\documentclass[nohyperref]{article}

\usepackage{microtype}
\usepackage{graphicx}
\usepackage{subfigure}
\usepackage{booktabs} 
\usepackage{hyperref}

\usepackage[accepted]{icml2022arxiv}

\usepackage{amsmath}
\usepackage{amssymb}
\usepackage{mathtools}
\usepackage{amsthm}
\usepackage{enumitem}
\usepackage{verbatim}
\usepackage{pifont}\newcommand{\cmark}{\ding{51}} \newcommand{\xmark}{\ding{55}} \usepackage{threeparttable}
\usepackage{multirow}
\usepackage{array}

\def\E{\mathop{\mathbf{E}\,\!}\nolimits}

\def\vec{\mathop{\rm vec}\nolimits}

\def\tr{\mathop{\rm tr}\nolimits}

\def\diag{\mathop{\rm diag}\nolimits}

\def\prox{\mathop{\rm prox}\nolimits}
\def\argmin{\mathop{\rm argmin}\nolimits}

\newtheoremstyle{custom}                {\topsep}                {\topsep}                {\itshape}                {}                {\bfseries}                {.}                {\newline}                {}
\newcommand\norm[1]{\left\lVert#1\right\rVert}
\newcommand{\iter}[2]{#1_{#2}}

\newtheorem{theorem}{Theorem}[section]

\newtheorem{lemma}{Lemma}[section]

\newtheorem{definition}{Definition}[section]

\newtheorem{proposition}{Proposition}[section]
\newtheorem{corollary}{Corollary}[section]

\newtheorem{remark}{Remark}[section]

\newtheorem{assumpA}{Assumption}
\newtheorem{assumpB}{Assumption}

\newenvironment{assumpAbis}[1]
  {\renewcommand{\theassumpA}{\ref{#1}$'$}   \addtocounter{assumpA}{-1}   \begin{assumpA}}
  {\end{assumpA}}

\usepackage[capitalize]{cleveref}
\Crefname{section}{Sect.}{Sects.}
\Crefname{figure}{Fig.}{Figs.}

\theoremstyle{plain}

\usepackage[textsize=tiny]{todonotes}

\icmltitlerunning{Statistical inference with implicit SGD}

\allowdisplaybreaks

\begin{document}

\twocolumn[
\icmltitle{Statistical inference with implicit SGD: \\
           proximal Robbins-Monro vs. Polyak-Ruppert}

\icmlsetsymbol{equal}{*}

\begin{icmlauthorlist}
\icmlauthor{Yoonhyung Lee}{equal,kec}
\icmlauthor{Sungdong Lee}{equal,snu}
\icmlauthor{Joong-Ho Won}{snu}
\end{icmlauthorlist}

\icmlaffiliation{kec}{Kakao Entertainment Corp.}
\icmlaffiliation{snu}{Department of Statistics, Seoul National University}

\icmlcorrespondingauthor{Joong-Ho Won}{wonj@stats.snu.ac.kr}

\icmlkeywords{implicit stochastic gradient descent, ISGD, stochastic proximal point}

\vskip 0.3in
]

\printAffiliationsAndNotice{\icmlEqualContribution} 
\begin{abstract}
The implicit stochastic gradient descent (ISGD), a proximal version of SGD, is gaining interest in the literature due to its stability over (explicit) SGD. In this paper, we conduct an in-depth analysis of the two modes of ISGD for smooth convex functions, namely proximal Robbins-Monro (proxRM) and proximal Poylak-Ruppert (proxPR) procedures, for their use in statistical inference on model parameters. Specifically, we derive non-asymptotic point estimation error bounds of both proxRM and proxPR iterates and their limiting distributions, and propose on-line estimators of their asymptotic covariance matrices that require only a single run of ISGD. 
The latter estimators are used to construct valid confidence intervals for the model parameters.
Our analysis is free of the  generalized linear model assumption that has limited the preceding analyses, and employs feasible procedures. Our on-line covariance matrix estimators appear to be the first of this kind in the ISGD literature.\end{abstract}

\section{Introduction}\label{sec:intro}
Consider the optimization problem of the form
\begin{equation}\label{eq:problem}
    \min_\theta L(\theta) := \E[\ell(Z, \theta)]
    \end{equation}
where $\theta \in \mathbb{R}^p$ is the variable (parameter) of interest,
$Z$ is a random variable, and $\E[\cdot]$ denotes the expectation over the distribution of $Z$.
Function $\ell$ is a real-valued \emph{sample function}, information on which can only be obtained by observing independent copies of $Z$.
For example, if $\ell$ refers to the negative log-likelihood of the model parameter $\theta$ given data $Z$, then problem \eqref{eq:problem} reduces to finding the ``true parameter''
$\iter{\theta}{\star}$ of the model.

A popular method for solving problem \eqref{eq:problem} is the stochastic gradient descent (SGD) method 
\begin{equation}\label{eq:sgd}    \theta_n = \theta_{n-1} - \gamma_n \nabla \ell(Z_n, \theta_{n-1})
    ,
\end{equation}
where the gradient is with respect to the second argument of $\ell$.
The $\gamma_n$ is the algorithm parameter called either \emph{step size} or \emph{learning rate}.
Assuming $\nabla L(\theta) = \E[\nabla\ell(Z, \theta)]$, $\nabla\ell(Z_n, \theta)$ is an unbiased estimator of $\nabla L(\theta)$, and SGD \eqref{eq:sgd} is an instance of stochastic approximation due to \citet{Robbins1951} for finding the root of $\nabla L$. 
The past decade witnessed a renewed interest in the Robbins-Monro procedure, mainly due to its adaptivity to large-scale data in machine learning problems \citep{Nemirovski2009,bottou2010large,bach2011,bottou2018optimization}. In particular, SGD and its variants are among the major driving forces of deep learning \citep{lecun2015deep,abadi2016tensorflow}. 
Its convergence property has been extensively studied \citep{zinkevich2003online,Nemirovski2009,bach2011}.
More recently, SGD has been studied as a tool for statistical inference from large datasets \citep{li2018statistical,liang2019statistical,chen2020statistical}.

A notable issue with SGD is its sensitivity to the step size selection. If it is too small, the convergence can be arbitrarily slow; if it is too large, then the iterate $\{\theta_n\}$ can diverge \citep{bach2011,RyuBoyd2014}.

As an alternative to SGD, consider the following iteration.
\begin{equation}\label{eq:spp}
    \theta_n = \prox_{\gamma_n \ell(Z_n, \cdot)}(\theta_{n-1}) 
    ,
\end{equation}
where
    $\prox_{\gamma f}(\theta) = \argmin_{\theta' \in \mathbb{R}^p}
    \left\{
        f(\theta') + \frac{1}{2\gamma}\norm{\theta' - \theta}^2
    \right\}	$
is the \emph{proximity operator} of $f$. Here the norm $\norm{\cdot}$ is the Euclidean $(\ell_2)$ norm. 
Using the optimality condition of the minimand in the definition of the operator,
iteration \eqref{eq:spp} can be written as an implicit equation
\begin{equation}\label{eq:implicit_sgd}    \theta_n = \theta_{n-1} - \gamma_n \nabla \ell(Z_n, \theta_n)
    .
	\tag{\theequation${}^\prime$}
\end{equation}
Iteration \eqref{eq:spp} can be considered as a noisy version of the proximal point algorithm due to \citet{rockafellar1976}, introduced to alleviate the sensitivity to the step size in the (noiseless) gradient descent method.
Being relatively new in the literature,
iteration \eqref{eq:spp} has been called in various names: incremental proximal method \citep{Bertsekas2011}, stochastic proximal point method \citep{RyuBoyd2014,bianchi2016}, and implicit SGD \citep[ISGD,][]{toulis2014, Toulis2017}. 
Convergence of ISGD is studied for the finite population case \citep{Bertsekas2011}. For infinite population, its stability (that the iterates do not diverge) and asymptotic rate of convergence are studied in \citet{toulis2014, RyuBoyd2014}. Non-asymptotic estimation error bounds are studied in \citet{Toulis2017,patrascu2017,asi2019stochastic}. 
These results can be summarized as that ISGD enjoys the same asymptotic rate of convergence as (explicit) SGD, while the former is much more stable and less sensitive to the choice of the step size.

Asymptotic normality of the ISGD is shown in \citet{Toulis2017} under several assumptions. 
However, the result of \citet{Toulis2017}
is limited by
the assumption that the sample function $\ell$ takes the form of $\ell(Z, \theta) = g(X^T \theta, Y)$ where $Z = (X, Y)$, which we shall call the \emph{generalized linear model (GLM) assumption}.
While the GLM model family is one of the most widely used in practice and the GLM assumption simplifies the computation of the proximity operator, it 
may not fit when dealing
with the general problem \eqref{eq:problem}. 
For example, consider estimating the $\alpha$ quantile of a univariate distribution with CDF $F(\theta)$. 
Then $\ell(Z, \theta) = \max(0, \theta - Z) - \alpha(\theta - Z)$ where $Z$ is drawn from $F$. This function does not satisfy the GLM assumption.
In addition, it is assumed that the Fisher information matrix coincides with the Hessian of the objective function $L$,
which is not in general true unless $\ell$ is the negative log-likelihood of $P$.
Furthermore, \citet{Toulis2017} assume that $L$ is both globally Lipschitz and globally strongly convex, a contradiction \citep{bach2011,asi2019stochastic}.

These restriction and contradiction are relaxed in some sense by \citet{toulis2021proximal}, who consider the \emph{proximal Robbins-Monro procedure} that idealizes ISGD:
\begin{equation}\label{eqn:PRM}
\begin{split}
        \theta_n^{+} &= \prox_{\gamma_n L}(\theta_{n-1}), \\
    \theta_n     &= \theta_{n-1} - \gamma_n \nabla \ell(Z_n, \theta_n^{+})
    ,
\end{split}
\end{equation}
and show asymptotic normality of $\theta_n$ under the assumption that $L$ is locally strongly convex at $\theta_{\star}$ and the \emph{gradient} of $L$ is globally Lipschitz \citep[Theorem 2.4]{toulis2021proximal}.
Nevertheless, procedure \eqref{eqn:PRM} is \emph{not feasible} since $\theta_n^{+}$ cannot be computed ($L$ is unknown or difficult to compute).
ISGD \eqref{eq:spp} can be understood as a plug-in procedure mimicking \eqref{eqn:PRM}, since $\theta_n$ in \eqref{eq:spp} is an unbiased estimator of $\theta_n^{+}$. 
In the sequel, we shall call the ISGD iteration \eqref{eq:spp} simply proximal Robbins-Monro (proxRM) and distinguish it from \eqref{eqn:PRM} by calling the latter the \emph{idealized} proxRM. Unlike the idealized counterpart, the consistency (convergence) and limiting distribution of proxRM has not been studied well, except for the special case mentioned above.

In the explicit SGD literature, averaging the iterates, known as the Polyak-Ruppert averaging \citep{polyak1992,ruppert1988} has been studied as a means to achieve the optimal asymptotic rate and adapt to large step sizes \citep{bach2011}.
Averaging the ISGD iterates \eqref{eq:spp}, i.e., taking $\iter{\bar{\theta}}{n} = \frac{1}{n}\sum_{k=0}^{n-1}\iter{\theta}{k}$, can thus be called the \emph{proximal Polyak-Ruppert} (proxPR) procedure.
Asymptotic normality of proxPR is shown by \citet{asi2019stochastic}. 
However, \emph{non-asymptotic} (point) estimation error bounds, which capture the transient behavior of $\{\iter{\bar{\theta}}{n}\}$, has not been studied well, although was conjectured to improve the rate \citep{toulis2021proximal}.

The goal of this paper is to fill these gaps in the literature, as well as to propose a device for efficient statistical inference with ISGD, in line with the similar works in explicit SGD \citep{li2018statistical,liang2019statistical,chen2020statistical}.
The latter goal is important since it enables to construct a valid confidence interval for the true parameter. Building the interval \emph{on-line} is also important as it means a single run of ISGD iteration suffices for legitimate inference.

\paragraph{Contributions.} 
Specifically, our contributions are as follows. 
(i) We extend the work by \citet{Toulis2017} on proxRM for GLM models to non-GLM settings. The relevant results include a non-asymptotic error bound on model parameter estimation, stability of the procedure, and its asymptotic normality under strong convexity. 
(ii) We elucidate that the above properties agree with those of the idealized proxRM \citep{toulis2021proximal}, thereby asserting the feasibility of the procedure.
(iii) We derive a non-asymptotic estimation error bound for proxPR, featuring the transient behavior of averaged ISGD.
(iv) We propose consistent \emph{online} estimators of the asymptotic covariance matrices of both proxRM and proxPR iterates, 
enabling statistical inference on the model parameter.
In addition to these results toward statistical inference, which requires strong convexity for identifiability of the  parameter,
(v) we provide a non-asymptotic analysis of the two procedures in the absence of strong convexity. In particular, we show that proxPR achieves the minimax optimal rate up to a logarithmic factor, confirming the conjecture by \citet{toulis2021proximal}.
\Cref{tab:contribution} summarizes the contributions.

\begin{table*}[ht]
\caption{Summary of contributions to implicit SGD. Learning rate schedule \eqref{enum:learning_rate} is assumed.}
\label{tab:contribution}
\begin{center}
\footnotesize
\begin{threeparttable}
\begin{tabular}{llllll}
\toprule
& GLM free & Feasibility & Non-asymptotic & Asympt. normality & Inference \\
\midrule
\citet{Toulis2017}         & \xmark & \cmark & \cmark~(RM) & \cmark~(RM) & \xmark \\
\citet{patrascu2017}       & \cmark & \cmark & \cmark~(RM) & \xmark & \xmark \\
\citet{asi2019stochastic}  & \cmark & \cmark & $\triangle$~(PR)\tnote{*} & \cmark~(PR) & \xmark     \\
\citet{toulis2021proximal} & \cmark & \xmark & \cmark~(RM) & \cmark~(RM) & \xmark  \\
This work                  & \cmark & \cmark & \cmark~(RM,~PR) & \cmark~(RM) & \cmark~(RM,~PR)   \\ 
\bottomrule
\end{tabular}
\begin{tablenotes}
   \item[*] Restricted to a certain class of ``easy'' problems.
	RM = proximal Robbins-Monro; PR = proximal Polyak-Ruppert.
\end{tablenotes}
\end{threeparttable}
\end{center}
\vskip -0.1in
\end{table*}

\section{Preliminary}\label{sec:prelim}

Let us begin with formally defining the objective function of problem \eqref{eq:problem}.
\[
    L(\theta) \triangleq \E[\ell(Z, \theta)]
    = \int_{\Omega} \ell(z,  \theta) dP(z)
    ,
\]
where $z$ is an element of the probability space $(\Omega, \mathcal{F}, P)$,
for which the probability measure $P$ can be considered as the distribution of a random variable $Z: \Omega \to \Omega: z \mapsto z$. 
In most practical situations $\Omega$ is the Euclidean space $\mathbb{R}^m$, $\mathcal{F}$ is the Borel sets of $\mathbb{R}^m$.
The sample function $\ell: \Omega\times\mathbb{R}^p \to \mathbb{R}$ is a real-valued. 
The ISGD procedure is implemented by sampling $Z_1, \dotsc, Z_n, \dotsc$ from $P$ independently and applying the update equation \eqref{eq:spp}. 
Finally, the filtration $\mathcal{F}_n$ is the smallest $\sigma$-algebra generated by $Z_1, \dotsc, Z_n$.

We make the following basic assumptions on the sample function $\ell$. 
\begin{assumpA}\label{enum:CCP} 
Function $\ell(z, \cdot)$ is real-valued convex function in $\mathbb{R}^p$ for each $z \in \Omega$;
Function $\ell(\cdot, \theta)$ is integrable for each $\theta \in \mathbb{R}^p$.
\end{assumpA}
\begin{assumpA}\label{enum:Lipschitz} 
Function $\ell(Z, \cdot)$ is $\beta(Z)$-smooth almost surely (a.s.), with $\E[\beta^2(Z)] = \beta_0^2  < \infty$ around a minimizer $\theta_{\star}$ of $L(\cdot) = \E{[\ell(Z, \cdot)]}$. That is, a sample function $\ell(Z, \theta)$ is continuously differentiable in $\theta$ and $\norm{\nabla \ell(Z,\theta) - \nabla \ell(Z,\theta_{\star})} \le \beta(Z) \norm{\theta - \theta_{\star}}$ for all $\theta$, with probability one.
\end{assumpA}

\begin{definition}[$M$-convexity \cite{RyuBoyd2014}]
    An extended real-valued function $f: \mathbb{R}^p \to \mathbb{R} \cup \{\infty\}$ is called $M$-convex at $x \in \mathbb{R}^p$ for a symmetric, positive semidefinite matrix $M \in \mathbb{R}^{p\times p}$ (denoted by $M \succeq 0$) if
    for $s \in \partial f(x)$
    \begin{equation}\label{eqn:strongcvx}
    f(y) \ge f(x) + s^T(y - x) + \textstyle\frac{1}{2}\norm{y - x}_M^2,
    \quad
    \forall y \in \mathbb{R}^p
    ,
    \end{equation}
    where
    $\norm{z}_M = (z^T M z)^{1/2}$.
\end{definition}
\begin{assumpA}\label{enum:M_str_conv} 
    Suppose $\theta_{\star}$ minimizes $L(\theta) = \E[\ell(Z, \theta)]$.
    Then $\ell(Z,\cdot)$ is $\Lambda(Z)$-convex at $\theta_{\star}$ a.s. with $\Lambda(Z) \succeq 0$ and $\Lambda_0 = \E[\Lambda(Z)]$ is positive definite so that $\lambda=\lambda_{\min}(\Lambda_0) > 0$,
    where $\lambda_{\min}(M)$ is the smallest eigenvalue of symmetric matrix $M$.
\end{assumpA}
\begin{assumpA}\label{enum:noiselevelcond} $\E{\|\nabla \ell(Z, \iter{\theta}{\star})\|^2} \le \sigma^2 < \infty$.
\end{assumpA}
\begin{remark}
If $f$ is differentiable, then condition \eqref{eqn:strongcvx} is equivalent to
$$
    (y - x)^T(\nabla f(y) - \nabla f(x)) \ge \norm{y - x}_M^2 
    .
$$
Observe that if $f$ is $M$-convex, then it is $\mu I$-convex with $\mu = \lambda_{\min}(M)$, where $I$ is the identity matrix. 
The latter is equivalent to the standard notion of $\mu$-convexity 
(if $\mu > 0$, then $f$ is strongly convex)
\cite{Bauschke:ConvexAnalysisAndMonotoneOperatorTheoryIn:2011}.
\end{remark}
\begin{remark}\label{rem:strongcvx}
From Assumptions \ref{enum:CCP}, \ref{enum:Lipschitz}, and \ref{enum:noiselevelcond} the objective function $L(\theta)=\E[\ell(Z, \theta)]$ is well-defined for all $\theta\in\mathbb{R}^p$. Furthermore, $L$ is continuously differentiable and its gradient has a representation $\nabla L(\theta) = \E[\nabla\ell(Z, \theta)]$ \citep{bertsekas1973}.
Assumption \ref{enum:M_str_conv} implies that $L$ is $\lambda$-convex at $\theta_{\star}$. 
Since $\lambda > 0$, it also implies that the minimizer $\theta_{\star}$ is unique.
\end{remark}
Throughout, we 
fix the learning rate schedule as follows.
\begin{equation}\label{enum:learning_rate}\tag{R}
\gamma_n = \gamma_1 n^{-\gamma}
\quad
\text{for some~} 
\gamma_1 > 0
\text{~and~}
\gamma > 0
.
\end{equation}
The valid range of the exponent $\gamma$ depends on the algorithm and the conditions on $L$; the subsequent discussions will elaborate on this.

\textbf{Notation}.
We employ the following asymptotic notation. 
For a sequence of random vectors/matrices $\{A_n\}$ defined on $(\Omega, \mathcal{F}, P)$ and a positive scalar sequence $\{b_n\}$, $A_n = O(b_n)$ means $\E[\norm{A_n}] \le c b_n$ for some $c > 0$ and for all $n=1, 2, \dotsc$. 
On the other hands, $A_n = o(b_n)$ means $\E[\norm{A_n}] / b_n \to 0$ as $n \to \infty$.
Here the norm $\norm{\cdot}$ refers to the (operator) 2-norm. 
Notation $b_n \downarrow 0$ means that $\{b_n\}$ is positive and converges monotonically toward zero.

\section{Approximation of ISGD by SGD}\label{sec:approx}
The results presented in \Cref{sec:strongcvx,sec:nonstronglycvx} rely on the following proposition bounding the difference between ISGD (proxRM) and (explicit) SGD iterates. This result holds without strong convexity.

\begin{proposition}[Approximation of ISGD by SGD]\label{thm:aproximation_ISGD}
In addition to Assumptions \ref{enum:CCP}, \ref{enum:Lipschitz}, and \ref{enum:noiselevelcond}, 
also assume that a minimizer $\iter{\theta}{\star}$ of $L$ exists (not necessarily unique).
Then, \vspace{-0.1cm}
\[
		\iter{\theta}{n} = 	\iter{\theta}{n-1} - \gamma_n\nabla\ell(Z_n,\iter\theta{n-1}) + \iter{R}{n}
		,
\]
\vspace{-0.1cm}
with
\begin{subequations}
    \begin{align}
    \E{[\norm{\iter{R}{n}} \!| \mathcal{F}_{n\!-\!1}]} \! &\leq \! \gamma_n^2\beta_0^2\norm{\iter{\theta}{n\!-\!1}\!-\!\iter{\theta}{\star}} \!+\! \textstyle\frac{\gamma_n^2}{2} (\beta_0^2\!+\!\sigma^2)     \label{eqn:Rn_conditional}
    \\
    \E{[\norm{\iter{R}{n}}^2 | \mathcal{F}_{n-1}]} &\leq  8\gamma_n^2\beta_0^2\norm{\iter{\theta}{n-1} - \iter{\theta}{\star}}^2 + 8 \gamma_n^2\sigma^2	\label{eqn:Rn2_conditional}
    \\
    \E{\norm{\iter{R}{n}}} &\leq \gamma_n^2[\beta_0^2(r + 1/2) + \sigma^2/2] \label{eqn:Rn_unconditional}
    \\
    \E{\norm{\iter{R}{n}}^2} &\leq  8\gamma_n^2(\beta_0^2 r^2 + \sigma^2),   \label{eqn:Rn2_unconditional}
    \end{align}
\end{subequations}
where
$r = \big(\norm{\theta_0 - \theta_{\star}}^2 + \sigma^2\sum_{k=1}^{\infty}\gamma_k^2\big)^{1/2}$. Inequalities \eqref{eqn:Rn_conditional} and \eqref{eqn:Rn2_conditional} hold for $\gamma \in (0, 1]$, and inequalities \eqref{eqn:Rn_unconditional} and \eqref{eqn:Rn2_unconditional} hold for $\gamma \in (1/2, 1]$.
\end{proposition}
\Cref{thm:aproximation_ISGD} relies on the following intermediate result:
\begin{lemma}\label{lem:difference}
    Under Assumptions \ref{enum:CCP}, we have
    $$
    \norm{\iter{\theta}{n} - \iter{\theta}{n-1}} =  \gamma_n\norm{\nabla\ell(Z_n, \iter{\theta}{n})} \leq \gamma_n\norm{\nabla\ell(Z_n, \iter{\theta}{n-1})}
    .
    $$
\end{lemma}

\Cref{thm:aproximation_ISGD} and \Cref{lem:difference} jointly play the role of Theorem 3.1 in \citet{Toulis2017}, which states that $\nabla\ell(Z_n, \iter{\theta}{n})$ has the same direction as $\nabla\ell(Z_n, \iter{\theta}{n-1})$  under the GLM assumption, expressing ISGD as a variant of SGD.
In the absence of this special relation, knowing that the norm of $R_n$ is $O(\gamma_n^2)$, not just $O(\gamma_n)$ as can be inferred from $\Vert R_n \Vert^2 = O(\gamma_n^2)$, is crucial since it allows (in principle) the techniques of bounding the estimation errors of explicit SGD \citep[e.g.,][]{bach2011} can be employed by controlling the impact of the additional term $R_n$ (e.g., \Cref{thm:finite_sample}).
Furthermore, in the proof of asymptotic normality (\Cref{thm:asymp_normality}), \Cref{thm:aproximation_ISGD} is used to obtain the $o(\gamma_n^{3/2})=n^{-\frac{3}{2}\gamma}\cdot o(1)$ error term in the recursive equation for the estimation error $\theta_n - \theta_{\star}$ (see \cref{eqn:normalapprox,eqn:fabianrecursion}). The resulting recursion admits the use of the central limit theorem due to \citet[Theorem 2.2]{Fabian1968} for classical stochastic approximation (including the explicit SGD) can be employed almost directly.
\Cref{thm:aproximation_ISGD} is also indispensable (through \Cref{thm:asymp_normality}) in showing consistency of the proposed on-line estimators of the asymptotic covariance matrices (\Cref{thm:plugin,cor:plugin}).
\section{Strongly convex objectives}\label{sec:strongcvx}
\subsection{Proximal Robbins-Monro}
\subsubsection{Stability}
Under Assumptions \ref{enum:CCP}--\ref{enum:noiselevelcond}, \citet{bianchi2016} and \citet[Proposition 3.8]{asi2019stochastic} show that the proxRM iterate $\theta_n$ converges to the unique solution $\theta_{\star}$ a.s. 
A finite-sample (non-asymptotic) mean-squared error bound of $\iter{\theta}{n}$ in estimating model parameter $\iter{\theta}{\star}$ is obtained by \citet[Theorem 14]{patrascu2017} for $\gamma \in (0, 1]$.
A simpler bound for $\gamma \in (1/2, 1]$ can be found as follows. We provide the proof in \Cref{sec:proofs} since it shows how the analysis of \citet{Toulis2017} extends to non-GLM settings (and without the contradictory assumptions).
\begin{definition}\label{def:phi}
$\phi_\gamma (n) \triangleq (n^{1 - \gamma} - 1)/(1 - \gamma)$ if $\gamma \neq 1$, and $\phi_\gamma (n) \triangleq \log n$ if $\gamma = 1$.
\end{definition}
\begin{theorem}[Non-asymptotic point estimation error bound]\label{thm:finite_sample}
Under Assumptions \ref{enum:CCP}--\ref{enum:noiselevelcond}, 
for any initial step size $\gamma_1 > 0$ when $\gamma \in (1/2, 1)$ and for $\gamma_1 > 1/(2\lambda)$ when $\gamma = 1$,
there exist a fixed integer $n_0$ and constants $K_1$, $D_{n_0}$ such that
\begin{align}\label{eq:finite_property}
     \E{\norm{\theta_n - \theta_{\star}}^2} &\leq
        K_1 n^{-\gamma} +  \exp\left(-\textstyle\frac{1}{2}\log(1+2\lambda\gamma_1) \phi_{\gamma}(n)\right)
		\nonumber\\
		&\quad
		\times
        ( \norm{\theta_0 - \theta_{\star}}^2 + D_{n_0} )
		,
		\quad 
		n \in \mathbb{N}
		.
\end{align}
\end{theorem}
All of the constants in inequality \eqref{eq:finite_property} are explicit, and are presented at the end of the proof given in \Cref{sec:proofs:RMerror}.

The result of \Cref{thm:finite_sample} can be summarized as $\E\norm{\theta_n - \theta_\star}^2 = O(\gamma_n)$. 
This asymptotic rate of $O(n^{-\gamma})$ matches that of SGD \citep{bach2011} and the idealized proxRM with the same learning rate schedule \eqref{enum:learning_rate} with $\gamma \in (0, 1]$.
For $\gamma < 1$, the effect of the initial point is forgotten at an exponential rate (second term in the right-hand side of the inequality \eqref{eq:finite_property}). 
For $\gamma = 1$, the rate is also $O(\gamma_n)$ provided that  $\gamma_1 \geq (e^2 - 1)/(2\lambda)$, although the ``exponential forgetting'' behavior is not so much prominent (it is polynomial).
Either way, this insensitivity to the initial step size
is one that features ISGD in contrast to SGD, where in the latter the impact of the initial point may \emph{exponentially increase} with the initial step size $\gamma_1$
in the transient phase
\cite{bach2011},
while in the former it always decreases.

The stability of proxRM can also be formalized in a non-MSE fashion. In order to analyze the error $\iter{\theta}{n} - \iter{\theta}{\star}$ (not $\norm{\iter{\theta}{n} - \iter{\theta}{\star}}^2$), we need an additional assumption:
\begin{assumpB}\label{enum:third_derivative} 
The objective function $L$ is twice differentiable at $\theta_{\star}$. 
\end{assumpB}    
The Hessian $\nabla^2 L$ of $L$ at $\theta_{\star}$ is denoted by $\mathcal{H}(\theta_{\star})$.
Note $0 < \lambda \leq \lambda_{\min}(\mathcal{H}(\iter{\theta}{\star}))$.
\begin{theorem}[Stability]\label{thm:stability}
Under Assumptions \ref{enum:CCP}--\ref{enum:noiselevelcond} and \ref{enum:third_derivative},
if $\gamma \in (1/2, 1)$ or $\gamma=1$ and $\gamma_1 \geq (e^2 - 1)/\lambda$, then
the proxRM iterate $\{\theta_n\}$ satisfies the following.
\begin{equation}\label{eqn:stability}
    \E[\theta_n - \theta_{\star}] = Q_1^n(\theta_0 - \theta_{\star}) + o(1)
\end{equation}
where $Q_1^n = \prod_{i=1}^n [I + \gamma_i \mathcal{H}(\theta_\star)]^{-1}$.
\end{theorem}
In contrast, if $\{\vartheta_n\}$ denotes the explicit SGD iterate (started from the same initial point), then
\[
    \E[\iter{\vartheta}{n} - \theta_{\star}] = P_1^n(\iter\theta{0} - \iter{\theta}{\star}),
    \quad
    P_1^n = \prod_{i=1}^n [I - \gamma_i \mathcal{H}(\iter{\theta}{\star})]
    ,
\]
ignoring the remainder term in the second-order Taylor expansion of $L$ at $\theta_{\star}$.
To avoid the explosion of the leading eigenvalue of $P_1^n$, it is desirable to control the initial step size $\gamma_1 < 2 / \lambda_{\max}(\mathcal{H}(\theta_{\star})) \leq 2/\lambda$.
In other words, in SGD the initial step size should not be large.
On the other hand, in proxRM the eigenvalues of $Q_1^n$ is always strictly smaller than $1$ \emph{regardless of the choice of $\gamma_1$} (when $\gamma < 1$). Thus in proxRM the step sizes can be taken large to promote fast convergence. 
A similar informal discussion can be found in \citep[Sect. 2.5]{Toulis2017}; we derive equation \eqref{eqn:stability} formally under weaker assumptions on differentiablility and without the GLM model.

\subsubsection{Inference}
\paragraph{Asymptotic normality.}
The following result generalizes that of \citet{Toulis2017} under weaker differentiability assumptions and without the GLM model. 
As already mentioned in \Cref{sec:approx},
the key instrument for deriving this result is \Cref{thm:aproximation_ISGD}, which quantifies the degree of approximation to SGD by ISGD. 
Our proof also fixes a flaw in the proof of \citet[Theorem 2.4]{Toulis2017}; see \Cref{rem:normalapprox}.

In order to establish asymptotic normality, we need additionally the following assumption on the stochastic error:
\begin{assumpB}\label{enum:Lindeberg}  Let $\sigma^2_{n,s} =  \E(I_{\norm{\varepsilon_n(\theta_\star)}^2 \ge s/\gamma_n} \norm{\varepsilon_n(\theta_\star)})$ where $\varepsilon_n(\theta_\star) = \nabla \ell(Z_n, \theta_\star) - \nabla L(\theta_{\star})$. Then for all $s > 0$, $\sum_{i=1}^n \sigma^2_{i,s} = o(n)$ if $\gamma = 1$, and $\sigma^2_{n,s} = o(1)$ otherwise.
\end{assumpB}

\begin{theorem}[Asymptotic normality]\label{thm:asymp_normality}
Suppose  Assumptions \ref{enum:CCP}-- \ref{enum:noiselevelcond} and 
\ref{enum:third_derivative}--\ref{enum:Lindeberg} hold. 
Then, 
the proxRM iterate $\theta_n$ is asymptotically normal, such that
\begin{equation}\label{eqn:CLT}
    n^{\gamma/2} 
    (\theta_n - \theta_\star) \stackrel{d}{\rightarrow} \mathcal{N}_p (0, \Sigma) 
\end{equation}
where 
\begin{equation}\label{eqn:asymptoticcovariance}
\Sigma = \begin{cases}
\gamma_1^2 \mathcal{L}_{2 \gamma_1 \mathcal{H}(\theta_\star) - I}^{-1} (\mathcal{I}(\theta_\star)), & \gamma = 1, ~ \gamma_1 \ge \frac{e^2-1}{\lambda}, 
\\
\gamma_1^2 \mathcal{L}_{ 2\gamma_1\mathcal{H}(\theta_\star)}^{-1} (\mathcal{I}(\theta_\star)), & \gamma \in (1/2, 1),
\end{cases}
\end{equation}
for $\mathcal{I}(\theta_{\star})=\E[\nabla\ell(Z, \theta_{\star})\nabla\ell(Z, \theta_{\star})^T]$.
Here, $\mathcal{L}_{P}^{-1}$ denotes the inverse operator of the Lyapunov linear map $\mathcal{L}_{P}(X) = \frac{1}{2}(PX + XP)$ for symmetric, positive definite matrix $P$.
\end{theorem}
\begin{remark}\label{rem:inversion}
    The inverse of the Lyapunov map has a closed form:
    $$
        \Vec(\mathcal{L}_{2B}^{-1}(Y))
        = (I\otimes B + B \otimes I)^{-1}\Vec(Y)
    $$
    where 
    $\otimes$ denotes the Kronecker product and $\Vec(\cdot)$ refers to the usual vectorization operator for matrices. 
\end{remark}
\Cref{thm:asymp_normality} emphasizes that, in order for proxRM to converge (in distribution), $\gamma_1 > 1/(2\lambda) \geq 1/[2\lambda_{\min}(\mathcal{H}(\theta_{\star}))]$ is necessary, at least for $\gamma=1$. 
(Note $\gamma_1 \ge (e^2-1)/\lambda$ implies this condition.)
Thus in proxRM a large initial step is \emph{promoted}, rather than prohibited for a stability concern as in explicit SGD; see the discussion after \Cref{thm:stability}.

\paragraph{Estimation of the asymptotic covariance matrix.}
An obvious way of consistently estimating the the covariance matrix $\Sigma$ of the asymptotic distribution \eqref{eqn:CLT} is to run $n$ proxRM  iterations $B$ times independently, and take
\begin{equation}\label{eqn:empirical_cov}
    	\textstyle
	B^{-1}
	\sum_{i=1}^{B} (\theta_n^{(i)} - \bar{\theta})(\theta_n^{(i)} - \bar{\theta})^T
    ,
\end{equation}
where $\theta_n^{(i)}$ denotes the $n$th iterate for the $i$th run.
This estimator is of course not very practical since many runs are required. 
Here we show a consistent estimator based on only a \emph{single} run can be constructed.
Specifically, we propose to use
\[
\hat{\Sigma}_n = \begin{cases}
\gamma_1^2 \mathcal{L}_{2 \gamma_1 \hat{H}_n - I}^{-1} (\hat{I}_n), & \gamma = 1, ~ \gamma_1 \ge \frac{e^2-1}{\lambda}, 
\\
\gamma_1^2 \mathcal{L}_{ 2\gamma_1\hat{H}_n}^{-1} (\hat{I}_n), & \gamma \in (1/2, 1),
\end{cases}
\]
where 
\begin{equation}\label{eqn:plugin_H_I}
\begin{split}
\hat{H}_n &= \frac{1}{n}\sum_{k=1}^n \nabla^2\ell(Z_k, \theta_{k-1}),
\\
\hat{I}_n &= \frac{1}{n}\sum_{k=1}^n \nabla\ell(Z_k, \theta_{k-1})\nabla\ell(Z_k, \theta_{k-1})^T
\end{split}
\end{equation}
are plug-in estimators of $\mathcal{H}(\theta_\star)$ and $\mathcal{I}(\theta_\star)$, respectively. It is clear that both $\hat{H}_n$ and $\hat{I}_n$ can be computed in an on-line fashion, so can $\hat{\Sigma}_n$. Since $\hat{H}_n$ may not be positive definite in practice, a bit of adjustment on its eigenvalues may be needed to ensure invertibility (see \Cref{rem:inversion}).

In \Cref{thm:plugin} below, we show that with additional regulatory assumptions along with the positive-definite adjustment, $\hat{\Sigma}_n$ is a consistent estimator of $\Sigma$, from which valid inference on $\theta_{\star}$ can be implemented.
The following assumptions are adopted from \citet{chen2020statistical}.
\begin{assumpB}\label{enum:fourthmoment}
    For the error sequence $\varepsilon_n = \nabla L(\iter{\theta}{n-1}) - \nabla\ell(Z_n, \iter{\theta}{n-1})$, the fourth conditional moment is bounded as follows.
    \[
        \E[\norm{\varepsilon_n}^4|\mathcal{F}_{n-1}] \le \Sigma_3 + \Sigma_4\norm{\iter{\theta}{n-1} - \iter{\theta}{\star}}^4
    \]
    for some constants $\Sigma_3$ and $\Sigma_4$.
\end{assumpB}
\begin{assumpB}\label{enum:Hessian}
The sample function $\ell(Z, \cdot)$ is
twice differentiable a.s., and is $M$-Lipschitz continuous at the minimizer $\theta_{\star}$ of $L(\cdot) = \E{[\ell(Z, \cdot)]}$. 
That is, $\nabla^2\ell(Z, \theta)$ exists for all $\theta$ and $\norm{\nabla^2 \ell(Z,\theta) - \nabla^2 \ell(Z,\theta_{\star})} \le M \norm{\theta - \theta_{\star}}$ for all $\theta$, with probability one.
\end{assumpB}
\begin{assumpB}\label{enum:hessian}
    The sample function $\ell(Z, \cdot)$ is twice differentiable a.s.
                        	The second moment of the Hessian is bounded:
    \[
        \norm{\E[(\nabla^2\ell(Z, \iter{\theta}{\star}))^2] - [\mathcal{H}(\iter{\theta}{\star})]^2} \le L_4
    \]
    for some constant $L_4$.
\end{assumpB}

\begin{remark}
According to Lemma 3.1 of \citet{chen2020statistical}, Assumption \ref{enum:fourthmoment} is satisfied if $||\nabla^2\ell(\theta,Z)|| \leq H(Z)$ for some $H$ with a bounded fourth moment, Thus, it can be easily checked that these assumptions hold for common losses such as quadratic or logistic loss.
In fact Assumptions \ref{enum:fourthmoment}--\ref{enum:hessian} correspond to part 3 of Assumption 3.2 and Assumption 4.1 of \citet{chen2020statistical}. Note part 2 of their Assumption 3.2 is not used in the present paper. \end{remark}

\begin{theorem}[Consistency of plug-in estimator]\label{thm:plugin}
    Suppose Assumptions \ref{enum:CCP}--\ref{enum:noiselevelcond},
\ref{enum:third_derivative}, and \ref{enum:fourthmoment}--\ref{enum:hessian} hold.
        Let $\hat{H}_n$ and $\hat{I}_n$ be as given in \Cref{eqn:plugin_H_I}.
    Let $\beta_{+} = 1$ if $\gamma=1$ and $\beta_{+}=0$ otherwise.
    Choose $\delta \in (\beta_{+}/(2\gamma_1), \lambda_{\min}(\mathcal{H}(\theta_{\star})))$ and let
    $\tilde{H}_n = P\diag(\max(d_1, \delta),\allowbreak  \dotsc,\allowbreak \max(d_p, \delta))P^T$ for the spectral decomposition $P\diag(d_1, \dotsc, d_p)P^T$ of $\hat{H}_n$. Then, for the asymptotic covariance matrix  \eqref{eqn:asymptoticcovariance},
    \[
        \E\norm{ \gamma_1^2\mathcal{L}_{2\gamma_1\tilde{H}_n - I}^{-1}(\hat{I}_n)  - \Sigma         }
        = O(\gamma_n^{1/2})
                                            \]
    if $\gamma=1$,     $\gamma_1 \geq \frac{e^2-1}{\lambda}$,
    $2\gamma_1\mathcal{H}(\theta_{\star}) \succ I$;
    if $\gamma \in (\frac{1}{2}, 1)$,
    \[
        \E\norm{\gamma_1^2\mathcal{L}_{2\gamma_1\tilde{H}_n}^{-1}(\hat{I}_n) - \Sigma 
                }
        = O(\gamma_n^{1/2})
                                                .
    \]
    \end{theorem}
Thus the $100(1-\alpha)\%$ confidence interval for the $j$-th component of $\iter{\theta}{\star}$ can be approximated by
$\theta_{n,j} \pm z_{\alpha/2}\hat{\sigma}_{n,j}$ where $z_{\alpha/2}$ is the $1-\alpha/2$ quantile of the standard normal distribution, and
\begin{equation}\label{eqn:proxRM-sigma}
    \hat{\sigma}_{n,j} =
    \begin{cases}
        n^{-1/2}\gamma_1\sqrt{[\mathcal{L}_{2\gamma_1\tilde{H}_n - I}^{-1}(\hat{I}_n)]_{jj}},
        & \gamma = 1,
        \\
        n^{-\gamma/2}\gamma_1\sqrt{[\mathcal{L}_{2\gamma_1\tilde{H}_n}^{-1}(\hat{I}_n)]_{jj}},
        & \gamma\in(1/2, 1)
        .
    \end{cases}
\end{equation}

\subsection{Proximal Polyak-Ruppert}
\subsubsection{Stability}
With slightly stronger assumptions on the objective function, the stability of proxPR can be analyzed:
\begin{assumpAbis}{enum:Lipschitz}\label{enum:Lipschitz2}
Function $\ell(Z, \cdot)$ is $\beta_0$-smooth a.s. around the minimizer $\theta_{\star}$ of $L(\cdot) = \E{[\ell(Z, \cdot)]}$. That is, a sample function $\ell(Z, \theta)$ is continuously differentiable in $\theta$ and $\norm{\nabla \ell(Z,\theta) - \nabla \ell(Z,\theta_{\star})} \le \beta_0 \norm{\theta - \theta_{\star}}$ for all $\theta$, with probability one.
\end{assumpAbis}
\begin{assumpAbis}{enum:noiselevelcond}\label{enum:noiselevelcond2} 
$\E{\|\nabla \ell(Z, \iter{\theta}{\star})\|^4} \le \sigma^4 < \infty$.
\end{assumpAbis}
Note Assumption \ref{enum:Lipschitz2} (\textit{resp}. \ref{enum:noiselevelcond2}) implies Assumption \ref{enum:Lipschitz} (\textit{resp}. \ref{enum:noiselevelcond}).
\begin{theorem}[Non-asymptotic point estimation error bound]\label{thm:polyakruppert}
Under Assumptions \ref{enum:CCP}, \ref{enum:Lipschitz2}, \ref{enum:M_str_conv},  \ref{enum:noiselevelcond2}, \ref{enum:third_derivative}, and \ref{enum:Hessian}, for $\iter{\bar{\theta}}{n} = \frac{1}{n}\sum_{k=0}^{n-1}\iter{\theta}{k}$, the following holds for $\gamma \in (1/3, 1)$.
\begin{align}\label{eqn:proxPRbound}
    &(\E{\norm{\iter{\bar{\theta}}{n}-\iter{\theta}{\star}}^2})^{1/2}
    \leq
	\textstyle
    \frac{1}{\sqrt{n}}\left[\tr{(\mathcal{H}(\iter{\theta}{\star})^{-1}\mathcal{I}(\iter{\theta}{\star})\mathcal{H}(\iter{\theta}{\star})^{-1})}\right]^{1/2}
	\nonumber\\
    &\quad
    +
	\textstyle
    \frac{K_1^{1/2}}{\lambda^{1/2}n}
    \Big(
    (\beta_0 + \gamma_1^{-1})n^{\gamma/2}
    +
    2\beta_0\phi_{\gamma}^{1/2}(n)   
	\nonumber\\
	&\quad\quad\quad\quad\quad
    +
	\textstyle
    \gamma(\beta_0 + \gamma_1^{-1})
    \phi_{1-\gamma/2}(n)
    +
    \frac{M}{2}\phi_{\gamma}(n)
    \Big)
	\nonumber\\
    &\quad
    + 
	\textstyle
	\frac{\tilde{\mathcal{A}}}{\lambda^{1/2}n}
    +
	\textstyle
	\frac{M\tilde{\mathcal{B}}}{2\lambda^{1/2}n}
	\nonumber\\
    & \quad
    +
	\textstyle
    \frac{\beta_0 + \gamma_1^{-1}}{\lambda^{1/2}}
    \exp\big(-\frac{1}{4}\log(1+2\lambda\gamma_1) \phi_{\gamma}(n)\big)
	\nonumber\\
    & \quad\quad\quad\quad\quad
    \qquad\qquad
	\times
    ( \norm{\theta_0 - \theta_{\star}}^2 + D_{n_0} )^{1/2}   
    ,
\end{align}
where $n_0, K_1, D_{n_0}$ are the same as in \Cref{thm:finite_sample}.
\end{theorem}
The constants $\tilde{\mathcal{A}}$ and $\tilde{\mathcal{B}}$ in inequality \eqref{eqn:proxPRbound} are explicit, and are presented at the end of the proof given in \Cref{sec:proofs:PRerror}.

Compared with the (explicit) SGD \citep[Theorem 3]{bach2011}, the allowed range of $\gamma$ is a bit narrower. (This is due to \Cref{lemma:toulis} and \Cref{cor:toulis} in the \Cref{sec:proofs}.)
However, when $M > 0$, to obtain the optimal $O(1/n)$ rate (in mean square error) independent of $\gamma_n$, we need $\gamma \in (1/2, 1]$, just as SGD (for $\gamma = 1$ we get a simpler bound by averaging the bound in \Cref{thm:finite_sample}); when $M = 0$ ($L$ is quadratic) then the rate is $O(1/n)$ for all $\gamma \in (1/3, 1]$.
The second slowest term has an order of either $O(n^{-(2-\gamma)})$ or $O(n^{-(1+\gamma)})$ due to the second line in inequality \eqref{eqn:proxPRbound}, suggesting $\gamma = 2/3$ to get a balance.

The rate of ``forgetting the initial condition'' consists of two parts, one with the rate of $O(1/n^2)$ involving quantities $\tilde{\mathcal{A}}$ and $\tilde{\mathcal{B}}$ and the other with an exponential rate of $O(\exp\left(-\textstyle\frac{1}{2}\log(1+2\lambda\gamma_1) \phi_{\gamma}(n)\right))$. 
Furthermore, the constant $\tilde{B}$ can be quite large: it involves the term exponential in $\phi_{\frac{5}{3}\gamma}(k)$, which is increasing if $\gamma < 3/5$ (see \Cref{sec:proofs}).
Thus, unlike proxRM, the impact of the initial condition many remain for a long time, and this is what we observed in the experiments in \Cref{sec:experiments}. 
Like in SGD, having a burn-in period before taking averages appears to be beneficial.

The key difference between proxPR the forward Polyak-Ruppert (averaged explicit SGD) is that there is no exponential term multiplied to the constant $\tilde{\mathcal{B}}$ in \eqref{eqn:proxPRbound}, which corresponds to the constant ``$A$'' in Theorem 3 of \citet{bach2011}. \emph{Thus there is no ``catastrophic term'' in proxPR.} This suggests that we can take $\gamma_1$ large.

\begin{remark}
In \citet{toulis2016}, a non-asymptotic error bound with rate $O(1/n)$ for any $\gamma \in (1/2, 1]$
is obtained for the averaged iterate, however under the contradictory assumption that the objective function $L$ is both globally Lipschitz and globally strongly convex, as well as the GLM assumption. In particular, their Lemma 5 crucially depends on the Lipschitz continuity of $L$.
\end{remark}

\subsubsection{Inference}
\paragraph{Asymptotic normality.}
The asymptotic normality of proxPR, i.e., that of $\bar{\theta}_n = \frac{1}{n}\sum_{k=0}^{n-1}\theta_k$, is established in \citet[Theorem 3.11]{asi2019stochastic} for $\gamma\in(1/2, 1)$. The asymptotic covariance matrix is  $n^{-1}\mathcal{H}(\iter{\theta}{\star})^{-1}\mathcal{I}(\iter{\theta}{\star})\mathcal{H}(\iter{\theta}{\star})^{-1}$, which achieves the Cram{\'e}r-Rao lower bound.
Its rate of vanishing matches that of the asymptotic covariance matrix of the proxRM iterate $\iter{\theta}{n}$, but $\iter{\bar{\theta}}{n}$ is statistically more efficient.

\paragraph{Estimation of the asymptotic covariance matrix.}
The proof of \Cref{thm:plugin} involves consistency of $\hat{H}_n$ and $\hat{I}_n$ in estimating $\mathcal{H}(\iter{\theta}{\star})$ and $\mathcal{I}(\iter{\theta}{\star})$ respectively (see \Cref{lem:consistency}) and that $\tilde{H}_n$ is asymptotically equivalent to $\hat{H}_n$.
Hence $\tilde{H}_n^{-1}\hat{I}_n\tilde{H}_n^{-1}$ is a consistent estimator of  $\mathcal{H}(\iter{\theta}{\star})^{-1}\mathcal{I}(\iter{\theta}{\star})\mathcal{H}(\iter{\theta}{\star})^{-1}$, the  asymptotic covariance matrix of the scaled Polyak-Ruppert average 
$\sqrt{n}\bar{\theta}_n = \frac{1}{\sqrt{n}}\sum_{k=0}^{n-1}\theta_k$.
It follows:
\begin{corollary}[Plug-in estimator]\label{cor:plugin}
    Suppose Assumptions \ref{enum:CCP}--\ref{enum:noiselevelcond},
\ref{enum:third_derivative}, and \ref{enum:fourthmoment}--\ref{enum:hessian} hold.
	Then, for $\tilde{H}_n$ and $\hat{I}_n$ as appeared in \Cref{thm:plugin}
    and for $\gamma \in (1/2, 1)$,
    \[
        \E\norm{ \tilde{H}_n^{-1}\hat{I}_n\tilde{H}_n^{-1}\  - \mathcal{H}(\iter{\theta}{\star})^{-1}\mathcal{I}(\iter{\theta}{\star})\mathcal{H}(\iter{\theta}{\star})^{-1} }
        = O(\gamma_n^{1/2})
        .
    \]
\end{corollary}
An asymptotic $100(1-\alpha)\%$ confidence interval for the $j$-th component of $\iter{\theta}{\star}$ is given by
$\bar{\theta}_{n,j} \pm z_{\alpha/2}\tilde{\sigma}_{n,j}$ where
\begin{equation}\label{eqn:proxPR-sigma}
    \tilde{\sigma}_{n,j} =
    n^{-1/2}\sqrt{[\tilde{H}_n^{-1}\hat{I}_n\tilde{H}_n^{-1}]_{jj}}
    ,
    \quad
    \gamma \in (1/2, 1)
    .
\end{equation}

\section{Non-strongly convex objectives}\label{sec:nonstronglycvx}

\subsection{Proximal Robbins-Monro}
In this and the next subsection, we do not assume that the objective function $L$ is strongly convex. However we do assume that the minimum is attained. In other words, we replace Assumption \ref{enum:M_str_conv} with a weaker one:
\begin{assumpAbis}{enum:M_str_conv}\label{enum:nonstrconv}
	There exists a minimizer $\iter{\theta}{\star}$ of the objective $L$. The minimizer may not be unique.
\end{assumpAbis}
Since the minimizer is not unique, we derive a finite-sample bound on the objective value.
\begin{theorem}[Non-asymptotic optimization error bound]\label{thm:nonstronglycvxRM}
Under Assumptions \ref{enum:CCP}, \ref{enum:Lipschitz2}, \ref{enum:nonstrconv}, and \ref{enum:noiselevelcond}, the following holds for the proxRM iterate \eqref{eq:spp}:
\begin{align*}
	&\E{[L(\iter{\theta}{n}) - L(\iter{\theta}{\star})]}
	\leq
	\\
	&
	        \begin{cases}
                        \Gamma_1 \cdot
            \frac{\delta_0 + \gamma_1^2(1+\phi_{2\gamma}(n))}{\phi_{1-\gamma/2}(n) - \phi_{1-\gamma/2}(n_1-1)}, & \gamma \in \textstyle (0, \frac{1}{2}],
            \\
                        \Gamma_2 \cdot
            \frac{\delta_0 + \gamma_1^2\zeta(2\gamma)}{\phi_{1-\gamma/2}(n) - \phi_{1-\gamma/2}(n_1-1)}, & \gamma \in \textstyle (\frac{1}{2}, \frac{2}{3}),
            \\
                        \Gamma_3 \cdot
            \frac{\delta_0 + \gamma_1^2\zeta(2\gamma)}{\phi_{\gamma}(n) - \phi_{\gamma}(n_1-1)}, & \gamma \in \textstyle [\frac{2}{3}, 1],
        \end{cases}
\end{align*}
    for $n \geq n_1 \triangleq \max\{\inf\{n \in \mathbb{N}: \gamma_n \leq 1/\beta_0\}, 3\}$,
	where $\delta_0 = \norm{\iter{\theta}{0} - \iter{\theta}{\star}}^2$; 
	$\zeta(\alpha) = \sum_{k=1}^{\infty}k^{-\alpha}$ is the Riemann zeta function.
\end{theorem}
The constants $\Gamma_1$, $\Gamma_2$, and $\Gamma_3$ are explicit, and are presented at the end of the proof given in \Cref{sec:proofs:RMerror2}.

This bound is decreasing only if $\gamma > 2/5$.
Thus the rate is $O(n^{-(1 - 5\gamma/2)})$ if $2/5 < \gamma < 1/2$, $O(n^{-1/4}\log n)$ if $\gamma=1/2$, $O(n^{-\gamma/2})$ if $1/2 < \gamma < 2/3$, $O(n^{-(1-\gamma)})$ if $2/3 < \gamma < 1$, and $O(1/\log n)$ if $\gamma = 1$.
    The best rate is attained for $\gamma = 2/3$ which is $O(n^{-1/3})$.
	This is the same rate as the explicit SGD counterpart \citep[Theorem 4]{bach2011}, while the latter is valid for $\gamma \in [1/2, 1]$. Furthermore, no catastrophic term appears in the bounds in \Cref{thm:nonstronglycvxRM}.

The convergence rates and stability implied in \Cref{thm:nonstronglycvxRM} fully agree with \citet[Theorem 2]{toulis2021proximal} obtained for the idealized, but infeasible, proxRM procedure ($\gamma \in (1/2, 1]$). Therefore, realizing the ideal procedure \eqref{eqn:PRM} by equation \eqref{eq:spp} loses nothing.

\subsection{Proximal Poylak-Ruppert}
\begin{theorem}[Non-asymptotic optimization error bound]\label{thm:nonstronglycvxPR}
Under Assumptions \ref{enum:CCP}, \ref{enum:Lipschitz2}, \ref{enum:nonstrconv}, and \ref{enum:noiselevelcond}, the following holds.
	\begin{align*}
	&\E{[L(\iter{\bar{\theta}}{n}) - L(\iter{\theta}{\star})]}
    \leq
	\\
	&
	\begin{cases}
						\frac{\tilde{\Gamma}_1}{n}
                        + \frac{\sigma^2\gamma_1^2}{(1 - 2\gamma)n^{\gamma}}
            + \frac{2\tilde{\sigma}^2\gamma_1}{(1 - \gamma)n^{\gamma}},
			            &
            \gamma \in \textstyle (0, \frac{1}{2}),
            \\
						\frac{\tilde{\Gamma}_2}{n}
                        + \frac{\sigma^2\gamma_1^2(1 + \log n)}{\sqrt{n}}
            + \frac{2\tilde{\sigma}^2\gamma_1}{(1 - \gamma)\sqrt{n}},
			            &
            \gamma = \textstyle \frac{1}{2}
            ,
			\\
			\frac{\tilde{\Gamma}_3}{n}
            + \frac{\sigma^2\gamma_1^2\zeta(2\gamma)}{n^{1-\gamma}}
            + \frac{2\tilde{\sigma}^2\gamma_1}{(1-\gamma)n^{\gamma}},
            & \gamma \in \textstyle (\frac{1}{2}, 1),
            \\
            \frac{\tilde{\Gamma}_3}{n}
                        + \sigma^2\gamma_1^2\zeta(2)
            + \frac{2\tilde{\sigma}^2\gamma_1\log n}{(1-\gamma)n},
            & \gamma = 1
    \end{cases}
	\end{align*}
for $n \geq n_* \triangleq \inf\{k \in \mathbb{N}: (1 - \gamma_k\beta_0) \geq 1/2 \}$.
Also,
    \begin{align*}
        &\E{[L(\iter{\bar{\theta}}{n}) - L(\iter{\theta}{\star})]}
        \leq
		\\
        &
        \begin{cases}
            \frac{\beta_0}{2}\Big(
            \delta_0 + \frac{\sigma^2\gamma_1^2}{(1-2\gamma)(2 - 2\gamma)}n^{1-2\gamma}
            \Big), & \gamma \in \textstyle (0, \frac{1}{2}), 
            \\
            \frac{\beta_0}{2}\Big(\delta_0 + \sigma^2\gamma_1^2\frac{\log(n+1)}{n} \Big), & \gamma = \textstyle \frac{1}{2}, 
            \\
            \frac{\beta_0}{2}\Big(\delta_0 + \frac{\sigma^2\gamma_1^2\zeta(2\gamma)}{n} \Big), & \gamma \in \textstyle (\frac{1}{2}, 1],
        \end{cases}
    \end{align*}
for $n < n_*$, where $\delta_0 = \norm{\iter{\theta}{0} - \iter{\theta}{\star}}^2$.
\end{theorem}
The constants $\tilde{\Gamma}_1$, $\tilde{\Gamma}_2$,  $
\tilde{\Gamma}_3$, and $\tilde{\sigma}^2$ are explicit, and are presented at the end of the proof given in \Cref{sec:proofs:PRerror2}.

The convergence rate can be summarized as $O(n^{-\gamma}\vee n^{-(1-\gamma)})$.
Thus with $\gamma=1/2$, we attain the minimax optimal asymptotic rate of $O(n^{-1/2})$ \citep{ruppert1988} up to a logarithmic factor.
We have thereby confirmed the conjecture by \citet[Remark 3.1]{toulis2021proximal} that the minimax rate may be achieved by averaging the (ideal) proxRM iterates, i.e., proxPR. Moreover, our procedure is feasible.
Compared to explicit SGD \citep[Theorem 6]{bach2011}, again no catastrophic term is involved and the initial step size $\gamma_1$ can be taken large to obtain the best rate.

\section{Experiments}\label{sec:experiments}
\subsection{Point estimation and optimization}
\begin{figure}[ht!]
    \centering
        \includegraphics[width=0.9\columnwidth]{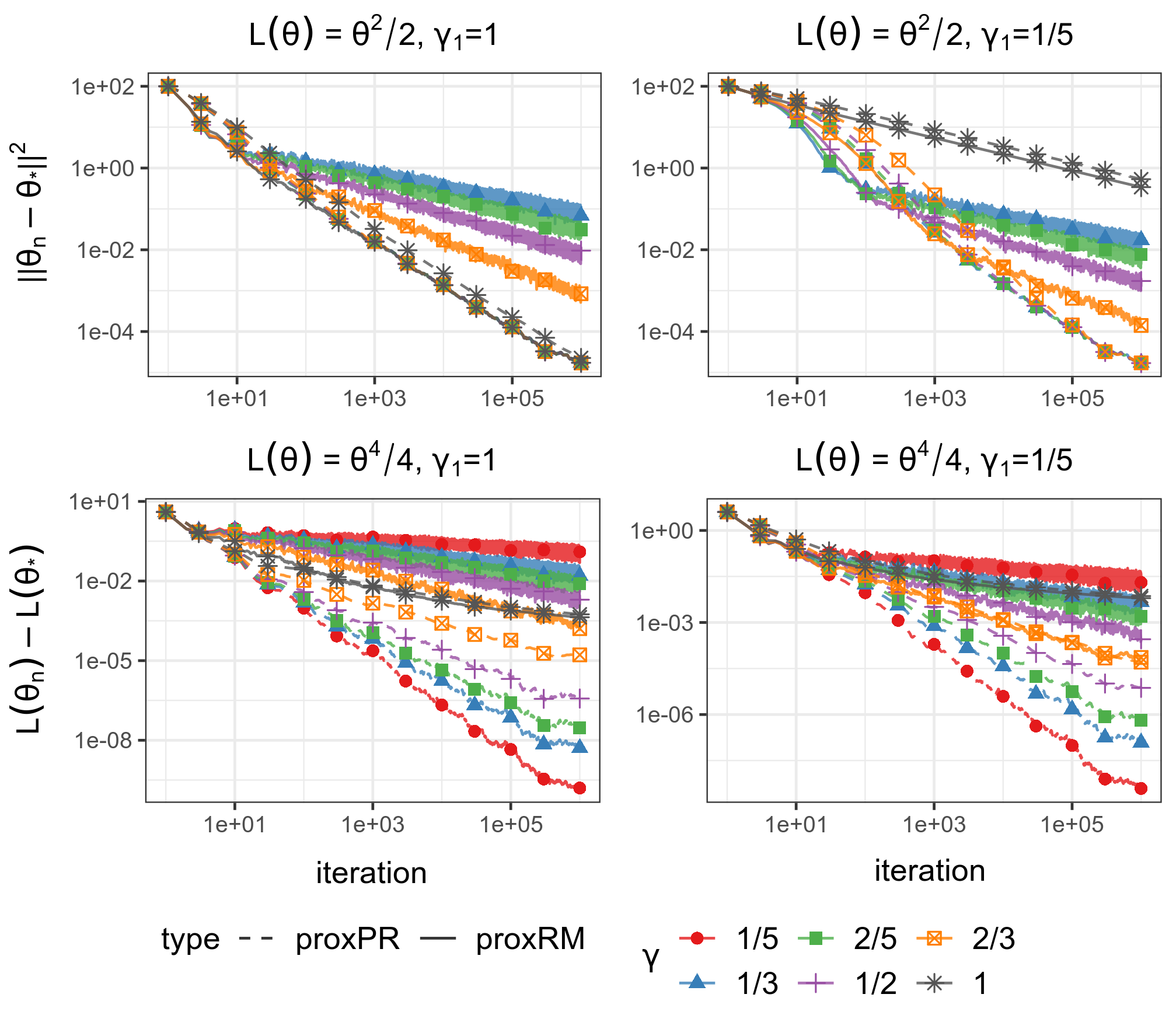}
        \caption{Error reduction for various learning rates.}
    \vskip -0.15in
    \label{fig:mse-fixed-gamma1}
\end{figure}
\begin{figure}[ht!]
    \centering
        \includegraphics[width=0.9\columnwidth]{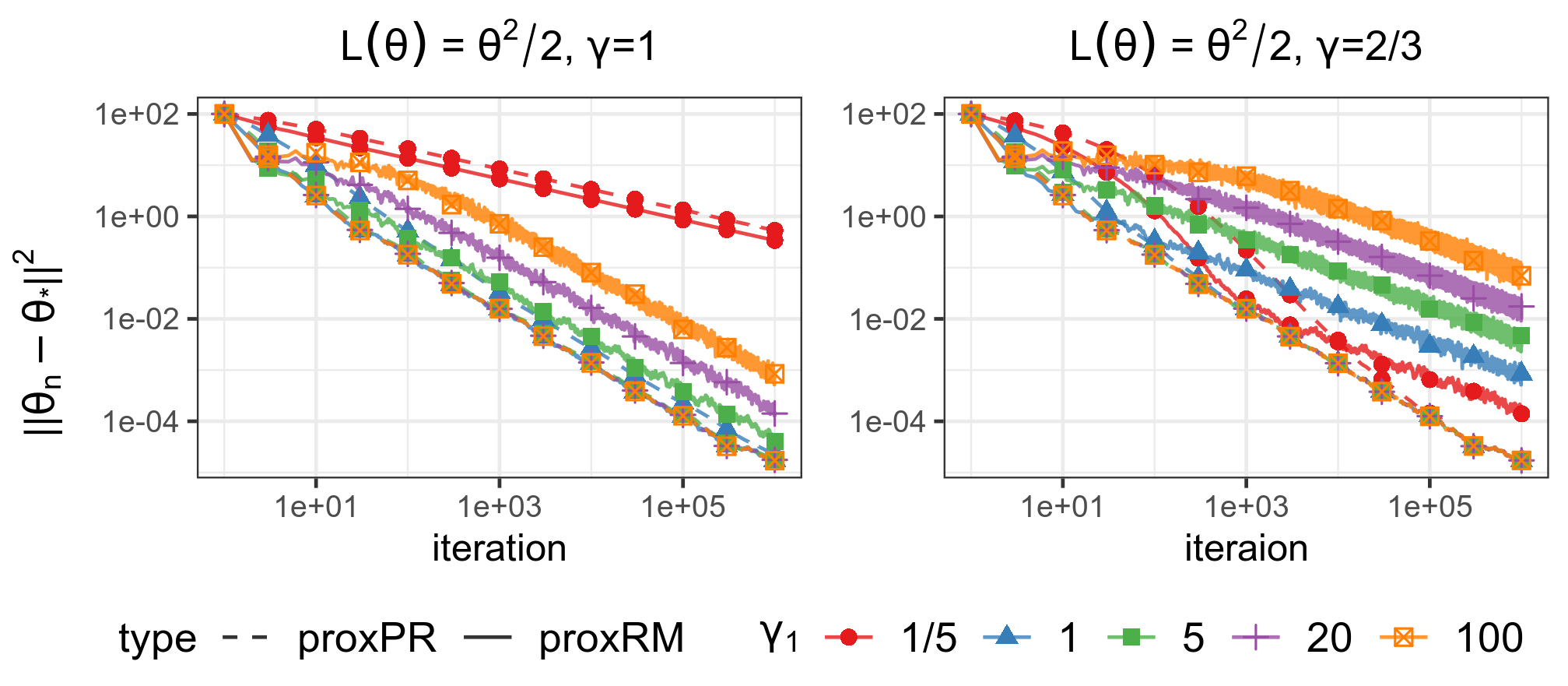}
    \vskip -0.15in
        \caption{Error reduction for various values of initial step size.}
    \label{fig:mse-fixed-gamma}
\end{figure}
Following \citet{bach2011}, we examined the convergence behavior of proxRM and proxPR using two univariate functions: $L(\theta) = \frac{1}{2}\theta^2$ (strongly convex) and $L(\theta) = \frac{1}{4}\theta^4$ (non-strongly convex);
the sample functions were chosen $\ell(Z, \theta) = \frac{1}{2}\theta^2 + Z\theta$ and $\ell(Z, \theta) = \frac{1}{4}\theta^4 + Z\theta$, where $Z\sim N(0,4^2)$.
We fixed the initial point $\theta_0 = 10$ for the quadratic and $\theta_0 = 2$ for the quartic function, 
and observed 100 independent runs of $n=10^6$ ISGD iterations
for initial step size $\gamma_1 \in \{1/5, 1, 5, 20, 100\}$ and exponent $\gamma \in \{1/5,  1/3, 2/5, 1/2, 2/3, 1\}$.

\Cref{fig:mse-fixed-gamma1,fig:mse-fixed-gamma} plot the squared estimation ($\norm{\iter{\theta}{n} - \iter{\theta}{\star}}^2$) or optimization ($L(\iter{\theta}{n}) - L(\iter{\theta}{\star}$)) error averaged over all replications for each iteration.
The trend generally aligns with the results in \Cref{sec:strongcvx,sec:nonstronglycvx}.
First, several instances of proxPR showed slower reduction than proxRM in the early stage of the iteration, as expected by the theory.
For the strongly convex case, the error reduction was proportional to $\gamma$ in proxRM and at the same fastest convergence rate in proxPR for all values except for $\gamma=1$, for which too small an initial step size showed a noticeable slowdown.
For a given $\gamma$, the rate was proportional to $\gamma_1$ in proxRM, ending up with parallel lines in \Cref{fig:mse-fixed-gamma}; in proxPR, the rate was independent of $\gamma$ and $\gamma_1$. Note that an initial step size as large as $100$ was allowed, with no explosion, 
confirming the stability of ISGD over SGD.
For the non-strongly convex case, $\gamma=1$ was the fastest among proxRM if $\gamma_1$ was not too small but became the slowest among proxPR, for which slower decays led to faster convergence.

\subsection{Interval estimation and inference}
\begin{table*}[ht!]
\begin{minipage}{.6\linewidth}
\caption{Statistical inference for linear regression model parameters.}
\label{table:LinearModel}
\centering
\tiny
\begin{tabular}{>{\centering}p{0.07\textwidth}>{\raggedleft}p{0.02\textwidth}>{\raggedleft}p{0.03\textwidth}>{\raggedleft}p{0.03\textwidth}>{\raggedleft}p{0.08\textwidth}>{\raggedleft}p{0.03\textwidth}>{\raggedleft}p{0.08\textwidth}>{\raggedleft}p{0.03\textwidth}>{\raggedleft}p{0.08\textwidth}>{\raggedleft}p{0.03\textwidth}>{\raggedleft\arraybackslash}p{0.08\textwidth}}
\toprule
\multicolumn{3}{r}{~}  & \multicolumn{4}{c}{ProxRM}  & \multicolumn{4}{c}{ProxPR}   \\
\cmidrule(lr){4-7}
\cmidrule(lr){8-11}
$\Sigma$ & $\gamma$ & $p$   & cover & MSE & lenCI & covdiff & cover & MSE & lenCI & covdiff  \\ 
\midrule
Identity  & 0.6   & 5   & 95.92    & 4.991e-3 & 0.285    & 5.817e-4 & 95.00    & 1.177e-5 & 0.013    & 1.299e-6  \\
          &       & 20  & 97.54    & 4.424e-3 & 0.298    & 1.640e-3 & 95.01    & 1.270e-5 & 0.014    & 2.657e-6  \\
          &       & 100 & 99.59    & 3.348e-3 & 0.334    & 4.180e-3 & 95.11    & 1.611e-5 & 0.016    & 7.309e-6  \\
          &       & 200 & 99.97    & 2.511e-3 & 0.352    & 5.783e-3 & 95.05    & 1.782e-5 & 0.017    & 1.129e-5  \\
          & 1     & 5   & 94.56    & 5.278e-5 & 0.028    & 3.436e-6 & 94.52    & 1.170e-5 & 0.013    & 1.344e-6  \\
          &       & 20  & 95.14    & 5.245e-5 & 0.028    & 1.050e-5 & 94.70    & 1.151e-5 & 0.013    & 2.361e-6  \\
          &       & 100 & 95.33    & 5.260e-5 & 0.029    & 2.353e-5 & 94.39    & 1.184e-5 & 0.013    & 5.456e-6  \\
          &       & 200 & 95.48    & 5.197e-5 & 0.029    & 3.299e-5 & 94.44    & 1.192e-5 & 0.013    & 7.547e-6  \\ 
\midrule
Toeplitz  & 0.6   & 5   & 95.88    & 4.890e-3 & 0.285    & 6.979e-4 & 94.96    & 1.766e-5 & 0.017    & 2.037e-6  \\
          &       & 20  & 97.40    & 4.439e-3 & 0.298    & 1.617e-3 & 94.96    & 2.082e-5 & 0.018    & 4.247e-6  \\
          &       & 100 & 99.61    & 3.347e-3 & 0.334    & 4.185e-3 & 94.99    & 2.679e-5 & 0.020    & 1.194e-5  \\
          &       & 200 & 99.96    & 2.525e-3 & 0.352    & 5.779e-3 & 95.19    & 2.959e-5 & 0.021    & 1.894e-5  \\
          & 1     & 5   & 94.92    & 5.486e-5 & 0.029    & 4.976e-6 & 94.32    & 1.809e-5 & 0.016    & 1.874e-6  \\
          &       & 20  & 95.05    & 5.518e-5 & 0.029    & 1.066e-5 & 93.95    & 1.985e-5 & 0.017    & 4.767e-6  \\
          &       & 100 & 94.96    & 5.520e-5 & 0.029    & 2.487e-5 & 93.61    & 2.067e-5 & 0.017    & 9.691e-6  \\
          &       & 200 & 95.47    & 5.396e-5 & 0.029    & 3.424e-5 & 93.88    & 2.083e-5 & 0.017    & 1.337e-5  \\ 
\midrule
EquiCorr & 0.6   & 5   & 95.64    & 4.968e-3 & 0.285    & 6.030e-4 & 94.68    & 1.306e-5 & 0.014    & 1.490e-6  \\
          &       & 20  & 97.42    & 4.454e-3 & 0.298    & 1.604e-3 & 94.90    & 1.532e-5 & 0.015    & 3.125e-6  \\
          &       & 100 & 99.60    & 3.394e-3 & 0.334    & 4.159e-3 & 95.20    & 1.985e-5 & 0.018    & 8.990e-6  \\
          &       & 200 & 99.95    & 2.583e-3 & 0.353    & 5.773e-3 & 95.13    & 2.228e-5 & 0.019    & 1.412e-5  \\
          & 1     & 5   & 94.72    & 5.307e-5 & 0.029    & 3.199e-6 & 94.00    & 1.307e-5 & 0.014    & 1.518e-6  \\
          &       & 20  & 95.15    & 5.326e-5 & 0.029    & 1.058e-5 & 94.49    & 1.403e-5 & 0.014    & 2.853e-6  \\
          &       & 100 & 95.19    & 5.337e-5 & 0.029    & 2.391e-5 & 94.25    & 1.480e-5 & 0.015    & 6.827e-6  \\
          &       & 200 & 95.45    & 5.267e-5 & 0.029    & 3.334e-5 & 94.36    & 1.497e-5 & 0.015    & 9.483e-6  \\
          \bottomrule
\end{tabular}
\end{minipage}
\begin{minipage}{.4\linewidth}
\caption{Statistical inference for quantile estimation.}
\label{table:QuantileEstimation}
\centering
\tiny
\begin{tabular}{>{\centering}p{0.07\textwidth}>{\raggedleft}p{0.03\textwidth}>{\raggedleft}p{0.06\textwidth}>{\raggedleft}p{0.03\textwidth}>{\raggedleft}p{0.11\textwidth}>{\raggedleft}p{0.03\textwidth}>{\raggedleft\arraybackslash}p{0.11\textwidth}}
\toprule
Method &
$\gamma$ & $\mu$   & cover & MSE & lenCI & covdiff  \\ 
\midrule
\multirow[t]{6}{*}{ProxRM} &
0.6   & 1e-3 & 95.20    & 1.339e-3 & 0.148    & 1.185e-4  \\
&      & 1e-2 & 95.60    & 1.315e-3 & 0.146    & 9.965e-5  \\
&      & 1e-1 & 95.40    & 1.191e-3 & 0.140    & 8.716e-5  \\
&
1     & 1e-3 & 96.60    & 4.933e-5 & 0.028    & 1.689e-6 \\
&      & 1e-2 & 96.00    & 4.887e-5 & 0.028    & 1.021e-6 \\
&      & 1e-1 & 91.40    & 5.769e-5 & 0.027    & 6.967e-7 \\
\midrule
\multirow[t]{6}{*}{ProxPR} &
0.6   & 1e-3 & 73.80    & 2.626e-5 & 0.016    & 4.363e-6  \\
&      & 1e-2 & 73.40    & 2.587e-5 & 0.015    & 3.262e-6  \\
&      & 1e-1 & 96.40    & 1.266e-5 & 0.015    & 2.755e-6  \\
&      
1     & 1e-3 & 94.60    & 1.603e-5 & 0.016    & 1.246e-6  \\
&      & 1e-2 & 95.60    & 1.589e-5 & 0.015    & 5.692e-7  \\
&      & 1e-1 & 87.40    & 2.872e-5 & 0.015    & 5.506e-7 \\ 
\bottomrule      
\end{tabular}
\end{minipage}
\end{table*}

To evaluate the performance of the plug-in estimators of the asymptotic covariance matrices (\Cref{thm:plugin} and  \Cref{cor:plugin}), we conducted statistical inference on model parameters in linear regression and quantile estimation models using ISGD; see \citet{chen2020statistical,toulis2021proximal}. 
Based on the observations on proxPR in the previous subsection,
we discarded the first tenth of the iterations when calculating $\tilde{H}_n, \hat{I}_n$, and $\bar{\theta}_n$.
Then, we gathered the average rate of the nominal 95\% confidence interval \eqref{eqn:proxRM-sigma}, \eqref{eqn:proxPR-sigma} covering $\theta_*$ for each coordinate (``cover''), mean squared error from $\theta_*$ (``MSE''), and the length of the 95\% confidence interval (``lenCI'') from $B=500$ independent replications. 
We also compared the plug-in estimator of each run with the multi-run estimator \eqref{eqn:empirical_cov} using the Frobenius norm (normalized by dimension $p$); its average is reported (``covdiff''). 

\textbf{Linear regression}. We generated $Z_n = (y_n, \mathbf{x}_n)$ where $y_n = \mathbf{x}_n^T \theta_* + \epsilon_n$, $\mathbb{R}^p \ni \mathbf{x}_n \sim N(0, \Sigma)$, and $\epsilon_n \sim N(0, 1)$, for $\ell(Z_n, \theta) = \frac{1}{2}(y_n - \mathbf{x}_n^T\theta)^2$.
Three different covariance structures were considered: identity ($\Sigma = I_p$), Toeplitz ($\Sigma_{ij} = (0.5)^{|i-j|}$), and equicorrelation ($\Sigma_{ij} = 0.2$ for $i \neq j$, $\Sigma_{ii} = 1$).
We fixed $\iter{\theta}{\star} = (1, \dotsc, 1)^T$ and ran $n=10^5$ iterations of ISGD for $\gamma \in \{0.6, 1.0\}$, $p \in \{5, 20, 100, 200\}$ with  $\theta_0 = 0$ for each type of $\Sigma$. 
\Cref{table:LinearModel} collects the results. The coverage rates generally observed the nominal level, with proxRM showing a slightly better coverage, possibly because proxPR exhibited shorter confidence intervals.
 
\textbf{Quantile estimation}. 
Since the sample function for quantile estimation introduced in \Cref{sec:intro} is nondifferentiable, we instead used a smoothed version in which the $\max(0, \cdot)$ part is replaced by a quadratic function $(4\mu)^{-1}(x+\mu)^2$ on $[-\mu, \mu]$. 
The smoothing parameter $\mu$ makes $\nabla\ell(Z, \theta)$ $(2\mu)^{-1}$-Lipschitz and introduces a bias.
We nevertheless set $\iter{\theta}{\star}$ to be the 99\%-ile
of the standard normal and let $Z \sim N(0, 1)$. 
The $n=10^6$ iterations were started with $\theta_0 = 0$ for each replication, where $\gamma \in \{0.6, 1\}$ and $\mu \in \{10^{-1}, 10^{-2}, 10^{-3}\}$;
we used $\gamma_1 = 250$ when $\gamma = 1$ and $\gamma_1 = 30$ when $\gamma = 0.6$.
\Cref{table:QuantileEstimation} summarizes the results. While proxPR exhibits smaller MSE and covdiff, the coverage rate of proxRM was close to 95\% and there are few cases that proxPR deteriorates. This result may be misleading since the actual minimizer of $L$ is not $\iter{\theta}{\star}$ here, and the length of the confidence interval is smaller for proxPR.
On the other hand, if the genuine goal is to find the quantile, the better coverage of (biased) proxRM might be a blessing.
\section{Conclusion}\label{sec:conclusion}
In both point estimation of the model parameter for the strongly convex case and optimization of non-strongly convex functionals, the behavior of implicit SGD as revealed by our non-asymptotic analysis, either proxRM or proxPR, resembles the explicit SGD when the gradient is uniformly bounded a.s. on a bounded domain \citep[Theorems 2, 5, 7]{bach2011}, which replaces Assumption(s) \ref{enum:noiselevelcond} (and, with strong convexity, \ref{enum:Lipschitz}).
Thus ISGD effectively imposes the latter condition without requiring it, operating under weaker assumptions on the gradient map. 

Like its explicit counterpart, our analysis (and experiments) states that averaging brings to ISGD robustness to the initial step size and lack of strong convexity. On the other hand, averaged ISGD (proxPR) may suffer from slow progress in the early phase of iterations. Burn-in is a viable option. 
It is interesting to note that the plug-in estimators of the asymptotic covariance matrices of proxRM and proxPR, indispensable for valid statistical inference on the model parameter, does not benefit from averaging. Since this paper appears to be the first to propose estimators based only on a single run, investigation of more statistically efficient estimators as well as online ones as in SGD \citep{chen2020statistical} will make a promising avenue of future research.

\newpage
\appendix
\onecolumn
\numberwithin{equation}{section}

\section{Proofs of main results}\label{sec:proofs}
\subsection{Preliminary}
In this section, we collect known facts useful for the subsequent proofs.
\begin{proposition}[\citet{bach2011}, p. 17]\label{prop:normpower}
    \[
    \norm{a + b}^k \leq 2^{k-1}(\norm{a}^k + \norm{b}^k),
    \quad
    k = 1, 2, 3, 4.
    \]
\end{proposition}

\begin{proposition}\label{prop:stepsize}
    \begin{align*}
        \gamma_n \leq \gamma_{n-1} &\leq 2\gamma_n, \quad n \geq 2;
        \\
        \gamma_n^{1/2} - \gamma_{n+1}^{1/2} &\geq \frac{\gamma_1^{1/2}}{4n^{\gamma/2}}, \quad n \geq 3.
    \end{align*}
\end{proposition}

\begin{proposition}
    Let $\phi_\gamma (n) = (n^{1 - \gamma} - 1)/(1 - \gamma)$ if $\gamma > 0$ and $\gamma \neq 1$, and $\phi_\gamma (n) = \log n$ if $\gamma = 1$.
    Then,
    \[
        \sum_{k=1}^n\frac{1}{k^{\gamma}} \leq 1 + \phi_{\gamma}(n) \leq
        \begin{cases}
            \textstyle \frac{n^{1-\gamma}}{1 - \gamma}, & \gamma \in (0, 1), \\
            1 + \log n, & \gamma = 1, \\
            \zeta(\gamma) := \sum_{k=1}^{\infty}\frac{1}{k^{\gamma}} < \infty, & \gamma > 1
        \end{cases}
    \]
    and
    \[
        \frac{1}{2}[\phi_{\gamma}(n) - \phi_{\gamma}(m)]
        \leq \sum_{k=m+1}^n \frac{1}{k^{\gamma}} 
        \leq \phi_{\gamma}(n) - \phi_{\gamma}(m)
        ,
        \quad
        n > m \geq 1
		.
    \]
\end{proposition}
The last inequality is from \citet[p. 13]{bach2011}.

\subsection{Proof of \Cref{thm:aproximation_ISGD}}
\Cref{thm:aproximation_ISGD} can be proved by combining the following intermediate result and \Cref{lem:difference}.
\begin{lemma}[Theorem 3.3 and Corollary 3.3 of \citet{asi2019stochastic}]\label{lem:asibound}
In addition to Assumptions \ref{enum:CCP}, \ref{enum:Lipschitz}, and \ref{enum:noiselevelcond}, 
also assume that a minimizer $\iter{\theta}{\star}$ of $L$ exists (not necessarily unique).
Then, the iterates from the ISGD update \eqref{eq:spp} with learning rate sequence \eqref{enum:learning_rate} satisfies
\begin{align*}
    \E[\norm{\theta_n - \theta_{\star}}^2 | \mathcal{F}_{n-1}]
    &\le
    \norm{\theta_{n-1} - \theta_{\star}}^2 + \sigma^2\gamma_n^2, \quad \gamma \in (0, 1],
    \\
    \E\norm{\theta_n - \theta_{\star}}^2
    &\le
    \norm{\theta_0 - \theta_{\star}}^2 + \sigma^2\sum_{k=1}^{\infty}\gamma_k^2
    = r^2
	< \infty
	,
	\quad
	\gamma \in (1/2, 1].
\end{align*}
for all $n = 1, 2, \dotsc$.
\end{lemma}

\begin{proof}[Proof of \Cref{thm:aproximation_ISGD}]
Note that
\begin{align*}
		\iter{\theta}{n} &= \iter{\theta}{n-1} - \gamma_n\nabla\ell(Z_n,\iter{\theta}{n}) \\
			&= [\iter{\theta}{n-1} - \gamma_n\nabla\ell(Z_n,\iter{\theta}{n-1})] 
				+ [\gamma_n\nabla\ell(Z_n,\iter{\theta}{n-1}) - \gamma_n\nabla\ell(Z_n,\iter{\theta}{n})]
		.
	\end{align*}
	Thus $\iter{R}{n} = \gamma_n[\nabla\ell(Z_n,\iter{\theta}{n-1}) - \nabla\ell(Z_n,\iter{\theta}{n})]$.
	Then,
	\begin{equation}\label{eqn:Rbound}
	    \begin{split}
		\|\iter{R}{n}\| &= \gamma_n\|\nabla\ell(Z_n,\iter{\theta}{n-1}) - \nabla\ell(Z_n,\iter{\theta}{n})]\|
		\\
		&\le \gamma_n \beta(Z_n) \| \iter{\theta}{n-1} - \iter{\theta}{n} \|
				\le \gamma_n^2 \beta(Z_n) \|\nabla\ell(Z_n,\iter{\theta}{n-1})\|
		,
		\end{split}
	\end{equation}
where the first inequality is due to Assumption \ref{enum:Lipschitz}, and the second due to \Cref{lem:difference}.

We show the bounds on $\norm{\iter{R}{n}}$. 
First,
observe from the triangular inequality and Assumption \ref{enum:Lipschitz} that
\begin{equation}\label{eqn:triangle}
	\norm{\nabla\ell(Z_n,\iter{\theta}{n-1})} 
	\leq 
	\beta(Z_n)\norm{\iter{\theta}{n-1}-\iter{\theta}{\star}} + \norm{\nabla\ell(Z_n,\iter{\theta}{\star})}
	.
\end{equation}
Then,
\begin{equation}\label{eqn:Rnbound}
\begin{split}
    \E{[\norm{\iter{R}{n}}|\mathcal{F}_{n-1}]} 
    &\leq \gamma_n^2 \E{[\beta^2(Z_n)]}\norm{\iter{\theta}{n-1} - \iter{\theta}{\star}} \\
    & \quad
    + (1/2)\gamma_n^2\E{[\beta^2(Z_n)]} + (1/2)\gamma_n^2\E{\norm{\nabla\ell(Z_n, \iter{\theta}{\star})}^2}
    \\
    &\leq 
    \gamma_n^2\beta_0^2\norm{\iter{\theta}{n-1} - \iter{\theta}{\star}} + (1/2)\gamma_n^2\beta_0^2 + (1/2)\gamma_n^2\sigma^2
\end{split}
\end{equation}
using $2ab \le a^2 + b^2$ and Assumption \ref{enum:Lipschitz}.
Therefore for $\gamma \in (1/2, 1]$ with which $r < \infty$, 
\begin{align*}
    \E{\norm{\iter{R}{n}}} \leq \gamma_n^2\beta_0^2\E{\norm{\iter{\theta}{n-1} - \iter{\theta}{\star}}} + (1/2)\gamma_n^2(\beta_0^2 + \sigma^2)
    \leq \gamma_n^2[\beta_0^2(r + 1/2) + \sigma^2/2]
    .
\end{align*}
The last inequality is due to \Cref{lem:asibound} and Jensen's inequality.

To establish bounds on $\norm{\iter{R}{n}}^2$,
from inequality \eqref{eqn:triangle},
\begin{align*}
	\norm{\nabla\ell(Z_n,\iter{\theta}{n-1})}^2 &\leq 2\beta^2(Z_n)\norm{\iter{\theta}{n-1}-\iter{\theta}{\star}}^2 + 2\norm{\nabla\ell(Z_n,\iter{\theta}{\star})}^2
\end{align*}
using  $(a + b)^2 \le 2a^2 + 2b^2$.
So
\begin{equation}\label{eqn:gradconditionalbound}
    \E{[\norm{\nabla\ell(Z_n,\iter{\theta}{n-1})}^2 | \mathcal{F}_{n-1}]}
            \leq
    2\beta_0^2\norm{\iter{\theta}{n-1} - \iter{\theta}{\star}}^2 + 2\sigma^2
\end{equation}
from Assumption \ref{enum:noiselevelcond}.
Then, by the definition of $\iter{R}{n}$,
\begin{align*}
    \E{[\norm{\iter{R}{n}}^2 | \mathcal{F}_{n-1}]} 
    &= 
    \gamma_n^2 \E{[\norm{\nabla\ell(Z_n, \iter{\theta}{n}) - \nabla\ell(Z_n, \iter{\theta}{n-1})}^2 | \mathcal{F}_{n-1}]}
    \\
    &=
    \gamma_n^2 \E{[\norm{\nabla\ell(Z_n, \iter{\theta}{n})}^2 | \mathcal{F}_{n-1}]}
    - 2\gamma_n^2 \E{[\nabla\ell(Z_n, \iter{\theta}{n})^T\nabla\ell(Z_n, \iter{\theta}{n-1}) | \mathcal{F}_{n-1}]}
    \\
    &\quad + \gamma_n^2 \E{[\norm{\nabla\ell(Z_n, \iter{\theta}{n-1})}^2 | \mathcal{F}_{n-1}]}
    \\
    &\leq 
    4\gamma_n^2 \E{[\norm{\nabla\ell(Z_n, \iter{\theta}{n-1})}^2 | \mathcal{F}_{n-1}]}
            \leq
    8\gamma_n^2\beta_0^2\norm{\iter{\theta}{n-1} - \iter{\theta}{\star}}^2 + 8 \gamma_n^2\sigma^2
\end{align*}
using the Cauchy-Schwarz inequality and \Cref{lem:difference}.
Then, for $\gamma \in (1/2, 1]$,
\begin{equation}\label{eqn:Rn2_unconditional2}
    \E{\norm{\iter{R}{n}}^2} 
    \leq 
    8\gamma_n^2\beta_0^2\E{\norm{\iter{\theta}{n-1} - \iter{\theta}{\star}}^2} + 8\gamma_n^4\sigma^2
    \leq 
        8\gamma_n^2(\beta_0^2 r^2 + \sigma^2)
\end{equation}
from \Cref{lem:asibound}.
\end{proof}

\subsection{Proof of \Cref{thm:finite_sample}}\label{sec:proofs:RMerror}
We enhance the result by \citet[Lemma 2.2 in the Supplement]{Toulis2017}:
\begin{lemma}\label{lemma:toulis}
    Consider positive sequences $a_n$     such that $\sum_{k=1}^{\infty} a_k = A < \infty$, $b_n \downarrow 0$, and $c_n \downarrow 0$,     and there is $n'$ such that $c_n/b_n < 1$ for all $n > n'$. 
    Let
    \[
        \delta_n = \frac{1}{a_n}\left(\frac{a_{n-1}}{b_{n-1}} - \frac{a_n}{b_n}\right),
        \quad
        \zeta_n = \frac{c_n}{b_{n-1}}\frac{a_{n-1}}{a_n}
    \]
    and suppose $\delta_n \downarrow \delta \geq 0$ and $\zeta_n \downarrow 0$.
    Then, for a nonnegative sequence $y_n$ such that
    \begin{equation}\label{eqn:recursion}
        y_n \le \frac{1 + c_n}{1 + (1 + \delta)b_n}y_{n-1} + a_n
        ,
    \end{equation}
        there holds
    \begin{equation}\label{eqn:finitesample}
        y_n \le 
        K_0 \frac{a_n}{b_n} + Q_1^n y_0 + Q_{n_0+1}^n [(1 + c_1)^{n_0} A + B]
    \end{equation}
    for every $n = 1, 2, \dotsc$, 
    where 
    $n_0$ is an integer such that $\delta_n + \zeta_n < 1  + \delta$ and $c_n < (1 + \delta)b_n$ for all $n \geq n_0$,  
    $K_0 = [1 + (1 + \delta)b_1]/(1 + \delta - \delta_{n_0}-\zeta_{n_0})$,
    $B = K_0 a_{n_0}/b_{n_0}$,
    and
    $Q_i^n=\prod_{j=i}^n(1 + c_j) / (1 + (1 + \delta) b_j)$ if $n \ge i$ and $Q_i^n = 1$ otherwise.
\end{lemma}
\begin{proof}
	See Sect. \ref{sec:technical}.
\end{proof}

\begin{corollary}\label{cor:toulis}
        Let $\alpha > \beta > \gamma$
    where $\alpha > 1$ and $\gamma \in (0, 1]$.     Consider a nonnegative sequence $\{y_n\}$ such that
    \begin{equation}\label{eqn:y_recursion2}
                y_n \leq (1 - \eta n^{-\gamma} + \nu n^{-\beta}) y_{n-1} + a_1 n^{-\alpha}
    \end{equation}
        with $\eta > 0$, $\nu \geq 0$, and $a_1 \geq 0$.
    If $\gamma = 1$, assume that     $\eta > \alpha - \gamma$.
    Then, there holds
    \begin{equation}\label{eqn:y_n_bound}
        y_n \le 
        K_1 n^{-(\alpha - \gamma)} + K_2(n) \exp\left(-\textstyle\frac{1}{2}\log(1+\eta) \phi_{\gamma}(n)\right)
        ( y_0 + D_{n_0} )
        ,
    \end{equation}
    where 
    $$
    \phi_{\gamma}(n) = \begin{cases}
        (n^{1 - \gamma} - 1)/(1 - \gamma), & \gamma \in (0, 1), \\
        \log n, & \gamma = 1,
    \end{cases}
    \quad
    \text{and}
    \quad
    \delta = \begin{cases}
        0, & \gamma \in (0, 1), \\
                1 / (\frac{\eta}{\alpha - \gamma} - 1), & \gamma = 1.
    \end{cases}
    $$
    and
    \begin{subequations}\label{eqn:toulis}
    \begin{align}
        K_1 &= K_0\frac{a_1(1 + \delta)}{\eta},          \label{eqn:toulis:K1}
        \\
        K_2(n) &= \begin{cases}
        \exp\left(\nu(1 + \eta)\sum_{k=1}^{\infty}k^{-\beta} \right) < \infty, & \beta > 1, \\
        \exp\big(\nu(1 + \eta)\phi_{\beta}(n)\big), & \beta \leq 1,
        \end{cases}
        \label{eqn:toulis:K2}
        \\
        D_{n_0} &= (1 + \eta)^{n_0}
        \big([1 + \nu(1+\eta)]^{n_0}A + B\big), \label{eqn:toulis:D_n0}
        \\
        A &= a_1\sum_{k=1}^{\infty}k^{-\alpha} < \infty, 
        \quad
        B =  K_0(1 + \delta)\eta^{-1} a_1 n_0^{-(\alpha - \gamma)}
                . \label{eqn:toulis:AB}
    \end{align}
    \end{subequations}
    The $n_0$ is an integer such that 
    \begin{subequations}\label{eqn:toulis:n0}
    \begin{align}
        n^{\gamma}\big( [n/(n-1)]^{\alpha - \gamma} - 1\big)
        +
                \nu(1 + \eta)
        [n / (n-1)]^{\alpha - \gamma}n^{-\gamma}
        &< \eta         \\
                    \nu(1 + \eta) n^{-\gamma} &< \eta
    \end{align}
    \end{subequations}    
    for all $n \geq n_0$. 
    The constant $K_0$ is given by
    $$
        K_0 = 
        \frac{(1 + \eta)/(1 + \delta)}{1 -
            \left(
            \frac{n_0^{-\gamma}}{\eta}([\frac{n_0}{n_0-1}]^{\alpha - \gamma} - 1) 
            + 
            \frac{\nu(1 + \eta)}{\eta}n_0^{-(\beta-\gamma)}
        [\frac{n_0}{n_0-1}]^{\alpha - \gamma}
        \right)}
        .
    $$
In particular, if $\nu = 0$, then $K_2(n) \equiv 1$.
\end{corollary}
\begin{proof}
	See Sect. \ref{sec:technical}.
\end{proof}

\begin{proof}[Proof of \Cref{thm:finite_sample}]
From \Cref{thm:aproximation_ISGD}, one can obtain
\begin{subequations}\label{eq:finite_sixth}
\begin{align}
\E\norm{\theta_n - \theta_{\star}}^2 =& \E\norm{\theta_{n-1} - \theta_{\star}}^2    
\label{eq:finite_first}
\\
&- 2\gamma_n \E[(\theta_{n-1} - \theta_{\star})^T \nabla \ell(Z_n, \theta_{n-1})]    
\label{eq:finite_second}
\\
& + \gamma_n^2 \E\norm{\nabla \ell(Z_n, \theta_{n-1})}^2
\label{eq:finite_third}
\\
& + 2 \E[(\theta_{n-1} - \theta_{\star})^T R_n]    
\label{eq:finite_fourth}
\\
& - 2 \E[R_n^T \nabla \ell(Z_n, \theta_{n-1})]    
\label{eq:finite_fifth}
\\
& + \E\norm{R_n}^2    
.
\end{align}
\end{subequations}

The second term \eqref{eq:finite_second} is upper bounded as follows. 
\begin{equation}\label{eqn:strongcvxbound}
\begin{split}
    -2 \gamma_n & \E[(\theta_{n-1} - \theta_{\star})^T \nabla \ell(Z_n, \theta_{n-1})] \\
    &= -2 \gamma_n \E\E[(\theta_{n-1} - \theta_{\star})^T \nabla \ell(Z_n, \theta_{n-1}) | \mathcal{F}_{n-1}] \\
    &= -2\gamma_n \E[(\theta_{n-1} - \theta_{\star})^T \E[\nabla \ell(Z_n, \theta_{n-1}) | \mathcal{F}_{n-1}]] \\
    &= -2 \gamma_n \E[(\theta_{n-1} - \theta_{\star})^T \nabla L(\theta_{n-1})] \\
    &= -2 \gamma_n \E[(\theta_{n-1} - \theta_{\star})^T (\nabla L(\theta_{n-1}) - \nabla L(\theta_{\star}))]
    \\
    &\le - 2 \gamma_n \lambda \E \norm{\theta_{n-1} - \theta_\star}^2
    .
\end{split}
\end{equation}
The last inequality is due to Assumption \ref{enum:M_str_conv} and \Cref{rem:strongcvx}; the penultimate inequality is due to $\nabla L(\theta_{\star})=0$.

In order to bound the third term \eqref{eq:finite_third},
from inequality \eqref{eqn:gradconditionalbound} in the proof of \Cref{thm:aproximation_ISGD},
\begin{align*}
    \gamma_n^2 \E\norm{\nabla \ell(Z_n, \theta_{n-1})}^2
    &\leq
    2\gamma_n^2 \beta_0^2 \E{\norm{\iter{\theta}{n-1} - \iter{\theta}{\star}}^2} + 2\gamma_n^2\sigma^2
    \\
    &\leq
    2\gamma_n^2 (\beta_0^2 r^2 + \sigma^2)
    .
\end{align*}
The last line follows from \Cref{lem:asibound}.

Also for the term \eqref{eq:finite_fourth}, from the Cauchy-Schwarz and inequality \eqref{eqn:Rn_conditional}  in \Cref{thm:aproximation_ISGD}  we have
\begin{align*}
    \E{[(\iter{\theta}{n-1} - \iter{\theta}{\star})^T R_n | \mathcal{F}_{n-1} ]}
    & \leq
    \E{[\norm{\iter{\theta}{n-1} - \iter{\theta}{\star}}\norm{\iter{R}{n}} | \mathcal{F}_{n-1}]}
    \\
    &=
    \norm{\iter{\theta}{n-1} - \iter{\theta}{\star}} \E{[\norm{\iter{R}{n}} | \mathcal{F}_{n-1}]}
    \\
    &\leq
    \gamma_n^2\beta_0^2\norm{\iter{\theta}{n-1} - \iter{\theta}{\star}}^2 + \gamma_n^2(\beta_0^2 + \sigma^2)\norm{\iter{\theta}{n-1} - \iter{\theta}{\star}}
    .
\end{align*}
Thus
$$
    2 \E{[(\iter{\theta}{n-1} - \iter{\theta}{\star})^T R_n]}
    \leq
        2\gamma_n^2[\beta_0^2 r^2 + (\beta_0^2 + \sigma^2)r]
$$
using 
\Cref{lem:asibound} and Jensen's inequality.

For the term \eqref{eq:finite_fifth}, we obtain
\begin{align*}
    -&\E{[\iter{R}{n}^T \nabla\ell(Z_n, \iter{\theta}{n-1}) | \mathcal{F}_{n-1} ]}
    =
    -\E{[\gamma_n[\nabla\ell(Z_n, \iter{\theta}{n-1}) - \nabla\ell(Z_n, \iter{\theta}{n})]^T\nabla\ell(Z_n, \iter{\theta}{n-1}) | \mathcal{F}_{n-1} ]}
    \\
    &\quad =
    -\gamma_n \E{[\norm{\nabla\ell(Z_n, \iter{\theta}{n-1})}^2 | \mathcal{F}_{n-1} ]}
      +\gamma_n \E{\nabla\ell(Z_n, \iter{\theta}{n})^T\nabla\ell(Z_n, \iter{\theta}{n-1}) | \mathcal{F}_{n-1}]}
    \\
    &\quad \leq
    -\gamma_n \E{[\norm{\nabla\ell(Z_n, \iter{\theta}{n-1})}^2 | \mathcal{F}_{n-1} ]}
      +\gamma_n \E{\norm{\nabla\ell(Z_n, \iter{\theta}{n})}\norm{\nabla\ell(Z_n, \iter{\theta}{n-1})} | \mathcal{F}_{n-1}]}    
    \\
    &\quad \leq
    -\gamma_n \E{[\norm{\nabla\ell(Z_n, \iter{\theta}{n-1})}^2 | \mathcal{F}_{n-1} ]}
        + \gamma_n \E{[\norm{\nabla\ell(Z_n, \iter{\theta}{n-1})}^2 | \mathcal{F}_{n-1} ]}
    \\
    &\quad = 0
    ,
\end{align*}
where the last inequality is due to \Cref{lem:difference}.
Thus we have $-2\E{[\iter{R}{n}^T\nabla\ell(Z_n, \iter{\theta}{n-1})]} \leq 0$.

The final term \eqref{eq:finite_sixth} is bounded by inequality \eqref{eqn:Rn2_unconditional} in \Cref{thm:aproximation_ISGD}.

Combining these results yields
\begin{align*}
    \E{\norm{\theta_n - \theta_{\star}}^2} 
    &\leq 
                                                                        (1 - 2\lambda\gamma_n) \E\norm{\theta_{n-1} - \theta_{\star}}^2 
    + 
    2[6\beta_0^2 r^2 + 5\sigma^2 + (\beta_0^2 + \sigma^2)r] \gamma_n^2 
    .
\end{align*}
Now we can apply \Cref{cor:toulis} by setting $y_n = \E{\norm{\theta_n - \theta_{\star}}^2} $, $\alpha = 2\gamma$, $a_1 = 2[6\beta_0^2 r^2 + 5\sigma^2 + (\beta_0^2 + \sigma^2)r]$, $\gamma_1^2$, $\eta = 2\lambda\gamma_1 > 0$, and $\nu = 0$. 
This proves the claim.

Below we put the explicit values of the constants:
\begin{align*}
    \delta &= \begin{cases}
        0, & \gamma \in (1/2, 1), \\
        1 / (2\lambda\gamma_1 - 1), & \gamma = 1, 
    \end{cases}
    \\
    K_0 &= 
        \frac{1/(1 + \delta) + \gamma_1}{1 -
            \frac{n_0^{\gamma}}{2\lambda \gamma_1}([\frac{n_0}{n_0-1}]^{\gamma} - 1)},
    \\
    K_1 &= K_0\frac{2[6\beta_0^2 r^2 + 5\sigma^2 + (\beta_0^2 + \sigma^2)r] \gamma_1(1 + \delta)}{2\lambda},
    \\
    K_2(n) &\equiv 1,
    \\
    D_{n_0} &= (1 + 2\lambda\gamma_1)^{n_0}
        (A + B), 
        \\
    A &= 2[6\beta_0^2 r^2 + 5\sigma^2 + (\beta_0^2 + \sigma^2)r] \gamma_1^2\sum_{k=1}^{\infty}k^{-2\gamma} < \infty, 
    \\
    B &= 2K_0[6\beta_0^2 r^2 + 5\sigma^2 + (\beta_0^2 + \sigma^2)r] \gamma_1n_0^{-\gamma}
    .
\end{align*}
\end{proof}

\subsection{Proof of \Cref{thm:stability}}
\begin{proof}[Proof of Theorem \ref{thm:stability}]
Suppose the following holds.
\begin{equation}\label{eqn:grad}
    \E{[\nabla \ell(Z_n, \theta_{n})]} =  \mathcal{H}(\theta_\star)\E{[\theta_{n} - \theta_\star]} 
    + O(\gamma_n^{1/2})
    .
\end{equation}
Taking expectations on the update equation $\theta_n = \theta_{n-1} - \gamma_n \nabla\ell(Z_n, \theta_n)$, equation \eqref{eqn:grad} entails
\[
    \E{[\theta_n - \theta_\star]}
    = 
    \E{[\theta_{n-1} - \theta_\star]} -\gamma_n  \mathcal{H}(\theta_\star) \E{[\theta_n - \theta_\star]} 
    + O(\gamma_n^{1/2})
    ,
\]
or
\begin{align*}
    \E{[\theta_n - \theta_\star]}
    &= [I + \gamma_n\mathcal{H}(\theta_\star)]^{-1} \E{[\theta_{n-1} - \theta_\star]} + [I + \gamma_n\mathcal{H}(\theta_\star)]^{-1} 
    O(\gamma_n^{1/2})
    .
\end{align*}
By recursively applying the above equation, we see
\begin{align*}
    \E{[\theta_n - \theta_\star]}
    &= Q_1^n(\iter{\theta}{0} - \iter{\theta}{\star})
    + Q_1^n O(\sum_{k=1}^n \gamma_k^{1/2})
    = Q_1^n(\iter{\theta}{0} - \iter{\theta}{\star}) 
    + Q_1^n O(n^{1-\gamma/2})
    .
\end{align*}
Note
$\norm{Q_1^n} = \norm{\prod_{i=1}^n [I + \gamma_i\mathcal{H}(\theta_{\star})]^{-1}}
\leq (\prod_{i=1}^n(1 + \lambda\gamma_i))^{-1}$ where $\lambda \leq \lambda_{\min}(\mathcal{H}(\theta_{\star}))$.
From inequality \eqref{eqn:Qbound}, 
$Q_1^n = O(\exp(-\kappa n^{1-\gamma}))$ if $\gamma \in (1/2, 1)$ and $Q_1^n = O(n^{-\kappa})$ if $\gamma = 1$, where $\kappa = \log(1 + \lambda\gamma_1) > 0$. 
In either case, we have $Q_1^n O(n^{1-\gamma/2}) = o(1)$
as desired (when $\gamma = 1$, we are given $\gamma_1 \geq (e^2 - 1)/\lambda$ with which $\kappa \geq 1$).

To see equation \eqref{eqn:grad} holds, 
recall that the objective function $L$ is twice differentiable at $\theta_{\star}$. It follows that
\begin{equation}\label{eqn:gradtaylor}
\begin{split}
    \nabla\ell(Z_n, \theta_n) 
    &= 
    W_n + \mathcal{H}(\theta_{\star})(\theta_n - \theta_{\star}) + o(\norm{\theta_n - \theta_{\star}}),
    \\
    &=
    W_n + \mathcal{H}(\theta_{\star})(\theta_n - \theta_{\star}) + o(\gamma_n^{1/2}),
    \\
    W_n &= \nabla\ell(Z_n, \theta_n) - \nabla L(\theta_n)
    ,
\end{split}
\end{equation}
From \Cref{thm:aproximation_ISGD}, we see $\nabla\ell(Z_n, \theta_n) = \nabla\ell(Z_n, \theta_{n-1}) - \gamma_n^{-1}R_n$ and $\E[R_n] = O(\gamma_n^2)$. In other words,
$W_n = \nabla\ell(Z_n, \theta_{n-1}) - L(\theta_n) - \gamma_n^{-1}R_n$ 
and
\[ 
    \E[W_n] = \nabla L(\theta_{n-1}) - \nabla L(\theta_n) + O(\gamma_n)
    .
\]
The difference of the first two terms are bounded as follows.
\begin{align*}
    \norm{\nabla L(\theta_{n}) - \nabla L(\theta_{n-1})}
    &=
    \norm{\E[\nabla\ell(Z, \theta_{n}) - \nabla\ell(Z, \theta_{n-1})]}
    \\
    &\leq
    \E\norm{\nabla\ell(Z, \theta_{n}) - \nabla\ell(Z, \theta_{n-1})}
    \quad \text{(Jensen's inequality)}
    \\
    &\leq
    \E{\norm{\nabla\ell(Z, \iter{\theta}{n}) - \nabla\ell(Z, \iter{\theta}{\star})}} 
    + \E{\norm{\nabla\ell(Z, \iter{\theta}{n-1}) - \nabla\ell(Z, \iter{\theta}{\star})}}
    \\
    &\leq
    \E{[\beta(Z)\norm{\iter{\theta}{n} - \iter{\theta}{\star}}]}
    + \E{[\beta(Z)\norm{\iter{\theta}{n-1} - \iter{\theta}{\star}}]}
        \quad \text{(Assumption \ref{enum:Lipschitz})}
    \\
    &\leq
    (\E{[\beta^2(Z)]})^{1/2}(\E{\norm{\iter{\theta}{n} - \iter{\theta}{\star}}}^2)^{1/2}
    \\
    &\quad
    + 
    (\E{[\beta^2(Z)]})^{1/2}(\E{\norm{\iter{\theta}{n-1} - \iter{\theta}{\star}}}^2)^{1/2}
    \quad \text{(Cauchy-Schwarz)}
    \\
    &= O(\gamma_n^{1/2})
    ,
\end{align*}
where the last inequality is due to
Assumption \ref{enum:Lipschitz} and
\Cref{thm:finite_sample}.
Hence $\E[W_n] = O(\gamma_n^{1/2})$. 
Finally, by taking expectations on both sides of equation \eqref{eqn:gradtaylor}, equation \eqref{eqn:grad} follows.
\end{proof}

\subsection{Proof of \Cref{thm:asymp_normality}}
The proof utilizes Fabian's central limit theorem for stochastic approximation \citet[Theorem 2.2]{Fabian1968} to show the asymptotic normality \eqref{eqn:CLT}.
\begin{lemma}[Fabian]
\label{lemma:Fabian}
Let $U_n$, $V_n$, $T_n \in \mathbb{R}^p$ and $\Phi_n$, $\Gamma_n \in \mathbb{R}^{p \times p}$ satisfy the following equation
\begin{equation*}
U_n = (I - n^{-\alpha}\Gamma_n)U_{n-1} + n^{-(\alpha+\beta)/2}\Phi_n V_n + n^{-\alpha-\beta/2}T_n
\end{equation*}
for $\alpha \in (0, 1]$ and $\beta \ge 0$,
where
$\Gamma_n \rightarrow \Gamma \succ 0$, $\Phi_n \rightarrow \Phi$, $T_n \rightarrow T$ 
or 
$\E\norm{T_n - T} \rightarrow 0$, 
$\E[V_n | \mathcal{F}_{n-1}] = 0$, 
$C > \norm{\E[V_n V_n^T - \Sigma | \mathcal{F}_{n-1}}] \rightarrow 0$ for some $C$ and
$\mathcal{F}_n$ a non-decreasing sequence of $\sigma$-fields 
such that $\Gamma_n$, $\Phi_{n}$, and $V_{n}$ are $\mathcal{F}_n$-measurable. 
Suppose
$\sigma^2_{j,r} = \E[I_{\norm{V_j}^2 \ge rj^{\alpha}}\norm{V_j}^2]$, 
and either 
$\lim_{j \rightarrow \infty}\sigma^2_{j,r} = 0$ 
for every $r > 0$ 
or $\alpha = 1$ and
$\lim_{n \rightarrow \infty} n^{-1} \sum_{j=1}^n \sigma^2_{j,r} = 0$ 
for every $r > 0$. 
Let
$\beta_{+} = \beta$ if $\alpha = 1$, $\beta_{+} = 0$ if $\alpha \neq 1$
and assume $\Gamma \succ (\beta_{+}/2)I$.
Then the asymptotic distribution of $n^{\beta/2}U_n$ is normal with mean $(\Gamma - (\beta_{+}/2)I)^{-1}T$ and covariance matrix 
$\mathcal{L}_{2\Gamma - \beta_{+}I}^{-1}(\Phi\Sigma\Phi^T)$,
where 
$\mathcal{L}_P: X \mapsto (1/2)(PX + XP)$ is the Lyapunov linear map.
\end{lemma}
If $P \succ 0$, the inverse linear map $\mathcal{L}_P^{-1}$ is well-defined; if furthermore $C \succ 0$, then $\mathcal{L}_P^{-1}(C)\succ 0$.
\begin{proof}[Proof of \Cref{thm:asymp_normality}]
Using \Cref{thm:aproximation_ISGD}, the implicit update equation becomes
\begin{equation}\label{eqn:normalapprox}
\begin{split}
    \theta_n &= \theta_{n-1} - \gamma_n\nabla\ell(Z_n, \theta_n)
    \\
             &= \theta_{n-1} - \gamma_n\nabla\ell(Z_n, \theta_{n-1}) + R_n
    \\
             &= \theta_{n-1} - \gamma_n[\nabla L(\theta_{n-1}) + \nabla\ell(Z_n, \theta_{n-1}) - \nabla L(\theta_{n-1})] + O(\gamma_n^2)
    \\
             &= \theta_{n-1} - \gamma_n\nabla L(\theta_{n-1}) + \gamma_n\varepsilon_n + O(\gamma_n^2)
             ,
\end{split}             
\end{equation}
where $\varepsilon_n = \nabla L(\theta_{n-1}) - \nabla\ell(Z_n, \theta_{n-1})$. 
Since $L$ is twice differentiable at $\theta_{\star}$,  we further have
\begin{equation}\label{eqn:fabianrecursion}
\begin{split}
    \theta_n - \theta_{\star} 
             &= \theta_{n-1} - \theta_{\star} - \gamma_n\mathcal{H}(\theta_{\star})(\theta_{n-1} - \theta_{\star}) + \gamma_n o(\norm{\theta_{n-1}-\theta_{\star}}) + \gamma_n\varepsilon_n + O(\gamma_n^2)
    \\
             &= \theta_{n-1} - \theta_{\star} - \gamma_n\mathcal{H}(\theta_{\star})(\theta_{n-1} - \theta_{\star}) +  \gamma_n o(\gamma_{n-1}^{1/2}) + \gamma_n\varepsilon_n + O(\gamma_n^2)
    \\
             &= \theta_{n-1} - \theta_{\star} - \gamma_n\mathcal{H}(\theta_{\star})(\theta_{n-1} - \theta_{\star}) +  o(\gamma_{n}^{3/2}) + \gamma_n\varepsilon_n + O(\gamma_n^2)
    \\
             &= [I - \gamma_n\mathcal{H}(\theta_{\star})](\theta_{n-1} - \theta_{\star}) + \gamma_n\varepsilon_n + o(\gamma_n^{3/2})
    ,         
\end{split}
\end{equation}
where the second line is due to \Cref{thm:finite_sample} and $\theta_{n-1} \to \theta_{\star}$ a.s.
For the third line, recall that $\gamma_n = \gamma_1 n^{-\gamma}$ \eqref{enum:learning_rate}.

In order to apply \Cref{lemma:Fabian}, observe that $\E[\varepsilon_n | \mathcal{F}_{n-1}] = 0$ and
\begin{align*}
    \E[\varepsilon_n\varepsilon_n^T | \mathcal{F}_{n-1}]
    &= \E[(\nabla L(\theta_{n-1}) - \nabla\ell(Z_n, \theta_{n-1}))(\nabla L(\theta_{n-1}) - \nabla\ell(Z_n, \theta_{n-1}))^T | \mathcal{F}_{n-1}]
    \\
    &= \E[(\nabla L(\theta_{n-1}) - \nabla\ell(Z, \theta_{n-1}))(\nabla L(\theta_{n-1}) - \nabla\ell(Z, \theta_{n-1}))^T ]
    \\
    &= \mathcal{I}(\theta_{n-1})
\end{align*}
since $\theta_{n-1} \in \mathcal{F}_{n-1}$. 
To meet the conditions for \Cref{lemma:Fabian}, it suffices to show that $\mathcal{I}$ is continuous at $\iter{\theta}{\star}$.
Consider a non-random convergent sequence $\{\iter{\vartheta}{n}\}$ such that $\iter{\vartheta}{n} \to \iter{\theta}{\star}$. Fix $\epsilon > 0$. Then there exists $n_0$ such that for all $n \geq n_0$, $\norm{\iter{\vartheta}{n} - \iter{\theta}{\star}} \leq \epsilon$. Then,
\begin{align*}
    \norm{\nabla\ell(Z, \iter{\vartheta}{n})\nabla\ell(Z, \iter{\vartheta}{n})^T}
    &\leq
    \norm{\nabla\ell(Z, \iter{\vartheta}{n})}^2
    \\
    &\leq
    2\beta^2(Z)\norm{\iter{\vartheta}{n} - \iter{\vartheta}{\star}}^2 + 2\norm{\nabla\ell(Z, \theta_{\star})}^2
    \\
    &\leq
    2\beta^2(Z)\epsilon^2 + 2\norm{\nabla\ell(Z, \theta_{\star})}^2
\end{align*}
for all $n \geq n_0$.
The last line is integrable due to Assumptions \ref{enum:Lipschitz} and \ref{enum:noiselevelcond}.
Therefore, by the dominated convergence theorem, $\mathcal{I}(\iter{\vartheta}{n}) = \E{[\nabla\ell(Z, \iter{\vartheta}{n})\nabla\ell(Z, \iter{\vartheta}{n})^T]} - \nabla L(\iter{\vartheta}{n})L(\iter{\vartheta}{n})^T < \infty$ and $\mathcal{I}(\iter{\vartheta}{n}) \to \mathcal{I}(\iter{\theta}{\star})$ and $\mathcal{I}$ is continuous at $\iter{\theta}{\star}$.

Letting $U_n = \theta_n - \theta_{\star}$, $V_n = \varepsilon_n$, $T_n = o(1)$, $\Phi_n = \Phi = \gamma_1 I$,  $\Gamma_n = \Gamma = \gamma_1\mathcal{H}(\theta_{\star})$, $T=0$, $\Sigma=\mathcal{I}(\theta_{\star})$, and $\alpha=\beta=\gamma$ in \Cref{lemma:Fabian} results in the desired asymptotic normality.
\end{proof}
\begin{remark}\label{rem:normalapprox}
    The approximation of ISGD to SGD in \Cref{thm:aproximation_ISGD} is crucial since otherwise $\varepsilon_n$ would equal to $\nabla L(\theta_n) - \nabla\ell(Z_n, \theta_n)$. Since $\theta_n \notin \mathcal{F}_{n-1}$, we do not have $\E[\varepsilon_n | \mathcal{F}_{n-1}] = 0$. That is, $\varepsilon_n$ is not a martingale difference sequence and it is difficult to see if $\E[\varepsilon_n\varepsilon_n^T|\mathcal{F}_{n-1}]$ would converge to a known quantity.
    Toulis et al. \citep{Toulis2017,Toulis2017supp} instead employ $\nabla\ell(Z_n, \theta_{\star})$ in place of $\varepsilon_n$, but assume $\nabla^2\ell(Z_n, \theta_{\star})$ converges to $\mathcal{H}(\theta_{\star})=\nabla^2 L(\iter{\theta}{\star})$ almost surely, which rarely holds in general. 
\end{remark}

\subsection{Proof of \Cref{thm:plugin}}
The following result can be deduced from the proof of \citet[Lemma 4.1]{chen2020statistical}.
\begin{lemma}\label{lem:consistency}
    Suppose Assumptions \ref{enum:CCP}--\ref{enum:noiselevelcond},
\ref{enum:third_derivative}, and \ref{enum:fourthmoment}--\ref{enum:hessian} hold.
    Then,
    \[
        \E\norm{\hat{H}_n - \mathcal{H}(\iter{\theta}{\star})} = O(\gamma_n^{1/2}),
        \quad
        \text{and}
        \quad
        \E\norm{\hat{I}_n - \mathcal{I}(\iter{\theta}{\star})} = O(\gamma_n^{1/2})
        .
    \]
\end{lemma}
A key in the proof of \citet[Lemma 4.1]{chen2020statistical} is \citet[Lemma 3.2]{chen2020statistical} showing that $\E\norm{\iter{\theta}{n} - \iter{\theta}{\star}} = O(n^{-\gamma/2})=O(\gamma_n^{1/2})$ in (explicit) SGD, which can be replaced by \Cref{thm:finite_sample} in implicit SGD.

\begin{proof}[Proof of \Cref{thm:plugin}]
Let $B = \gamma_1\mathcal{H}(\iter{\theta}{\star}) - (\beta_{+}/2)I$,
$\tilde{B}_n = \gamma_1\tilde{H}_n - (\beta_{+}/2)I$, and $\hat{B}_n = \gamma_1\hat{H}_n - (\beta_{+}/2)I$, where $\beta_{+}=1$ if $\gamma=1$ and $\beta_{+}=0$ if $\gamma \in (0.5, 1)$.  By construction, $\lambda_{\min}(\tilde{B}_n) \geq \gamma_1\delta - \beta_{+}/2 > 0$ and $\lambda_{\min}(B) \geq \gamma_1\lambda_{\min}(\mathcal{H}(\iter{\theta}{\star})) - \beta_{+}/2 > 0$.

Recall that $\vec\big(\mathcal{L}_{2B_n}^{-1}(\hat{I}_n)\big) = (I\otimes B_n + B_n\otimes I)^{-1}\vec(\hat{I}_n)$ and $\vec\big(\mathcal{L}_{2B}^{-1}(\mathcal{I}(\iter{\theta}{\star})\big) = (I\otimes B + B\otimes I)^{-1}\vec(\mathcal{I}(\iter{\theta}{\star})$. 
It suffices to show that
\begin{equation}\label{eqn:estimrate}
    \E\norm{\vec\big(\mathcal{L}_{2B_n}^{-1}(\hat{I}_n)\big) - \vec\big(\mathcal{L}_{2B}^{-1}(\mathcal{I}(\iter{\theta}{\star})\big)}
    = O(\gamma_n^{1/2})
    .
\end{equation}

To see this, let
\begin{align*}
    E_B &= (I\otimes \tilde{B}_n + \tilde{B}_n\otimes I) - (I\otimes B + B\otimes I)
    ,
    \quad
    E_I = \hat{I}_n - \mathcal{I}(\iter{\theta}{\star}),
    \\
    F_B &= (I\otimes \tilde{B}_n + \tilde{B}_n\otimes I)^{-1} - (I\otimes B + B\otimes I)^{-1}
    .
\end{align*}
Then,
\begin{align*}
    \vec&\big(\mathcal{L}_{2B_n}^{-1}(\hat{I}_n)\big) - \vec\big(\mathcal{L}_{2B}^{-1}(\mathcal{I}(\iter{\theta}{\star})\big)
    =
    \\
    &
    [(I\otimes B + B\otimes I)^{-1} + F_B](\vec(\mathcal{I}(\iter{\theta}{\star}) + E_I)
    - (I\otimes B + B\otimes I)^{-1}\vec(\mathcal{I}(\iter{\theta}{\star})
    \\
    &=
    (I\otimes B + B\otimes I)^{-1} E_I
    + F_B E_I
    + F_B \vec(\mathcal{I}(\iter{\theta}{\star}))
    .
\end{align*}
Since the eigenvalues of $I\otimes A + A \otimes I$ consist of $\lambda_i(A) + \lambda_j(A)$ for $i, j=1, \dotsc, p$ if $A$ is a $p\times p$ symmetric matrix,
\[
    \norm{F_B} \le \norm{(I\otimes \tilde{B}_n + \tilde{B}_n\otimes I)^{-1}} + \norm{(I\otimes B + B\otimes I)^{-1}}
    \le 
    \frac{1}{2\gamma_1\delta - \beta_{+}} + \frac{1}{2\lambda_{\min}(B)}
    .
\]
Therefore, by \Cref{lem:consistency},
\begin{equation}\label{eqn:estimbound1}
    \E\norm{(I\otimes B + B\otimes I)^{-1} E_I
    + F_B E_I}
    \le 
    \left(\frac{1}{2\gamma_1\delta - \beta_{+}} + \frac{1}{\lambda_{\min}(B)}\right)
    \E\norm{E_I}
    = O(\gamma_n^{1/2})
    .
\end{equation}

In order to bound $F_B\vec(\mathcal{I}(\iter{\theta}{\star}))$, recall from \citet[Lemma C.1]{chen2020statistical} that
\[
    \norm{F_B}
    \leq
    \norm{E_B}
    \norm{(I\otimes B + B\otimes I)^{-1}}^2
    \leq
    \frac{1}{4\lambda_{\min}^2(B)}\norm{E_B}
\]
under the event $\norm{(I\otimes B + B\otimes I)^{-1}E_B} < 1/2$. 
Thus
\begin{align*}
    \E\norm{F_B} 
    &\leq
    \frac{1}{4\lambda_{\min}^2(B)}\E\norm{E_B} P(\norm{(I\otimes B + B\otimes I)^{-1}E_B} < 1/2) 
    \\
    &\quad + 
    \Big(\frac{1}{2\gamma_1\delta - \beta_{+}}
    + \frac{1}{2\lambda_{\min}(B)}\Big)
    P(\norm{(I\otimes B + B\otimes I)^{-1}E_B} \geq 1/2) 
    \\
    &\leq
    \frac{1}{4\lambda_{\min}^2(B)}\E\norm{E_B}
    +
    \Big(\frac{1}{2\gamma_1\delta - \beta_{+}}
    + \frac{1}{2\lambda_{\min}(B)}\Big)
    \frac{1}{\lambda_{\min}(B)}\E\norm{E_B}
    ,
\end{align*}
where the last line is due to Markov's inequality
\[
    P(\norm{(I\otimes B + B\otimes I)^{-1}E_B} \geq 1/2) 
    \leq
    2\norm{(I\otimes B + B\otimes I)^{-1}}\E\norm{E_B}
    .
\]

It remains to bound $\E\norm{E_B}$.
If $\hat{H}_n \succeq \delta I$, then $\tilde{H}_n = \hat{H}_n$. Otherwise, $\lambda_{\min}(\hat{H}_n) < \delta$ and
\begin{align*}
    \norm{E_B} &= \norm{(I\otimes \tilde{B}_n + \tilde{B}_n\otimes I) - (I\otimes B + B\otimes I)}
    \\
    &\leq
    \norm{(I\otimes \tilde{B}_n + \tilde{B}_n\otimes I) - (I\otimes \hat{B}_n + \hat{B}_n\otimes I)}
    \\
    &\quad\quad
    +
    \norm{(I\otimes \hat{B}_n + \hat{B}_n\otimes I) - (I\otimes B + B\otimes I)}
    \\
    &\leq
    2\delta 
        +
    \norm{(I\otimes \hat{B}_n + \hat{B}_n\otimes I) - (I\otimes B + B\otimes I)}
    .
\end{align*}
Thus, by using \Cref{lem:consistency},
\begin{align*}
    \E\norm{E_B}
    &\leq
    \E\norm{(I\otimes \hat{B}_n + \hat{B}_n\otimes I) - (I\otimes B + B\otimes I)}
    +
    2\delta P(\lambda_{\min}(\hat{H}_n) < \delta)
    \\
    &\leq
    \E\norm{I\otimes (\hat{B}_n - B) + (\hat{B}_n - B) \otimes I}
    \\
    &\quad\quad +
    2\delta P\left(\norm{\hat{H}_n - \mathcal{H}(\iter{\theta}{\star})} \geq \lambda_{\min}(\mathcal{H}(\iter{\theta}{\star})) - \delta\right)
    \\
    &\leq
    2\E\norm{\hat{H}_n - \mathcal{H}(\iter{\theta}{\star})}
    +
    \frac{2\delta}{\lambda_{\min}(\mathcal{H}(\iter{\theta}{\star})) - \delta}
    \E\norm{\hat{H}_n - \mathcal{H}(\iter{\theta}{\star})}
    ,
\end{align*}
where the second line is due to Weyl's inequalty
\[
    \lambda_{\min}(\mathcal{H}(\iter{\theta}{\star}))
    \leq
    \lambda_{\max}(\mathcal{H}(\iter{\theta}{\star}) - \hat{H}_n)
    +
    \lambda_{\min}(\hat{H}_n)
    \leq
    \norm{\mathcal{H}(\iter{\theta}{\star}) - \hat{H}_n}
    + \delta
\]
and the third line is Markov's inequality.
Hence,
\begin{equation}\label{eqn:estimbound2}
\begin{split}
    \E\norm{F_B \vec\mathcal{I}(\iter{\theta}{\star})}
    &\leq
    \norm{\mathcal{I}(\iter{\theta}{\star})}_F
    \left(
    \frac{1}{4\lambda_{\min}^2(B)}
    +
    \Big(\frac{1}{2\gamma_1\delta - \beta_{+}}
    + \frac{1}{2\lambda_{\min}(B)}\Big)
    \frac{1}{\lambda_{\min}(B)}
    \right) \times
    \\
    & \qquad
    \Big(
     2 + \frac{2\delta}{\lambda_{\min}(\mathcal{H}(\iter{\theta}{\star})) - \delta}
    \Big)
    \E\norm{\hat{H}_n - \mathcal{H}(\iter{\theta}{\star})}
    = O(\gamma_n^{1/2})
\end{split}
\end{equation}
by \Cref{lem:consistency}.

Combining inequalities \eqref{eqn:estimbound1} and \eqref{eqn:estimbound2}, we obtain inequality \eqref{eqn:estimrate}.
\end{proof}

\subsection{Proof of \Cref{thm:polyakruppert}}\label{sec:proofs:PRerror}
\begin{lemma}\label{lem:l4bound}
Under Assumptions \ref{enum:CCP}, \ref{enum:Lipschitz2}, \ref{enum:M_str_conv},  \ref{enum:noiselevelcond2}, \ref{enum:Hessian}, and \ref{enum:third_derivative}, if $\gamma \in (1/3, 1]$, it follows
\begin{align*}
    \E{\norm{\iter{\theta}{n} - \iter{\theta}{\star}}^4}
    & \leq
    \tilde{K}_1 n^{-2\gamma}
    + \exp\Big(
    \nu(1 + \lambda\gamma_1/2)\phi_{\frac{5}{3}\gamma}(n) 
    - \textstyle\frac{1}{2}\log(1 + \lambda\gamma_1/2)\phi_{\gamma}(n)\Big)
    \\
    & \quad    
    \times
    (\norm{\iter{\theta}{0} - \iter{\theta}{\star}}^4 
    + \textstyle\frac{8(\beta_0^2 + \sigma^2)}{\lambda}\gamma_1\norm{\iter{\theta}{0} - \iter{\theta}{\star}}^3
    + \tilde{D}_{\tilde{n}_0})
\end{align*}
    where     \begin{align*}
        \nu &= \textstyle 
        \frac{6(\beta_0^2 + \sigma^2)}{\lambda}\beta_0^{2/3}\gamma_1^{5/3} + 14\beta_0^2\gamma_1^{2} + 16\beta_0^3\gamma_1^{3} + 8\beta_0^4\gamma_1^{4}
                        + \textstyle
        2K_2
        [\textstyle (10\sigma^2 + \frac{32\sigma(\beta_0^2 + \sigma^2)}{\lambda})\gamma_1^{2} + \frac{16(\beta_0^2 + \sigma^2)^2}{\lambda}\gamma_1^{3} + 8\gamma_1^{4}]
        ,
    \end{align*}
    and
    \[
        K_2 = (K_1 + \norm{\theta_0 - \theta_{\star}}^2 + D_{n_0}) / \gamma_1
    \]
    with $K_1$ and $D_{n_0}$ as defined in \Cref{thm:finite_sample}.
    Here, $\tilde{n}_0$ is an integer such that 
    \begin{equation}\label{eqn:n0tilde}
        n^{\gamma}\big( [n/(n-1)]^{2\gamma} - 1\big)
        +
                \nu(1 + \eta)
        [n / (n-1)]^{2\gamma}n^{-\gamma}
        < \eta         ~
        \text{and}
        ~
        \nu(1 + \lambda\gamma_1/2) n^{-\gamma} < \eta
    \end{equation}
    for all $n \geq \tilde{n}_0$,
    and
    \begin{subequations}\label{eqn:tildeconstants}
    \begin{align}
        \tilde{K}_1 &= \frac{1 + \lambda\gamma_1/2}{1 -
            \left(
            \frac{\tilde{n}_0^{-\gamma}}{\eta}([\frac{\tilde{n}_0}{\tilde{n}_0-1}]^{2\gamma} - 1) 
            + 
            \frac{\nu(1 + \lambda\gamma_1/2)}{\lambda\gamma_1/2}\tilde{n}_0^{-2\gamma/3}
        [\frac{\tilde{n}_0}{\tilde{n}_0-1}]^{2\gamma}
        \right)}
        \frac{C_1\gamma_1^3}{\lambda\gamma_1/2},          \\
        C_1 &= \textstyle
        \frac{2(\beta_0^2 + \sigma^2)}{\lambda}\beta_0^2
        + (16\sigma^4 + \frac{56(\beta_0^2 + \sigma^2)}{\lambda}\sigma^3)\gamma_1   
        \\
        \tilde{D}_{\tilde{n}_0} &= (1 + \lambda\gamma_1/2)^{\tilde{n}_0}
        \big([1 + \nu(1 + \lambda\gamma_1/2)]^{\tilde{n}_0}\tilde{A} + \tilde{B}\big), 
        \\
        \tilde{A} &= C_1\gamma_1^3\sum_{k=1}^{\infty}k^{-3\gamma} < \infty, 
        \\
        \tilde{B} &=  \frac{1 + \lambda\gamma_1/2}{1 -
            \left(
            \frac{\tilde{n}_0^{-\gamma}}{\eta}([\frac{\tilde{n}_0}{\tilde{n}_0-1}]^{2\gamma} - 1) 
            + 
            \frac{\nu(1 + \lambda\gamma_1/2)}{\lambda\gamma_1/2}\tilde{n}_0^{-2\gamma/3}
        [\frac{n_0}{\tilde{n}_0-1}]^{2\gamma}
        \right)}\frac{C_1\gamma_1^3}{\lambda\gamma_1/2} \tilde{n}_0^{-2\gamma}
        .
    \end{align}
    \end{subequations}    
\end{lemma}

\begin{proof}[Proof of \Cref{thm:polyakruppert}]
We follow the decomposition by \citet[section C]{bach2011}
for explicit SGD.
The only difference is that
$$
    \nabla\ell(Z_n, \iter{\theta}{n}) = \frac{1}{\gamma_n}(\iter{\theta}{n-1} - \iter{\theta}{n})
$$
instead of $\nabla\ell(Z_n, \iter{\theta}{n-1}) = \frac{1}{\gamma_n}(\iter{\theta}{n-1} - \iter{\theta}{n})$,
which leads to 
\begin{align*}
    \norm{\nabla\ell(Z_n, \iter{\theta}{n-1})}
    &\leq
    \norm{\nabla\ell(Z_n, \iter{\theta}{n-1}) - \nabla\ell(Z_n, \iter{\theta}{n})}
    + \norm{\nabla\ell(Z_n, \iter{\theta}{n})}
    \\
    &\leq
    \beta_0\norm{\iter{\theta}{n-1} - \iter{\theta}{n}} + \frac{1}{\gamma_n}\norm{\iter{\theta}{n-1} - \iter{\theta}{n}}
    =
    \frac{\beta_0\gamma_n + 1}{\gamma_n}\norm{\iter{\theta}{n-1} - \iter{\theta}{n}}
    \\
    &\leq
    \frac{\beta_0\gamma_1 + 1}{\gamma_n}\norm{\iter{\theta}{n-1} - \iter{\theta}{n}}
\end{align*}
and
\begin{align*}
    \norm{\frac{1}{n}\sum_{k=1}^n\nabla\ell(Z_k, \iter{\theta}{k-1})}
    &\leq
    \frac{\beta_0\gamma_1 + 1}{n}\sum_{k=1}^{n-1}|\gamma_{k+1}^{-1} - \gamma_k^{-1}|\norm{\iter{\theta}{k} - \iter{\theta}{\star}}
    \\
    \quad
    &+
    \frac{\beta_0\gamma_1 + 1}{n\gamma_n}\norm{\iter{\theta}{n}-\iter{\theta}{\star}}
    +
    \frac{\beta_0\gamma_1 + 1}{n\gamma_1}\norm{\iter{\theta}{0} - \iter{\theta}{\star}}
    .
\end{align*}
This in turn yields
\begin{align*}
    (\E{\norm{\iter{\bar{\theta}}{n}-\iter{\theta}{\star}}^2})^{1/2}
    &\leq
    \frac{1}{\sqrt{n}}\left[\tr{(\mathcal{H}(\iter{\theta}{\star})^{-1}\mathcal{I}(\iter{\theta}{\star})\mathcal{H}(\iter{\theta}{\star})^{-1})}\right]^{1/2}
    \\
    &\quad
    +
    \frac{\beta_0\gamma_1 + 1}{\lambda^{1/2} n}\sum_{k=1}^{n-1}|\gamma_{k+1}^{-1} - \gamma_k^{-1}|(\E{\norm{\iter{\theta}{k} - \iter{\theta}{\star}}^2})^{1/2}
    \\
    &\quad
    + \frac{\beta_0\gamma_1 + 1}{\lambda^{1/2} n \gamma_n}(\E{\norm{\iter{\theta}{n} - \iter{\theta}{\star}}^2})^{1/2}
    + \frac{\beta_0\gamma_1 + 1}{\lambda^{1/2} \gamma_1 n}\norm{\iter{\theta}{0}-\iter{\theta}{\star}}
    \\
    &\quad
    +
    \frac{M}{2\lambda^{1/2}n}\sum_{k=1}^n(\E{\norm{\iter{\theta}{k} - \iter{\theta}{\star}}^4})^{1/2}
    + 
    \frac{2\beta_0}{\lambda^{1/2} n}(\sum_{k=0}^{n-1}\E{\norm{\iter{\theta}{k} - \iter{\theta}{\star}}^2})^{1/2}
    ,
\end{align*}
which corresponds to inequality (26) of \cite{bach2011}.
The above bound can be simplified to
\begin{subequations}\label{eqn:PRbound}
\begin{align}
    (\E{\norm{\iter{\bar{\theta}}{n}-\iter{\theta}{\star}}^2})^{1/2}
    &\leq
    \frac{1}{\sqrt{n}}\left[\tr{(\mathcal{H}(\iter{\theta}{\star})^{-1}\mathcal{I}(\iter{\theta}{\star})\mathcal{H}(\iter{\theta}{\star})^{-1})}\right]^{1/2}
    + 
    \frac{1/\gamma_1 + 3\beta_0}{\lambda^{1/2}  n}\norm{\iter{\theta}{0}-\iter{\theta}{\star}}  
    \nonumber
    \\
    &\quad 
    + 
    \frac{\beta_0\gamma_1 + 1}{\lambda^{1/2} n \gamma_n}(\E{\norm{\iter{\theta}{n} - \iter{\theta}{\star}}^2})^{1/2}
    \label{eqn:PRbound1}
    \\
    &\quad
    +
    \frac{2\beta_0}{\lambda^{1/2} n}(\sum_{k=1}^{n}\E{\norm{\iter{\theta}{k} - \iter{\theta}{\star}}^2})^{1/2}
    \label{eqn:PRbound2}
    \\
    &\quad
    +
    \frac{\beta_0\gamma_1 + 1}{\lambda^{1/2} n}\sum_{k=1}^{n-1}|\gamma_{k+1}^{-1} - \gamma_k^{-1}|(\E{\norm{\iter{\theta}{k} - \iter{\theta}{\star}}^2})^{1/2}
    \label{eqn:PRbound3}
    \\
    &\quad
    +
    \frac{M}{2\lambda^{1/2}n}\sum_{k=1}^n(\E{\norm{\iter{\theta}{k} - \iter{\theta}{\star}}^4})^{1/2}     
    \label{eqn:PRbound4}
\end{align}
\end{subequations}

The terms \eqref{eqn:PRbound1}--\eqref{eqn:PRbound3} can be further bounded using \Cref{thm:finite_sample} and Minkowski's inequality: \begin{align*}
    \text{\eqref{eqn:PRbound1}}
    &\leq
    \frac{(\beta_0\gamma_1 + 1)K_1^{1/2}}{\lambda^{1/2}\gamma_1 n^{1-\gamma/2}}
    \\
    &\quad
    +
    \frac{\beta_0\gamma_1 + 1}{\lambda^{1/2} \gamma_1 n^{1-\gamma}}
    \exp\left(-\textstyle\frac{1}{4}\log(1+2\lambda\gamma_1) \phi_{\gamma}(n)\right)
        ( \norm{\theta_0 - \theta_{\star}}^2 + D_{n_0} )^{1/2}
    \\
    &\leq 
    \frac{\beta_0\gamma_1 + 1}{\lambda^{1/2}\gamma_1}\frac{K_1^{1/2}}{ n^{1-\gamma/2}}    
    +
    \frac{\beta_0\gamma_1 + 1}{\lambda^{1/2} \gamma_1}
    \exp\left(-\textstyle\frac{1}{4}\log(1+2\lambda\gamma_1) \phi_{\gamma}(n)\right)
    ( \norm{\theta_0 - \theta_{\star}}^2 + D_{n_0} )^{1/2}
    ,
\\
    \text{\eqref{eqn:PRbound2}}
    & \leq
    \frac{2\beta_0}{\lambda^{1/2}n}
    \Bigg[
    K_1\sum_{k=1}^n k^{-\gamma}
    + 
    \sum_{k=1}^n \exp\left(-\textstyle\frac{1}{2}\log(1+2\lambda\gamma_1) \phi_{\gamma}(k)\right)
        ( \norm{\theta_0 - \theta_{\star}}^2 + D_{n_0} )
    \Bigg]^{1/2}
    \\
    & \leq
    \frac{2\beta_0 K_1^{1/2}}{\lambda^{1/2}}
    \frac{\phi_{\gamma}^{1/2}(n)}{n}
    + 
    \frac{2\beta_0}{\lambda^{1/2}}\frac{\mathcal{A}^{1/2}}{n}
    ( \norm{\theta_0 - \theta_{\star}}^2 + D_{n_0} )^{1/2}
    ,
    \\
    \text{\eqref{eqn:PRbound3}}
    & \leq
    \frac{\gamma(\beta_0 + 1/\gamma_1)}{\lambda^{1/2} n}
    \Bigg[
    K_1^{1/2}\sum_{k=1}^n  k^{\gamma/2-1} 
    \\
    &\quad
    +
    \sum_{k=1}^n
    k^{\gamma-1}   
    \exp\left(-\textstyle\frac{1}{4}\log(1+2\lambda\gamma_1) \phi_{\gamma}(k)\right)
        ( \norm{\theta_0 - \theta_{\star}}^2 + D_{n_0} )^{1/2}
    \Bigg]
    \\
    & \leq
    \frac{\gamma K_1^{1/2}(\beta_0 + 1/\gamma_1)}{\lambda^{1/2}}
    \frac{\phi_{1-\gamma/2}(n)}{n}
    +
    \frac{\gamma(\beta_0 + 1/\gamma_1)}{\lambda^{1/2}}\frac{\mathcal{A}}{n}
    ( \norm{\theta_0 - \theta_{\star}}^2 + D_{n_0} )^{1/2}
    ,
\end{align*}
where 
$$
\mathcal{A} = \textstyle    \sum_{k=1}^{\infty}
    \exp\big(-\textstyle\frac{1}{4}\log(1+2\lambda\gamma_1) \phi_{\gamma}(k)\big)
    < \infty.
$$
The first inequality for \eqref{eqn:PRbound3} is due to $|\gamma_{k+1}^{-1} - \gamma_k^{-1}| = \frac{1}{\gamma_1}[(k+1)^{\gamma} - k^{\gamma}] = \frac{k^{\gamma}}{\gamma_1}[(1 + 1/k)^{\gamma} - 1] \leq \frac{\gamma}{\gamma_1}k^{\gamma-1}$.

In order to bound the term \eqref{eqn:PRbound4},
from \Cref{lem:l4bound},
\begin{align*}
    (\E{\norm{\iter{\theta}{n} - \iter{\theta}{\star}}^4})^{1/2}
    & \leq
    \tilde{K}_1^{1/2}n^{-\gamma}
    + \exp\Big(
    \textstyle\frac{1}{2}\nu(1 + \lambda\gamma_1/2)\phi_{\frac{5}{3}\gamma}(n) 
    - \textstyle\frac{1}{4}\log(1 + \lambda\gamma_1/2)\phi_{\gamma}(n)\Big)
    \\
    & \quad
    \times
    (\norm{\iter{\theta}{0} - \iter{\theta}{\star}}^4 
    + \frac{8(\beta_0^2 + \sigma^2)}{\lambda}\gamma_1\norm{\iter{\theta}{0} - \iter{\theta}{\star}}^3
    + \tilde{D}_{n_0})^{1/2}
    .
\end{align*}
It follows that
\begin{align*}
    \text{\eqref{eqn:PRbound4}}
    & \leq
    \frac{M}{2\lambda^{1/2}n}\sum_{k=1}^n\frac{\tilde{K}_1^{1/2}}{k^{\gamma}}
    +
    \frac{M}{2\lambda^{1/2}n}
    (\norm{\iter{\theta}{0} - \iter{\theta}{\star}}^4 
    + \frac{8(\beta_0^2 + \sigma^2)}{\lambda}\gamma_1\norm{\iter{\theta}{0} - \iter{\theta}{\star}}^3
    + D_{n_0})^{1/2} 
    \\
    & \quad
    \times
    \sum_{k=1}^n
    \exp\Big(
    \textstyle\frac{1}{2}\nu(1 + \lambda\gamma_1/2)\phi_{\frac{5}{3}\gamma}(k) 
    - \textstyle\frac{1}{4}\log(1 + \lambda\gamma_1/2)\phi_{\gamma}(k)\Big)  
    \\
    & \leq
    \frac{M\tilde{K}_1^{1/2}}{2\lambda^{1/2}}\frac{\phi_{\gamma}(n)}{n}
    +
    \frac{M}{2\lambda^{1/2}}\frac{\mathcal{B}}{n}
    (\norm{\iter{\theta}{0} - \iter{\theta}{\star}}^4 
    + \frac{8(\beta_0^2 + \sigma^2)}{\lambda}\gamma_1\norm{\iter{\theta}{0} - \iter{\theta}{\star}}^3
    + D_{n_0})^{1/2} 
    ,
\end{align*}
where
$$
    \mathcal{B} = \textstyle    \sum_{k=1}^{\infty}
    \exp\Big(
    \frac{1}{2}\nu(1 + \lambda\gamma_1/2)\phi_{\frac{5}{3}\gamma}(k) 
    - \textstyle\frac{1}{4}\log(1 + \lambda\gamma_1/2)\phi_{\gamma}(k)\Big)  
    < \infty
    .
$$

Combining these bounds, we get
\begin{align*}
    (\E{\norm{\iter{\bar{\theta}}{n}-\iter{\theta}{\star}}^2})^{1/2}
    &\leq
    \frac{1}{\sqrt{n}}\left[\tr{(\mathcal{H}(\iter{\theta}{\star})^{-1}\mathcal{I}(\iter{\theta}{\star})\mathcal{H}(\iter{\theta}{\star})^{-1})}\right]^{1/2}
    + 
    \frac{1/\gamma_1 + 3\beta_0}{\lambda^{1/2}  }\frac{\norm{\iter{\theta}{0}-\iter{\theta}{\star}}}{n}      
    \\
    &\quad
    +
    \frac{\beta_0\gamma_1 + 1}{\lambda^{1/2}\gamma_1}\frac{\tilde{K}_1^{1/2}}{ n^{1-\gamma/2}}    
    +
    \frac{\beta_0\gamma_1 + 1}{\lambda^{1/2} \gamma_1}
    \exp\left(-\textstyle\frac{1}{4}\log(1+2\lambda\gamma_1) \phi_{\gamma}(n)\right)
    ( \norm{\theta_0 - \theta_{\star}}^2 + \tilde{D}_{n_0} )^{1/2}
    \\
    &\quad
    +
    \frac{2\beta_0 \tilde{K}_1^{1/2}}{\lambda^{1/2}}
    \frac{\phi_{\gamma}^{1/2}(n)}{n}
    + 
    \frac{2\beta_0}{\lambda^{1/2}}\frac{\mathcal{A}^{1/2}}{n}
    ( \norm{\theta_0 - \theta_{\star}}^2 + \tilde{D}_{n_0} )^{1/2}
    \\
    &\quad
    +
    \frac{\gamma \tilde{K}_1^{1/2}(\beta_0 + 1/\gamma_1)}{\lambda^{1/2}}
    \frac{\phi_{1-\gamma/2}(n)}{n}
    +
    \frac{\gamma(\beta_0 + 1/\gamma_1)}{\lambda^{1/2}}\frac{\mathcal{A}}{n}
    ( \norm{\theta_0 - \theta_{\star}}^2 + \tilde{D}_{n_0} )^{1/2}
    \\
    & \quad
    +
    \frac{M\tilde{K}_1^{1/2}}{2\lambda^{1/2}}\frac{\phi_{\gamma}(n)}{n}
    +
    \frac{M}{2\lambda^{1/2}}\frac{\mathcal{B}}{n}
    (\norm{\iter{\theta}{0} - \iter{\theta}{\star}}^4 
    + \frac{8(\beta_0^2 + \sigma^2)}{\lambda}\gamma_1\norm{\iter{\theta}{0} - \iter{\theta}{\star}}^3
    + \tilde{D}_{n_0})^{1/2}     
    ,
\end{align*}
which further simplifies to
\begin{align*}
    (\E{\norm{\iter{\bar{\theta}}{n}-\iter{\theta}{\star}}^2})^{1/2}
    &\leq
    \frac{1}{\sqrt{n}}\left[\tr{(\mathcal{H}(\iter{\theta}{\star})^{-1}\mathcal{I}(\iter{\theta}{\star})\mathcal{H}(\iter{\theta}{\star})^{-1})}\right]^{1/2}
    \\
    &\quad
    +
    \frac{\tilde{K}_1^{1/2}}{\lambda^{1/2}n}\left(
    (\beta_0 + \gamma_1^{-1})n^{\gamma/2}
    +
    2\beta_0\phi_{\gamma}^{1/2}(n)   
    +
    \gamma(\beta_0 + \gamma_1^{-1})
    \phi_{1-\gamma/2}(n)
    +
    \frac{M}{2}\phi_{\gamma}(n)
    \right)
    \\
    &\quad
    + 
    \frac{1}{\lambda^{1/2}n}\Big(
    (\gamma_1^{-1} + 3\beta_0)\norm{\iter{\theta}{0}-\iter{\theta}{\star}}
    + 
    2\beta_0\mathcal{A}^{1/2}
    ( \norm{\theta_0 - \theta_{\star}}^2 + \tilde{D}_{n_0} )^{1/2}
    \\
    & \quad\quad\quad\quad
    +
    \gamma(\beta_0 + \gamma_1^{-1})\mathcal{A}
    ( \norm{\theta_0 - \theta_{\star}}^2 + \tilde{D}_{n_0} )^{1/2}
    \Big)
    \\
    & \quad
    +
    \frac{M\mathcal{B}}{2\lambda^{1/2}n}
    (\norm{\iter{\theta}{0} - \iter{\theta}{\star}}^4 
    + \textstyle\frac{8(\beta_0^2 + \sigma^2)}{\lambda}\gamma_1\norm{\iter{\theta}{0} - \iter{\theta}{\star}}^3
    + \tilde{D}_{n_0})^{1/2}     
    \\
    & \quad
    +
    \frac{\beta_0 + \gamma_1^{-1}}{\lambda^{1/2}}
    \exp\left(-\textstyle\frac{1}{4}\log(1+2\lambda\gamma_1) \phi_{\gamma}(n)\right)
    ( \norm{\theta_0 - \theta_{\star}}^2 + \tilde{D}_{n_0} )^{1/2}    
    .
\end{align*}
Below we put the explicit values of the constants:
\begin{align*}
    \tilde{\mathcal{A}} &= (\gamma_1^{-1} + 3\beta_0)\norm{\iter{\theta}{0}-\iter{\theta}{\star}}
    + 
    2\beta_0\mathcal{A}^{1/2}
    ( \norm{\theta_0 - \theta_{\star}}^2 + D_{n_0} )^{1/2}
    +
    \gamma(\beta_0 + \gamma_1^{-1})\mathcal{A}
    ( \norm{\theta_0 - \theta_{\star}}^2 + D_{n_0} )^{1/2}
    ,
    \\
    \tilde{\mathcal{B}} &= \mathcal{B} \cdot
    (\norm{\iter{\theta}{0} - \iter{\theta}{\star}}^4 
    + \textstyle\frac{8(\beta_0^2 + \sigma^2)}{\lambda}\gamma_1\norm{\iter{\theta}{0} - \iter{\theta}{\star}}^3
    + D_{n_0})^{1/2}     
    ,
    \\
    \mathcal{A} &= \textstyle    \sum_{k=1}^{\infty}
    \exp\big(-\textstyle\frac{1}{4}\log(1+2\lambda\gamma_1) \phi_{\gamma}(k)\big)
        ,
    \\
    \mathcal{B} &= \textstyle    \sum_{k=1}^{\infty}
    \exp\Big(
    \frac{1}{2}\nu(1 + \lambda\gamma_1/2)\phi_{\frac{5}{3}\gamma}(k) 
    - \textstyle\frac{1}{4}\log(1 + \lambda\gamma_1/2)\phi_{\gamma}(k)\Big)  
        ,
    \\
    \nu &= \textstyle 
        \frac{6(\beta_0^2 + \sigma^2)}{\lambda}\beta_0^{2/3}\gamma_1^{5/3} + 14\beta_0^2\gamma_1^{2} + 16\beta_0^3\gamma_1^{3} + 8\beta_0^4\gamma_1^{4}
        + \textstyle
        2K_2
        [\textstyle (10\sigma^2 + \frac{32\sigma(\beta_0^2 + \sigma^2)}{\lambda})\gamma_1^{2} + \frac{16(\beta_0^2 + \sigma^2)^2}{\lambda}\gamma_1^{3} + 8\gamma_1^{4}]
    ,
    \\
    K_2 &= (K_1 + \norm{\theta_0 - \theta_{\star}}^2 + D_{n_0}) / \gamma_1
    .
\end{align*}
\end{proof}

\subsection{Proof of \Cref{thm:nonstronglycvxRM}}\label{sec:proofs:RMerror2}
\begin{proof}[Proof of \Cref{thm:nonstronglycvxRM}]
    Pick a minimizer $\iter{\theta}{\star}$ of $L$.
    From Assumption \ref{enum:Lipschitz2}, $\nabla L$ is also $\beta_0$-Lipschitz continuous. Therefore,
    \begin{align}\label{eqn:objectivebound1}
        L(\iter{\theta}{n}) - L(\iter{\theta}{\star})
        &\leq
        L(\iter{\theta}{n-1}) - L(\iter{\theta}{\star})
        + \nabla L(\iter{\theta}{n-1})^T(\iter{\theta}{n} - \iter{\theta}{n-1}) + \frac{\beta_0}{2}\norm{\iter{\theta}{n} - \iter{\theta}{n-1}}^2
        \nonumber
        \\
        &\leq
        L(\iter{\theta}{n-1}) - L(\iter{\theta}{\star})
        + \nabla L(\iter{\theta}{n-1})^T(\iter{\theta}{n} - \iter{\theta}{n-1}) + \frac{\beta_0\gamma_n^2}{2}\norm{\nabla\ell(Z_n, \iter{\theta}{n})}^2
        ,
    \end{align}
    where the last inequality is due to equation \eqref{eq:implicit_sgd}.
    
    In order to find a recursive relation for the sequence $\Delta_n \triangleq \E{[L(\iter{\theta}{n}) - L(\iter{\theta}{\star})]}$, observe that
    \begin{align*}
        \norm{\nabla\ell(Z_n, \iter{\theta}{n})}^2
        &=
        \norm{\nabla\ell(Z_n, \iter{\theta}{n}) - \nabla\ell(Z_n, \iter{\theta}{\star}) + \nabla\ell(Z_n, \iter{\theta}{\star})}^2
        \\
        &\leq
        2\norm{\nabla\ell(Z_n, \iter{\theta}{n}) - \nabla\ell(Z_n, \iter{\theta}{\star})}^2 + 2\norm{\nabla\ell(Z_n, \iter{\theta}{\star})}^2
        \\
        &\leq
        2\beta_0^2\norm{\iter{\theta}{n} - \iter{\theta}{\star}}^2 + 2\norm{\nabla\ell(Z_n, \iter{\theta}{\star})}^2
        .
    \end{align*}
    Let $r_n ^2 = \delta_0 + \sigma^2\sum_{k=1}^n\gamma_k^2 = \norm{\iter{\theta}{0} - \iter{\theta}{\star}}^2 + \sigma^2\sum_{k=1}^n\gamma_k^2$.
    Since $\E{\norm{\iter{\theta}{n} - \iter{\theta}{\star}}^2} \le r_n ^2$ due to \Cref{lem:asibound}, we see
        \begin{equation}\label{eqn:gradbound1}
        \E{\norm{\nabla\ell(Z_n, \iter{\theta}{n})}^2}
        \leq
        2\beta_0^2\E{\norm{\iter{\theta}{n} - \iter{\theta}{\star}}^2}
        + 2\E{\norm{\nabla\ell(Z_n, \iter{\theta}{\star})}^2}
                        \leq
        2\beta_0^2 r_n^2 + 2\sigma^2
        .
    \end{equation}
        
    In addition, from equation \eqref{eq:implicit_sgd} and \Cref{lem:difference},
    \begin{align*}
        \nabla L(\iter{\theta}{n-1})^T(\iter{\theta}{n} - \iter{\theta}{n-1})
        & =
        -\gamma_n \nabla L(\iter{\theta}{n-1})^T \nabla\ell(Z_n, \iter{\theta}{n})
        \\
        & = 
        -\gamma_n \nabla L(\iter{\theta}{n-1})^T\nabla\ell(Z_n, \iter{\theta}{n-1}) 
        \\
        &\quad + \nabla L(\iter{\theta}{n-1})^T(\gamma_n[\nabla\ell(Z_n, \iter{\theta}{n-1}) + \nabla\ell(Z_n, \iter{\theta}{n})])
        \\
        & \leq
        -\gamma_n \nabla L(\iter{\theta}{n-1})^T\nabla\ell(Z_n, \iter{\theta}{n-1}) 
        + \norm{\nabla L(\iter{\theta}{n-1})}\norm{R_n}
        ,
    \end{align*}
    which leads to
    \begin{align*}
        \E{[\nabla L(\iter{\theta}{n-1})^T(\iter{\theta}{n} - \iter{\theta}{n-1}) | \mathcal{F}_{n-1}]}
        &\leq
        -\gamma_n \nabla L(\iter{\theta}{n-1})^T\nabla L( \iter{\theta}{n-1}) 
        + \norm{\nabla L(\iter{\theta}{n-1})}\E{[\norm{R_n} | \mathcal{F}_{n-1}]}
        \\
        & \leq
        -\gamma_n \norm{\nabla L(\iter{\theta}{n-1})}^2
        \\
        & \quad
        + \norm{\nabla L(\iter{\theta}{n-1}) - \nabla L(\iter{\theta}{\star})}
        \Big( \gamma_n^2\beta_0^2\norm{\iter{\theta}{n-1} - \iter{\theta}{\star}} + \frac{\beta_0^2 + \sigma^2}{2}\gamma_n^2 \Big)
        \\
        & \leq
        -\gamma_n \norm{\nabla L(\iter{\theta}{n-1})}^2
        \\
        & \quad
        + \beta_0\norm{\iter{\theta}{n-1} - \iter{\theta}{\star}}
        \Big( \gamma_n^2\beta_0^2\norm{\iter{\theta}{n-1} - \iter{\theta}{\star}} + \frac{\beta_0^2 + \sigma^2}{2}\gamma_n^2 \Big)
        \\
        & =
        -\gamma_n \norm{\nabla L(\iter{\theta}{n-1})}^2
        \\
        & \quad
        + 
        \gamma_n^2\beta_0^3\norm{\iter{\theta}{n-1} - \iter{\theta}{\star}}^2 + \frac{\beta_0(\beta_0^2 + \sigma^2)}{2}\gamma_n^2\norm{\iter{\theta}{n-1} - \iter{\theta}{\star}}
        .
    \end{align*}
    Then by convexity of $L$, 
    \[
        L(\iter{\theta}{n-1}) - L(\iter{\theta}{\star})
        \leq
        \nabla L(\iter{\theta}{n-1})^T(\iter{\theta}{n-1} - \iter{\theta}{\star})
        \leq
        \norm{\nabla L(\iter{\theta}{n-1})}
        \norm{\iter{\theta}{n-1} - \iter{\theta}{\star}}
        .
    \]
    Thus
    \begin{align*}
        \iter{\Delta}{n-1} = \E{[L(\iter{\theta}{n-1}) - L(\iter{\theta}{\star})]}
        & \leq
        \E{[\norm{\nabla L(\iter{\theta}{n-1})}
        \norm{\iter{\theta}{n-1} - \iter{\theta}{\star}}]}
        \\
        & \leq
        (\E{\norm{\nabla L(\iter{\theta}{n-1})}^2})^{1/2}
        (\E{\norm{\iter{\theta}{n-1} - \iter{\theta}{\star}}^2})^{1/2}
        \\
        &\leq
        \E{\norm{\nabla L(\iter{\theta}{n-1})}^2}^{1/2} r_n
    \end{align*}
    to obtain
    \[
        \E{\norm{\nabla L(\iter{\theta}{n-1})}^2} \geq \frac{1}{r_n^2}\iter{\Delta}{n-1}^2
        .
    \]
    Therefore
    \begin{align}\label{eqn:L1storderbound}
        \E{[\nabla L(\iter{\theta}{n-1})^T(\iter{\theta}{n} - \iter{\theta}{n-1})]}
        & \leq
        -\gamma_n \E{\norm{\nabla L(\iter{\theta}{n-1})}^2}
        + 
        \gamma_n^2\beta_0^3\E{\norm{\iter{\theta}{n-1} - \iter{\theta}{\star}}^2} + \frac{\beta_0(\beta_0^2 + \sigma^2)}{2}\gamma_n^2\E{\norm{\iter{\theta}{n-1} - \iter{\theta}{\star}}}
        \nonumber
        \\
        & \leq
        -\frac{\gamma_n}{r_n^2}\iter{\Delta}{n-1}^2
        + \gamma_n^2\beta_0^3r_n^2
        + \gamma_n^2\frac{\beta_0(\beta_0^2 + \sigma^2)}{2}r_n
        .
    \end{align}
    
    Combining inequalities \eqref{eqn:objectivebound1}, \eqref{eqn:gradbound1}, and \eqref{eqn:L1storderbound}, we have
    \begin{equation}\label{eqn:nonstrcvx_main}
        \iter{\Delta}{n} 
        \leq
        \iter{\Delta}{n-1} - \frac{\gamma_n}{r_n^2}\iter{\Delta}{n-1}^2
        + \frac{1}{2}\gamma_n^2\bar{\beta_n}\sigma^2
        ,
    \end{equation}
    where $\bar{\beta}_n = \sigma^{-2}(4\beta_0^3r_n^2 + \beta_0^3r_n + \beta_0\sigma^2r_n +2\beta_0\sigma^2)$.
    
    If we let $\psi_n(t) = t - \frac{\gamma_n}{r_n^2}t^2$, then $\iter{\Delta}{n} \leq \psi_n(\iter{\Delta}{n-1}) + \frac{1}{2}\gamma_n^2\bar{\beta_n}\sigma^2$.
    Since $\psi$ is increasing for $t \in [0, \frac{r_n^2}{2\gamma_n}]$ and 
    $\iter{\Delta}{n} = \E{[L(\iter{\theta}{n}) - L(\iter{\theta}{\star})]} \leq \E{[\nabla L(\iter{\theta}{\star})^T(\iter{\theta}{n} - \iter{\theta}{\star}) + \frac{\beta_0}{2}\norm{\iter{\theta}{n} - \iter{\theta}{\star}}^2]} = \frac{\beta_0}{2}\E{\norm{\iter{\theta}{n} - \iter{\theta}{\star}}^2]} \leq \frac{\beta_0}{2}r_n^2$, if we let $n_1 = \max\{\inf\{n \in \mathbb{N}: \gamma_n \leq 1/\beta_0\}, 3\}$, then $\psi_n(\iter{\Delta}{n})$ is increasing for $n \geq n_1$.
    
    Now let us consider a surrogate sequence defined as
    \[
        \iter{\tilde{\Delta}}{n} 
        =
        \iter{\tilde{\Delta}}{n-1} - \frac{\gamma_n}{r_n^2}\iter{\tilde{\Delta}}{n-1}^2 + \frac{1}{2}\gamma_n^2\bar{\beta}_n\sigma^2,
        \quad
        \iter{\tilde{\Delta}}{n_1} = \iter{\Delta}{n_1}
        .
    \]
    Then $\iter{\Delta}{n} \leq \iter{\tilde{\Delta}}{n}$ for $n \geq n_1$, since if we suppose $\iter{\Delta}{n-1} \leq \iter{\tilde{\Delta}}{n-1}$, then 
    \[
        \iter{\Delta}{n} 
        \leq 
        \psi_{n}(\iter{\Delta}{n-1}) + \frac{1}{2}\gamma_n^2\bar{\beta}_n\sigma^2
        \leq
        \psi_{n}(\iter{\tilde{\Delta}}{n-1}) + \frac{1}{2}\gamma_n^2\bar{\beta}_n\sigma^2
        = \iter{\tilde{\Delta}}{n}
    \]
    from the monotonicity of $\psi_n$.
    
    It suffices to bound $\iter{\tilde{\Delta}}{n}$. 
    Let $\varepsilon_n = (4\bar{\beta}_n^{1/2}\sigma\gamma_1^{3/2})^{-1}\min\{r_1, r_n n^{3\gamma/2 - 1}\}$, which is decreasing. 
    Since $\gamma_n^{1/2} - \gamma_{n-1}^{1/2} \geq \frac{\gamma_1^{1/2}}{4n^{\gamma/2}}$ for $n \geq 3$ (\Cref{prop:stepsize}), we have for all $n \geq n_1$,
    \begin{align*}
        \gamma_n^{1/2}(1 + \varepsilon_n)^{1/2} - \gamma_{n+1}^{1/2}(1 + \varepsilon_{n+1})^{1/2}
        &\geq 
        \gamma_n^{1/2}(1 + \varepsilon_{n+1})^{1/2} - \gamma_{n+1}^{1/2}(1 + \varepsilon_{n+1})^{1/2}
        \\
        &\geq 
        \frac{\gamma_1^{1/2}}{4n^{\gamma/2}}(1 + \varepsilon_{n+1})^{1/2}
        \geq 
        \frac{\gamma_1^{1/2}}{4n^{\gamma/2}}
        .
    \end{align*}
    On the other hand,
    \begin{align*}
        \varepsilon_n\bar{\beta}_n^{1/2}\sigma\gamma_n^2r_n^{-1}
        &=
        (1/4)\bar{\beta}_n^{-1/2}\sigma^{-1}\gamma_1^{-3/2}\bar{\beta}_n^{1/2}\sigma\gamma_1^2n^{-2\gamma}r_n^{-1}
        \min\{r_1, r_n n^{3\gamma/2 - 1}\}
        \\
        &=
        \frac{\gamma_1^{1/2}}{4n^{2\gamma}}
        \min\big\{\frac{r_1}{r_n}, n^{3\gamma/2 - 1}\big\}
        \leq
        \frac{\gamma_1^{1/2}}{4n^{2\gamma}}
        \leq
        \frac{\gamma_1^{1/2}}{4n^{\gamma/2}}
        .
    \end{align*}
    Thus
    \begin{equation}\label{eqn:gammabound}
        \gamma_n^{1/2}(1 + \varepsilon_n)^{1/2} - \gamma_{n+1}^{1/2}(1 + \varepsilon_{n+1})^{1/2}
        \geq 
        \varepsilon_n\bar{\beta}_n^{1/2}\sigma\gamma_n^2r_n^{-1}
        ,
        \quad
        n \geq n_1
        .
    \end{equation}
    
    Let $n_2 = \inf\{n \geq n_1: \iter{\tilde{\Delta}}{n-1}^2 \geq (1 + \varepsilon_n)\bar{\beta}_n\gamma_n^2\sigma^2 r_n^2/2 \}$. 
    Assume for now $n_2$ is finite.
    We want to show that 
    \begin{equation}\label{eqn:deltatilde2bound}
        \iter{\tilde{\Delta}}{n-1}^2 \geq (1 + \varepsilon_n)\bar{\beta}_n\gamma_n\sigma^2 r_n^2/2
        ,
        \quad
        n \geq n_2
        .
    \end{equation}
    Indeed, inequality \eqref{eqn:deltatilde2bound} is true for $n=n_2$ by construction. If $\iter{\tilde{\Delta}}{n-1} \geq (1 + \varepsilon_n)^{1/2}\bar{\beta}_n^{1/2}\gamma_n^{1/2}\sigma r_n/\sqrt{2}$, then since $\psi_n$ is increasing,
    \begin{align*}
        \iter{\tilde{\Delta}}{n}
        &=
        \psi_n(\iter{\tilde{\Delta}}{n-1}) + \frac{1}{2}\gamma_n^2\bar{\beta}_n\sigma^2
        \\
        &\geq 
        \psi_n((1 + \varepsilon_n)^{1/2}\bar{\beta}_n^{1/2}\gamma_n^{1/2}\sigma r_n/\sqrt{2}) + \frac{1}{2}\gamma_n^2\bar{\beta}_n\sigma^2
        \\
        &=
        (1 + \varepsilon_n)^{1/2}\bar{\beta}_n^{1/2}\gamma_n^{1/2}\sigma r_n/\sqrt{2}
        - \frac{\gamma_n}{2r_n^2}(1 + \varepsilon_n)\bar{\beta}_n\gamma_n\sigma^2 r_n^2
        + \frac{1}{2}\gamma_n^2\bar{\beta}_n\sigma^2
        \\
        &=
        \textstyle
        \big(\frac{1 + \varepsilon_n}{2}\big)^{1/2}\bar{\beta}_n^{1/2}\gamma_n^{1/2}\sigma r_n
        - \frac{1}{2}(1 + \varepsilon_n)\gamma_n^2\bar{\beta}_n\sigma^2
        + \frac{1}{2}\gamma_n^2\bar{\beta}_n\sigma^2
        \\
        &=
        \textstyle
        \big(\frac{1 + \varepsilon_{n+1}}{2}\big)^{1/2}\bar{\beta}_n^{1/2}\gamma_{n+1}^{1/2}\sigma r_n
        - \frac{1}{2}\varepsilon_n\gamma_n^2\bar{\beta}_n\sigma^2
        - \big(\frac{1 + \varepsilon_{n+1}}{2}\big)^{1/2}\bar{\beta}_n^{1/2}\gamma_{n+1}^{1/2}\sigma r_n
        + \big(\frac{1 + \varepsilon_n}{2}\big)^{1/2}\bar{\beta}_n^{1/2}\gamma_n^{1/2}\sigma r_n
        \\
        &=
        \textstyle
        \big(\frac{1 + \varepsilon_{n+1}}{2}\big)^{1/2}\bar{\beta}_n^{1/2}\gamma_{n+1}^{1/2}\sigma r_n
        - \frac{1}{2}\varepsilon_n\gamma_n^2\bar{\beta}_n\sigma^2
        + \frac{1}{\sqrt{2}}\bar{\beta}_n^{1/2}\sigma r_n
        [\gamma_n^{1/2}(1 + \varepsilon_n)^{1/2} - \gamma_{n+1}^{1/2}(1 + \varepsilon_{n+1})^{1/2}]
        \\
        & \stackrel{\eqref{eqn:gammabound}}{\geq}
        \textstyle
        \big(\frac{1 + \varepsilon_{n+1}}{2}\big)^{1/2}\bar{\beta}_n^{1/2}\gamma_{n+1}^{1/2}\sigma r_n
        - \frac{1}{2}\varepsilon_n\gamma_n^2\bar{\beta}_n\sigma^2
        + \frac{1}{\sqrt{2}}\varepsilon_n\gamma_n^2\bar{\beta}_n\sigma^2
        \\
        & \geq 
        \textstyle
        \big(\frac{1 + \varepsilon_{n+1}}{2}\big)^{1/2}\bar{\beta}_n^{1/2}\gamma_{n+1}^{1/2}\sigma r_n
        .
    \end{align*}
    Hence inequality \eqref{eqn:deltatilde2bound} holds for all $n \geq n_2$.
    
    Therefore, for $n \geq n_2$,
    \begin{align*}
        \iter{\tilde{\Delta}}{n} 
        &=
        \iter{\tilde{\Delta}}{n-1} - \frac{\gamma_n}{r_n^2}\iter{\tilde{\Delta}}{n-1}^2 + \frac{1}{2}\gamma_n^2\bar{\beta}_n\sigma^2
        \\
        & \stackrel{\eqref{eqn:deltatilde2bound}}{\leq}
        \iter{\tilde{\Delta}}{n-1} - \frac{\gamma_n}{r_n^2}\iter{\tilde{\Delta}}{n-1}^2 +
        \frac{\gamma_n}{^2}\frac{\iter{\tilde{\Delta}}{n-1}^2}{1 + \varepsilon_n}
        =
        \iter{\tilde{\Delta}}{n-1} -
        \frac{\gamma_n}{r_n^2}\frac{\varepsilon_n}{1 + \varepsilon_n}\iter{\tilde{\Delta}}{n-1}^2
    \end{align*}
    Divide the preceding inequality by $\iter{\tilde{\Delta}}{n-1}\iter{\tilde{\Delta}}{n}$ to obtain
    \[
        \iter{\tilde{\Delta}}{n-1}^{-1} 
        \leq 
        \iter{\tilde{\Delta}}{n}^{-1} - \frac{\gamma_n}{r_n^2}\frac{\varepsilon_n}{1 + \varepsilon_n}\frac{\iter{\tilde{\Delta}}{n-1}}{\iter{\tilde{\Delta}}{n}}
        \leq
        \iter{\tilde{\Delta}}{n}^{-1} - \frac{\gamma_n}{r_n^2}\frac{\varepsilon_n}{1 + \varepsilon_n}
        ,
    \]
    where the last inequality is from $\iter{\tilde{\Delta}}{n} \leq \iter{\tilde{\Delta}}{n-1}$. 
    It follows $\iter{\tilde{\Delta}}{n_2-1}^{-1} \leq \iter{\tilde{\Delta}}{n}^{-1} - \sum_{k=n_2}^n \frac{1}{r_k^2}\frac{\varepsilon_k}{1 + \varepsilon_k}\gamma_k$, or
    \[
        \iter{\tilde{\Delta}}{n} 
        \leq
        \frac{1}{\iter{\tilde{\Delta}}{n_2}^{-1} + \sum_{k=n_2}^n\frac{1}{r_k^2}\frac{\varepsilon_k}{1 + \varepsilon_k}\gamma_k},
        \quad
        n \ge n_2
        .
    \]
    By the definition of $n_2$, $\iter{\tilde{\Delta}}{n_2-1} \leq (1 + \varepsilon_{n_2})^{1/2}\bar{\beta}_{n_2}^{1/2}\gamma_{n_2}\sigma r_{n_2}/\sqrt{2}$. Hence
    $\iter{\tilde{\Delta}}{n_2}^{-1} \geq \iter{\tilde{\Delta}}{n_2-1}^{-1} \geq (1 + \varepsilon_{n_2})^{-1/2}\bar{\beta}_{n_2}^{-1/2}\gamma_{n_2}^{-1}\sigma^{-1}r_{n_2}^{-1}\sqrt{2}$.
    Now from inequality \eqref{eqn:gammabound}, since both $\varepsilon_n$ and $\gamma_n$ are decreasing, and $\bar{\beta}_n$ is increasing,
    \begin{align*}
        \gamma_{n+1}^{-1/2}(1 + \varepsilon_{n+1})^{-1/2} - \gamma_n^{-1/2}(1 + \varepsilon_n)^{-1/2}
        &\geq 
        \textstyle
        \bar{\beta}_n^{1/2}\sigma r_n^{-1}\frac{\varepsilon_n}{(1 + \varepsilon_n)^{1/2}(1 + \varepsilon_{n+1})^{1/2}}\frac{\gamma_n^2}{\gamma_n^{1/2}\gamma_{n+1}^{1/2}}
        \\
        & \geq
        \textstyle
        \sigma \bar{\beta}_n^{1/2}r_n^{-1}\frac{\varepsilon_n}{1 + \varepsilon_n}\gamma_n,
        \quad
        n \geq n_1
        .
    \end{align*}
    Also since $\gamma_{n_1}^{1/2}(1 + \varepsilon_{n_1})^{1/2} \geq \varepsilon_{n_1}\bar{\beta}_{n_1}^{1/2}\sigma\gamma_{n_1}r_{n_1}^{-1}$,
    \[
        \gamma_{n_1}^{-1/2}(1 + \varepsilon_{n_1})^{-1/2} 
        \geq 
        \textstyle
        \sigma\frac{\varepsilon_{n_1}}{1 + \varepsilon_{n_1}}\bar{\beta}_{n_1}^{1/2}r_{n_1}^{-1}\gamma_{n_1}
                        ,
        \quad
        n \geq n_1
        .
    \]
    Therefore
    \begin{align*}
        \gamma_n^{-1/2}(1 + \varepsilon_n)^{-1/2}
        &=
        \sum_{k=n_1}^{n-1}[\gamma_{k+1}^{-1/2}(1 + \varepsilon_{k+1})^{-1/2} - \gamma_k^{-1/2}(1 + \varepsilon_k)^{-1/2}] + \gamma_{n_1}^{-1/2}(1 + \varepsilon_{n_1})^{-1/2}
        \\
        &\geq 
        \textstyle
        \sigma\sum_{k=n_1}^{n-1}\frac{\bar{\beta}_k^{1/2}}{r_k}\frac{\varepsilon_k}{1 + \varepsilon_k}\gamma_k
        + 
        \sigma \frac{\bar{\beta}_{n_1}^{1/2}}{r_{n_1}}\frac{\varepsilon_{n_1}}{1 + \varepsilon_{n_1}}\gamma_{n_1}
        \\
        &\geq 
        \textstyle
        \sigma\sum_{k=n_1}^{n-1}\frac{\bar{\beta}_k^{1/2}}{r_k}\frac{\varepsilon_k}{1 + \varepsilon_k}\gamma_k
        .
    \end{align*}
    Since $\bar{\beta}_n^{1/2}/r_n$ is decreasing,
    \begin{equation}\label{eqn:deltan2}
        \iter{\tilde{\Delta}}{n_2}^{-1}
        \geq 
        \frac{\sqrt{2}}{\gamma_{n_2}}\frac{1}{r_{n_2}^2}\sum_{k=n_1}^{n_2 - 1}\frac{\bar{\beta}_k^{1/2}}{r_k}\frac{r_{n_2}}{\bar{\beta}_{n_2}^{1/2}}\frac{\varepsilon_k}{1 + \varepsilon_k}\gamma_k
        \geq 
        \frac{\sqrt{2}}{\gamma_{n_2}}\frac{1}{r_{n_2}^2}\sum_{k=n_1}^{n_2 - 1}\frac{\varepsilon_k}{1 + \varepsilon_k}\gamma_k
        .
    \end{equation}
    This entails, as $r_n^2$ is increasing, for $n \geq n_2$,
    \begin{equation}\label{eqn:deltatildebound}
        \iter{\tilde{\Delta}}{n} 
        \leq
        \frac{1}{\frac{\sqrt{2}}{\gamma_{n_2}}\frac{1}{r_{n_2}^2}\sum_{k=n_1}^{n_2 - 1}\frac{\varepsilon_k}{1 + \varepsilon_k}\gamma_k
        + \sum_{k=n_2}^n\frac{1}{r_k^2}\frac{\varepsilon_k}{1 + \varepsilon_k}\gamma_k}
        \leq
        \frac{\max\{\frac{\gamma_{n_2}}{\sqrt{2}}, 1\}r_n^2}{\sum_{k=n_1}^n\frac{\varepsilon_k}{1 + \varepsilon_k}\gamma_k}
        .
    \end{equation}
    Given the derivation of inequality \eqref{eqn:deltan2}, inequality \eqref{eqn:deltatildebound} holds for $n_1 \leq n < n_2$, thus it is true even if $n_2$ is infinite. Therefore, inequality \eqref{eqn:deltatildebound} holds for $n \geq n_1$.
    
    In order to obtain the rate, note
    \[
        r_n^2 = \delta_0 + \gamma_1^2\sum_{k=1}^n k^{-2\gamma}
        \leq
        \begin{cases}
            r^2 \triangleq \delta_0 + \gamma_1^2\zeta(2\gamma) < \infty, & \gamma \in (1/2, 1],
            \\
            \delta_0 + \gamma_1^2(1 + \phi_{2\gamma}(n)), & \gamma \in (0, 1/2]
            .
        \end{cases}
    \]
    Also, for $\gamma \in [2/3, 1]$, $\varepsilon_n=(4\bar{\beta}_n^{1/2}\sigma\gamma_1^{1/2})^{-1}r_1$. Since $\bar{\beta}_n \le \bar{\beta}_{\infty} \triangleq \sigma^2[4\beta_0^3r^2 + (\beta_0^3 + \beta_0\sigma^2)r + 2\beta_0\sigma^2] < \infty$,
    \begin{align*}
        \sum_{k=n_1}^n\frac{\varepsilon_k}{1 + \varepsilon_k}\gamma_k
        &=
        \sum_{k=n_1}^n\frac{1}{\varepsilon_k^{-1} + 1}\gamma_k
        =
        \sum_{k=n_1}^n\frac{1}{\frac{4\sigma\gamma_1^{1/2}}{r_1}\bar{\beta}_k^{1/2} + 1}\gamma_k
        \\
        & \geq 
        \frac{\gamma_1}{\frac{4\sigma\gamma_1^{1/2}}{r_1}\bar{\beta}_{\infty}^{1/2} + 1}\sum_{k=n_1}^n\frac{1}{k^{\gamma}}
                        \geq 
        \frac{\gamma_1/2}{\frac{4\sigma\gamma_1^{1/2}}{r_1}\bar{\beta}_{\infty}^{1/2} + 1}[\phi_{\gamma}(n) - \phi_{\gamma}(n_1 -1)]
        .
    \end{align*}
    If $\gamma \in (0, 2/3)$, then $\varepsilon_n=(4\bar{\beta}_n^{1/2}\sigma\gamma_1^{1/2})^{-1}r_n n^{3\gamma/2-1}$. Since $\bar{\beta}_n^{1/2}/r_n$ is decreasing,
    \begin{align*}
                        \sum_{k=n_1}^n\frac{1}{\varepsilon_k^{-1} + 1}\gamma_k
        &=
        \sum_{k=n_1}^n\frac{1}{4\sigma\gamma_1^{1/2}\frac{\bar{\beta}_k^{1/2}}{r_k}k^{1-3\gamma/2} + 1}\gamma_k
        =
        \sum_{k=n_1}^n\frac{\gamma_1}{4\sigma\gamma_1^{1/2}\frac{\bar{\beta}_k^{1/2}}{r_k}k^{1-\gamma/2} + k^{\gamma}}
        \\
        & \geq 
        \frac{\gamma_1}{4\sigma\gamma_1^{1/2}\frac{\bar{\beta}_1^{1/2}}{r_1} + 1}\sum_{k=n_1}^n\frac{1}{k^{1-\gamma/2}}
                        \geq 
        \frac{\gamma_1/2}{4\sigma\gamma_1^{1/2}\frac{\bar{\beta}_1^{1/2}}{r_1} + 1}[\phi_{1-\gamma/2}(n) - \phi_{1-\gamma/2}(n_1 -1)]
        .
    \end{align*}
    Therefore,
    \[
        \iter{\Delta}{n} 
        \leq
        \begin{cases}
            \frac{2\max\{\frac{\gamma_{n_1}}{\sqrt{2}}, 1\}}{4\sigma\gamma_1^{3/2}\bar{\beta}_1^{1/2}r_1^{-1} + \gamma_1}
            \frac{\delta_0 + \gamma_1^2(1+\phi_{2\gamma}(n))}{\phi_{1-\gamma/2}(n) - \phi_{1-\gamma/2}(n_1-1)}, & \gamma \in (0, 1/2],
            \\
            \frac{2\max\{\frac{\gamma_{n_1}}{\sqrt{2}}, 1\}}{4\sigma\gamma_1^{3/2}\bar{\beta}_1^{1/2}r_1^{-1} + \gamma_1}
            \frac{\delta_0 + \gamma_1^2\zeta(2\gamma)}{\phi_{1-\gamma/2}(n) - \phi_{1-\gamma/2}(n_1-1)}, & \gamma \in (1/2, 2/3),
            \\
            \frac{2\max\{\frac{\gamma_{n_1}}{\sqrt{2}}, 1\}}{4\sigma\gamma_1^{3/2}\bar{\beta}_{\infty}^{1/2}r_1^{-1} + \gamma_1}
            \frac{\delta_0 + \gamma_1^2\zeta(2\gamma)}{\phi_{\gamma}(n) - \phi_{\gamma}(n_1-1)}, & \gamma \in [2/3, 1],
        \end{cases}
    \]    
    for $n \geq n_1$ (note $\gamma_{n_2} \leq \gamma_{n_1}$).

Below we put the explicit values of the constants:
\begin{align*}
    \Gamma_1 &= \frac{2\max\{\frac{\gamma_{n_1}}{\sqrt{2}}, 1\}}{4\sigma\gamma_1^{3/2}\bar{\beta}_1^{1/2}r_1^{-1} + \gamma_1}
    ,
    \quad
    \Gamma_2 = \frac{2\max\{\frac{\gamma_{n_1}}{\sqrt{2}}, 1\}}{4\sigma\gamma_1^{3/2}\bar{\beta}_1^{1/2}r_1^{-1} + \gamma_1}
    \quad
    \Gamma_3 = \frac{2\max\{\frac{\gamma_{n_1}}{\sqrt{2}}, 1\}}{4\sigma\gamma_1^{3/2}\bar{\beta}_{\infty}^{1/2}r_1^{-1} + \gamma_1}
    \\
    r_n ^2 &= \delta_0 + \sigma^2\sum_{k=1}^n\gamma_k^2
    ,
    \\
	\bar{\beta}_n &= \sigma^{-2}(4\beta_0^3r_n^2 + \beta_0^3r_n + \beta_0\sigma^2r_n +2\beta_0\sigma^2)
	,
	\\
	\bar{\beta}_{\infty} &= \sigma^2[4\beta_0^3r^2 + (\beta_0^3 + \beta_0\sigma^2)r + 2\beta_0\sigma^2] < \infty
	,
	\\
	r &= \delta_0 + \sigma^2\sum_{k=1}^{\infty}\gamma_k^2
    .
\end{align*}
\end{proof}

\subsection{Proof of \Cref{thm:nonstronglycvxPR}}\label{sec:proofs:PRerror2}
\begin{proof}[Proof of \Cref{thm:nonstronglycvxPR}]
    Pick a minimizer $\iter{\theta}{\star}$ of $L$. 
        From the main iteration \eqref{eq:implicit_sgd}, we have $\iter{\theta}{k} - \iter{\theta}{\star} = \iter{\theta}{k-1} - \iter{\theta}{\star} - \gamma_k\nabla\ell(Z_k, \iter{\theta}{k})$ and
    \begin{align*}
        \norm{\iter{\theta}{k} - \iter{\theta}{\star}}^2
        &= 
        \norm{\iter{\theta}{k-1} - \iter{\theta}{\star}}^2
        - 2\gamma_k \nabla\ell(Z_k, \iter{\theta}{k})^T(\iter{\theta}{k-1} - \iter{\theta}{\star})
        + \gamma_k^2\norm{\nabla\ell(Z_k, \iter{\theta}{k})}^2
        \\
        &=
        \norm{\iter{\theta}{k-1} - \iter{\theta}{\star}}^2
        - 2\gamma_k \nabla\ell(Z_k, \iter{\theta}{k-1})^T(\iter{\theta}{k-1} - \iter{\theta}{\star})
        \\
        &\quad
        + 2\gamma_k (\nabla\ell(Z_k, \iter{\theta}{k-1}) - \nabla\ell(Z_k, \iter{\theta}{k}))^T(\iter{\theta}{k-1} - \iter{\theta}{\star})
        + \gamma_k^2\norm{\nabla\ell(Z_k, \iter{\theta}{k})}^2
        \\
        &\leq
        \norm{\iter{\theta}{k-1} - \iter{\theta}{\star}}^2
        - 2\gamma_k \nabla\ell(Z_k, \iter{\theta}{k-1})^T(\iter{\theta}{k-1} - \iter{\theta}{\star})
        \\
        &\quad
        + 2\norm{R_k}\norm{\iter{\theta}{k-1} - \iter{\theta}{\star}}
        + \gamma_k^2\norm{\nabla\ell(Z_{k-1}, \iter{\theta}{k})}^2
        ,
    \end{align*}
    where the final inequality is due to \Cref{lem:difference}.
    Therefore
    \begin{equation}\label{eqn:nonstrcvx_errorbound1}
        \begin{split}
        \E{[\norm{\iter{\theta}{k} - \iter{\theta}{\star}}^2 | \mathcal{F}_{k-1}]}
        &\leq
        \norm{\iter{\theta}{k-1} - \iter{\theta}{\star}}^2
        - 2\gamma_k \nabla L(\iter{\theta}{k-1})^T(\iter{\theta}{k-1} - \iter{\theta}{\star})
       \\
        &\quad
        + 2\E{[\norm{R_k} | \mathcal{F}_{k-1}]}\norm{\iter{\theta}{k-1} - \iter{\theta}{\star})}
        + \gamma_k^2\E{[\norm{\nabla\ell(Z_{k-1}, \iter{\theta}{k})}^2 | \mathcal{F}_{k-1}]}        
        .
        \end{split}
    \end{equation}
    From the cocoercivity of Lipschitz continuous gradient operator \citep{Bauschke:ConvexAnalysisAndMonotoneOperatorTheoryIn:2011}, we see
    \[
        \beta_0^{-1}\norm{\nabla\ell(Z_k, \iter{\theta}{k-1}) - \nabla\ell(Z_k, \iter{\theta}{\star})}^2
        \leq
        (\nabla\ell(Z_k, \iter{\theta}{k-1}) - \nabla\ell(Z_k, \iter{\theta}{\star}))^T(\iter{\theta}{k-1} - \iter{\theta}{\star})
    \]
    and thus
    \begin{align*}
        \norm{\nabla\ell(Z_k, \iter{\theta}{k-1})}^2
        &=
        \norm{\ell(Z_k, \iter{\theta}{k-1}) - \ell(Z_k, \iter{\theta}{\star}) + \ell(Z_k, \iter{\theta}{\star})}^2
        \\
        &\leq
        2\norm{\ell(Z_k, \iter{\theta}{k-1}) - \ell(Z_k, \iter{\theta}{\star})}^2 + 2\norm{\ell(Z_k, \iter{\theta}{\star})}^2
        \\
        &\leq
        2\beta_0(\nabla\ell(Z_k, \iter{\theta}{k-1}) - \nabla\ell(Z_k, \iter{\theta}{\star}))^T(\iter{\theta}{k-1} - \iter{\theta}{\star})
        + 2\norm{\ell(Z_k, \iter{\theta}{\star})}^2
        .
    \end{align*}
    Therefore from inequality \eqref{eqn:nonstrcvx_errorbound1}
    \begin{align*}
        \E{\norm{\iter{\theta}{k} - \iter{\theta}{\star}}^2}
        & \leq
        \E{\norm{\iter{\theta}{k-1} - \iter{\theta}{\star}}^2}
        - 2\gamma_k \E{[\nabla L(\iter{\theta}{k-1})^T(\iter{\theta}{k-1} - \iter{\theta}{\star})]}
        \\
        & \quad
        + 2\gamma_k^2\beta_0^2 \E{\norm{\iter{\theta}{k-1} - \iter{\theta}{\star}}^2}
        + (\beta_0^2 + \sigma^2)\gamma_k^2 \E{\norm{\iter{\theta}{k-1} - \iter{\theta}{\star}}^2}
        \\
        & \quad
        + 2\gamma_k^2\beta_0 \E{[\nabla L(\iter{\theta}{k-1})^T(\iter{\theta}{k-1} - \iter{\theta}{\star})]}
        + 2\gamma_k^2\sigma^2
        \\
        & \leq
        \E{\norm{\iter{\theta}{k-1} - \iter{\theta}{\star}}^2}
        - 2\gamma_k(1 - \beta_0\gamma_k)\E{[\nabla L(\iter{\theta}{k-1})^T(\iter{\theta}{k-1} - \iter{\theta}{\star})]}
        + 2\gamma_k^2\tilde{\sigma}^2
        ,
    \end{align*}
    where $\tilde{\sigma}^2 = \beta_0^2 r^2 + r(\beta_0^2 + \sigma^2)/2 + \sigma^2$;
    the last inequality is due to \Cref{lem:asibound}.
    Let $\iter{\delta}{k}$ denote $\E{\norm{\iter{\theta}{k} - \iter{\theta}{\star}}^2}$. The previous inequality implies
    \[
        2\gamma_k(1 - \beta_0\gamma_k) \E{[\nabla L(\iter{\theta}{k-1})^T(\iter{\theta}{k-1} - \iter{\theta}{\star})]}
        \leq
        \iter{\delta}{k-1} - \iter{\delta}{k} + 2\gamma_k^2\tilde{\sigma}^2
        .
    \]
    If we let $n_* = \inf\{k \in \mathbb{N}: (1 - \gamma_k\beta_0) \geq 1/2 \}$, then
    $\E{[\nabla L(\iter{\theta}{k-1})^T(\iter{\theta}{k-1} - \iter{\theta}{\star})]} \leq \gamma_k^{-1}(\iter{\delta}{k-1} - \iter{\delta}{k} + 2\gamma_k^2\tilde{\sigma}^2)$ for all $k \geq n_*$.
    
    Lipschitz continuity of $\nabla L$ implies
    $L(\iter{\theta}{k-1}) - L(\iter{\theta}{\star}) \leq \frac{\beta_0}{2}\norm{\iter{\theta}{k-1} - \iter{\theta}{\star}}^2$.
    Also, by convexity, $L(\iter{\theta}{k-1}) - L(\iter{\theta}{\star}) \leq \nabla L(\iter{\theta}{k-1})^T(\iter{\theta}{k-1} - \iter{\theta}{\star})$.
    Therefore, for $n > n_*$,
    \begin{align*}
        L(\iter{\bar{\theta}}{n}) - L(\iter{\theta}{\star})
        &= L(\frac{1}{n}\sum_{k=0}^{n-1}\iter{\theta}{k}) - L(\iter{\theta}{\star})
        \\
        &\leq 
        \frac{1}{n}\sum_{k=1}^n(L(\iter{\theta}{k-1}) - L(\iter{\theta}{\star}))
        \\
        &=
        \frac{1}{n}\sum_{k=1}^{n_*}(L(\iter{\theta}{k-1}) - L(\iter{\theta}{\star}))
        +
        \frac{1}{n}\sum_{k=n_* + 1}^n(L(\iter{\theta}{k-1}) - L(\iter{\theta}{\star}))
        \\
        &\leq
        \frac{\beta_0}{2n}\sum_{k=1}^{n_*}\norm{\iter{\theta}{k-1} - \iter{\theta}{\star}}^2 + \frac{1}{n}\sum_{k=n_* + 1}^n \nabla L(\iter{\theta}{k-1})^T(\iter{\theta}{k-1} - \iter{\theta}{\star})
        .
    \end{align*}
    Then it follows 
    \begin{align*}
        \E{[L(\iter{\bar{\theta}}{n}) - L(\iter{\theta}{\star})]}
        &\leq
        \frac{1}{n}\Big(
        \frac{\beta_0}{2}\sum_{k=1}^{n_*}\iter{\delta}{k-1} 
        + \sum_{k=n_*+1}^n\gamma_k^{-1}(\iter{\delta}{k-1} - \iter{\delta}{k} + 2\gamma_k^2\tilde{\sigma}^2) \Big)
        .
    \end{align*}
    Observe  that $\delta_{k-1} \leq r_k^2 \triangleq \delta_0 + \sigma^2\sum_{j=1}^k\gamma_j^2 \leq \delta_0 + \sigma^2\gamma_1^2[1 + \phi_{2\gamma}(k)]$. 
    Since $1 + \phi_{2\gamma}(k) \leq k^{1-2\gamma}/(1-2\gamma)$ if $\gamma < 1/2$ and $\phi_{2\gamma}(k) = \log k$ if $\gamma = 1/2$, and $\sum_{j=1}^k j^{-2\gamma} \leq \sum_{j=1}^{\infty} j^{-2\gamma} = \zeta(2\gamma) < \infty$ if $\gamma > 1/2$ where $\zeta(\cdot)$ is the Riemann zeta function, it follows that
    \[
        \sum_{k=1}^{n_*} r_k^2\leq
        \begin{cases} 
        n_*\delta_0 + \frac{\sigma^2\gamma_1^2}{1-2\gamma}\phi_{2\gamma-1}(n_*), & \gamma < 1/2, \\
        n_*\delta_0 + \sigma^2\gamma_1^2\log(n_* + 1), & \gamma = 1/2, \\
        n_*[\delta_0 + \sigma^2\gamma_1^2\zeta(2\gamma)], & \gamma > 1/2.
        \end{cases}
    \]
    On the other hand, 
    \begin{align*}
        \sum_{n_*+1}^n \gamma_k^{-1}(\delta_{k-1} - \delta_k) 
        &=
        \gamma_{n_*+1}^{-1}\delta_{n_*} + \sum_{k=n_*+1}^{n-1}\delta_k(\gamma_{k+1}^{-1} - \gamma_k^{-1}) - \gamma_n^{-1}\delta_n
        \\
        &\leq
        r_n^2\gamma_{n_*+1}^{-1} + r_n^2\sum_{k=n_*+1}^{n-1}(\gamma_{k+1}^{-1} - \gamma_k^{-1}) - \gamma_n^{-1}\delta_n
        \\
        &= 
        r_n^2 \gamma_n^{-1} - \gamma_n^{-1}\delta_n
        \leq
        r_n^2\gamma_n^{-1}
    \end{align*}
    \citep[p. 27]{bach2011}.
    Thus
    \begin{align*}
        \E{[L(\iter{\bar{\theta}}{n}) - L(\iter{\theta}{\star})]}
        &\leq
        \frac{1}{n}\Big(
        \frac{\beta_0}{2}\sum_{k=1}^{n_*}r_k^2
        + r_n^2\gamma_n^{-1} + 2\tilde{\sigma}^2\sum_{k=1}^n\gamma_k
        \Big)
        \\
        &\leq
        \frac{1}{n}\Big(
        \frac{\beta_0}{2}\sum_{k=1}^{n_*}r_k^2
        + \delta_0\gamma_n^{-1} + \sigma^2\gamma_1^2[1 + \phi_{2\gamma}(n)]\gamma_n^{-1} 
        + 2\tilde{\sigma}^2\gamma_1\phi_{\gamma}(n)
        \Big)
        \\
        &=
        \frac{\beta_0}{2n}\sum_{k=1}^{n_*}r_k^2
        + \frac{\delta_0}{\gamma_1 n^{1-\gamma}}
        + \sigma^2\gamma_1^2\frac{1 + \phi_{2\gamma}(n)}{n^{1-\gamma}}
        + 2\tilde{\sigma}^2\gamma_1\frac{\phi_{\gamma}(n)}{n}
        \\
        &\leq
        \begin{cases}
            \frac{\beta_0[n_*\delta_0 + \frac{\sigma^2\gamma_1^2}{1-2\gamma}\phi_{2\gamma-1}(n_*)]}{2n}
            + \frac{\sigma^2\gamma_1^2}{(1 - 2\gamma)n^{\gamma}}
            + \frac{2\tilde{\sigma}^2\gamma_1}{(1 - \gamma)n^{\gamma}},
            &
            \gamma < 1/2,
            \\
            \frac{\beta_0[n_*\delta_0 + \sigma^2\gamma_1^2\log(n_* + 1)]}{2n}
            + \sigma^2\gamma_1^2\frac{1 + \log n}{\sqrt{n}}
            + \frac{2\tilde{\sigma}^2\gamma_1}{(1 - \gamma)\sqrt{n}},
            &
            \gamma = 1/2
            .
        \end{cases}
    \end{align*}
    When $\gamma > 1/2$, we can replace the $1 + \phi_{2\gamma}(n)$ with $\zeta(2\gamma)$, hence
    \begin{align*}
        \E{[L(\iter{\bar{\theta}}{n}) - L(\iter{\theta}{\star})]}
        & \leq
        \begin{cases}
            \frac{\beta_0n_*[\delta_0 + \sigma^2\gamma_1^2\zeta(2\gamma)]}{2n}
            + \frac{\sigma^2\gamma_1^2\zeta(2\gamma)}{n^{1-\gamma}}
            + \frac{2\tilde{\sigma}^2\gamma_1}{(1-\gamma)n^{\gamma}},
            & \gamma \in (1/2, 1),
            \\
            \frac{\beta_0n_*[\delta_0 + \sigma^2\gamma_1^2\zeta(2\gamma)]}{2n}
            + \sigma^2\gamma_1^2\zeta(2)
            + \frac{2\tilde{\sigma}^2\gamma_1\log n}{(1-\gamma)n},
            & \gamma = 1.
        \end{cases}
    \end{align*}
    
    The preceding argument immediately yields that for $n \leq n_*$,
    \begin{align*}
        \E{[L(\iter{\bar{\theta}}{n}) - L(\iter{\theta}{\star})]}
        & \leq
        \begin{cases}
            \frac{\beta_0}{2}\Big(
            \delta_0 + \frac{\sigma^2\gamma_1^2}{(1-2\gamma)(2 - 2\gamma)}n^{1-2\gamma}
            \Big), & \gamma < 1/2, 
            \\
            \frac{\beta_0}{2}\Big(\delta_0 + \sigma^2\gamma_1^2\frac{\log(n+1)}{n} \Big), & \gamma = 1/2, 
            \\
            \frac{\beta_0}{2}\Big(\delta_0 + \frac{\sigma^2\gamma_1^2\zeta(2\gamma)}{n} \Big), & \gamma > 1/2,
        \end{cases}
    \end{align*}
    Since $\phi_{2\gamma-1}(n) \leq n^{2-2\gamma}/(2 - 2\gamma)$ when $\gamma < 1/2$.
    
Below we put the explicit values of the constants:
\begin{align*}
    \tilde{\Gamma}_1 &= \frac{\beta_0[n_*\delta_0 + \frac{\sigma^2\gamma_1^2}{1-2\gamma}\phi_{2\gamma-1}(n_*)]}{2}
    ,
    \quad
    \tilde{\Gamma}_2 = \frac{\beta_0[n_*\delta_0 + \sigma^2\gamma_1^2\log(n_* + 1)]}{2}
    \quad
    \tilde{\Gamma}_3 = \frac{\beta_0n_*[\delta_0 + \sigma^2\gamma_1^2\zeta(2\gamma)]}{2}
    ,
    \\
    \tilde{\sigma}^2 &= \beta_0^2 r^2 + r(\beta_0^2 + \sigma^2)/2 + \sigma^2
    ,
    \\
	r &= \delta_0 + \sigma^2\sum_{k=1}^{\infty}\gamma_k^2
    .
\end{align*}    
\end{proof}

\section{Proofs of technical lemmas}\label{sec:technical}

\begin{proof}[Proof of \Cref{lem:difference}]
    Note from the ISGD update \eqref{eq:spp} that $\iter{\theta}{n-1} - \iter{\theta}{n} = \gamma_n\nabla\ell(Z_n, \iter{\theta}{n})$. Therefore
    \begin{align*}
        \gamma_n[\ell(Z_n, \iter{\theta}{n}) - \ell(Z_n, \iter{\theta}{n-1})]
        \le
        \gamma_n \nabla\ell(Z_n, \iter{\theta}{n})^T(\iter{\theta}{n} - \iter{\theta}{n-1})
        = -\norm{\iter{\theta}{n} - \iter{\theta}{n-1}}^2
        ,
    \end{align*}
    where the first inequality follows from the convexity of $\ell(Z_n, \cdot)$, i.e., Assumptions \ref{enum:CCP}.
    Using this convexity once more, we see $\ell(Z_n, \iter{\theta}{n}) - \ell(Z_n, \iter{\theta}{n-1}) \ge \nabla\ell(Z_n, \iter{\theta}{n-1})^T(\iter{\theta}{n} - \iter{\theta}{n-1})$. It follows that
    $$
        \norm{\iter{\theta}{n} - \iter{\theta}{n-1}}^2 \le \gamma_n\nabla\ell(Z_n, \iter{\theta}{n-1})(\iter{\theta}{n-1} - \iter{\theta}{n})
        \le \gamma_n\norm{\nabla\ell(Z_n, \iter{\theta}{n-1})}\norm{\iter{\theta}{n-1} - \iter{\theta}{n}}
    $$
    and the claim is proved.
\end{proof}

\begin{proof}[Proof of \Cref{lemma:toulis}]
    Let us first consider the case $n < n_0$. Since $c_n \downarrow 0$,
    $Q_{i+1}^n < (1 + c_{i+1})\dotsb(1 + c_n) \le (1 + c_1)^{n_0}$. 
    By expanding inequality \eqref{eqn:recursion}, we see
    \begin{equation}\label{eqn:n0bound}
    \begin{aligned}
        y_n &\leq Q_1^n y_0 + \sum_{i=1}^n Q_{i+1}^n a_i 
        \leq Q_1^n y_0 + (1 + c_1)^{n_0}\sum_{i=1}^n a_i 
        \\
        &\leq Q_1^n y_0 + (1 + c_1)^{n_0} A \\
        &\leq K_0\frac{a_n}{b_n} + Q_1^n y_0 + (1 + c_1)^{n_0} A + B
        ,
    \end{aligned}
    \end{equation}
    where the last line is inequality \eqref{eqn:finitesample}.
    
    Now consider the case $n \geq n_0$. Since
    $$
        \frac{1 + (1 + \delta) b_1}{1 + \delta - \delta_n - \zeta_n}
        \leq
        \frac{1 + (1 + \delta) b_1}{1 + \delta - \delta_{n_0} - \zeta_{n_0}} = K_0
    $$
    and $b_n \downarrow 0$, it follows that
    $K_0(\delta_n + \zeta_n) + 1 + (1 + \delta) b_n \leq K_0(1 + \delta)$, or
    $$
        \frac{K_0}{a_n}\left(\frac{a_{n-1}}{b_{n-1}} - \frac{a_n}{b_n} + \frac{c_n a_{n-1}}{b_{n-1}}\right) + 1 + (1 + \delta) b_n 
        \leq
        K_0 (1 + \delta)
        .
    $$
    This implies
    $$
        a_n[1 + (1 + \delta) b_n] \leq
                        K_0\left( \frac{[1 + (1 + \delta)b_n]a_n}{b_n} - \frac{(1 + c_n)a_{n-1}}{b_{n-1}}\right)
    $$
    and
    \begin{equation}\label{eqn:anbound}
        a_n \leq
        K_0\left( \frac{a_n}{b_n} - \frac{1 + c_n}{1 + (1 + \delta)b_n}\frac{a_{n-1}}{b_{n-1}}\right)
        .
    \end{equation} 
    Combining inequalities \eqref{eqn:recursion} and \eqref{eqn:anbound}, we obtain
    $$
        y_n - K_0\frac{a_n}{b_n}
        \leq
        \frac{1 + c_n}{1 + (1 + \delta)b_n}
        \left( y_{n-1} - K_0\frac{a_{n-1}}{b_{n-1}}\right)
        .
    $$
    Define $s_n = y_n - K_0 a_n / b_n$. Then $|s_n| \le (1 + c_n)[1 + (1 + \delta)b_n]^{-1}|s_{n-1}|$ and thus $|s_n| \leq Q_{n_0 + 1}^n |s_{n_0}|$. Therefore,
    \begin{align*}
        y_n - K_0\frac{a_n}{b_n} \leq |s_n|
        &\leq
        Q_{n_0+1}^n\left|y_{n_0} - K_0\frac{a_{n_0}}{b_{n_0}}\right|
        \\
        &\leq
        Q_{n_0+1}^n\left(y_{n_0} + K_0\frac{a_{n_0}}{b_{n_0}}\right)
        \\
        &=
        Q_{n_0+1}^n y_{n_0} + Q_{n_0+1}^n B
        \\
        &\leq
        Q_1^n y_0 + Q_{n_0+1}^n(1 + c_1)^{n_0} A + Q_{n_0+1}^n B
        ,
    \end{align*}
    where the last line follows from inequality \eqref{eqn:n0bound}.
\end{proof}

\begin{proof}[Proof of \Cref{cor:toulis}]
It is easy to verify that
\[
    1 - \eta n^{-\gamma}  + \nu n^{-\beta} 
    \le
    \frac{1 + \nu(1+\eta)n^{-\beta}}{1 + \eta n^{-\gamma}}
\]
for all $n \geq 1$.
Let $a_n = a_1n^{-\alpha}$, $b_n = \eta n^{-\gamma}/(1+\delta)$, $c_n = \nu(1+\eta)n^{-\beta}$.
Then inequality \eqref{eqn:recursion} in \Cref{lemma:toulis} holds for $(a_n, b_n, c_n)$.

To see if the conditions for \Cref{lemma:toulis} hold, verify that $a_n \downarrow 0$, 
$\sum_{n=1}^{\infty}a_n < \infty$,
$b_n \downarrow 0$, $c_n \downarrow 0$, 
$c_n/b_n \propto n^{-(\beta-\alpha)} \downarrow 0$,
hence $c_n / b_n < 1$ for all sufficiently large $n$, and
\begin{align*}
    \delta_n &= \frac{1}{a_n}\left(\frac{a_{n-1}}{b_{n-1}} - \frac{a_n}{b_n}\right) 
            = \frac{1 + \delta}{\eta}n^{\gamma}\left(\left[\frac{n}{n-1}\right]^{\alpha - \gamma} - 1\right) 
            \downarrow
            \delta,
            \\
    \zeta_n &= \frac{c_n}{b_{n-1}}\frac{a_{n-1}}{a_n}  = \nu(1 + \eta)(1 + \delta)\left(\frac{n}{n-1}\right)^{\alpha-\gamma} n^{-(\beta-\gamma)}
            \downarrow 0
    .
\end{align*}
Recall that $\delta = \frac{1+\delta}{\eta}(\alpha - \gamma)$ if $\gamma = 1$.

Therefore inequality \eqref{eqn:finitesample} from \Cref{lemma:toulis} translates to
\begin{equation}\label{eqn:finitesamplebound1}
    \begin{split}
    y_n
    &\leq
    K_1 n^{-(\alpha - \gamma)}
    + Q_1^n y_0
    + Q_{n_0+1}^n\left[(1 + c_1)^{n_0} A
    + B\right]
    ,
    \end{split}
\end{equation}
where $K_1$, $A$, and $B$ are given in equations \eqref{eqn:toulis:K1} and \eqref{eqn:toulis:AB}; the conditions for the $n_0$ translates to inequalities \eqref{eqn:toulis:n0}.
Furthermore,
\[
    Q_i^n = \begin{cases} \prod_{j=i}^n \frac{1 + \nu(1+\eta) n^{-\beta}}{1 + \eta n^{-\gamma}}, & n \ge i, \\
        1, & n < i.
    \end{cases}
\]
Since $b_n, c_n \downarrow 0$, 
    $Q_1^n = \prod_{j=1}^n \frac{1 + c_j}{1+(1+\delta)b_j} \geq \prod_{j=1}^n \frac{1}{1+ (1+\delta) b_j} \ge (1 + \eta)^{-n}$.
Hence 
\begin{equation}\label{eqn:finitesamplebound2}
    Q_{n_0+1}^n = Q_1^n / Q_1^{n_0} \le (1 +  \eta)^{n_0} Q_1^n
    = (1 + \eta)^{n_0} Q_1^n
    .
\end{equation}
In order to bound $Q_1^n$, take a logarithm to see
\[
    \log Q_1^n = \sum_{k=1}^n\log(1 + \nu(1+\eta) k^{-\beta}) 
        - \sum_{k=1}^n \log(1 + \eta k^{-\gamma})
    .
\]
For the first term, use $\log(1+x) \le x$ for $x \ge 0$ to get
\begin{align*}
    \sum_{k=1}^n \log(1 + \nu(1+\eta) k^{-\beta}) 
    &\leq 
    \nu(1+\eta) \sum_{k=1}^{n} \frac{1}{k^{\beta}}
    \leq 
    \begin{cases}
        \nu(1+\eta) \sum_{k=1}^{\infty} \frac{1}{k^{\beta}}, & \beta > 1, \\
        \nu(1+\eta) \phi_{\beta}(n), & \beta \leq 1,
    \end{cases}
        ,
\end{align*}
since $\sum_{k=1}^n k^{-\beta} \leq \phi_{\beta}(n)$ for $\beta \leq 1$.
For the second term, since $x \mapsto x^{-1}\log(1+x)$ is decreasing for $x > -1$ and $k^{-\gamma} \downarrow 0$, we have
    $\log(1+\eta k^{-\gamma}) \geq k^{-\gamma}\log(1+\eta)$
to get
\begin{equation}\label{eqn:Qbound}
    \begin{split}
    -\sum_{i=1}^n\log(1+\eta k^{-\gamma}) 
    &\le 
    -\log(1+\eta)\sum_{k=1}^n\frac{1}{k^{\gamma}}
    \le
    -\frac{1}{2}\log(1+\eta) \phi_{\gamma}(n)
    ,
    \end{split}
\end{equation}
since $\sum_{k=1}^n k^{-\gamma} \ge \frac{1}{2}\phi_{\gamma}(n)$ for $\gamma \leq 1$.
Thus
\begin{equation}\label{eqn:finitesamplebound3}
    Q_1^n \le K_2(n) \exp\left(-\textstyle\frac{1}{2}\log(1+\eta) \phi_{\gamma}(n)\right)
    ,
\end{equation}
where $K_2(n)$ is given in equation \eqref{eqn:toulis:K2}.
Combining inequalities \eqref{eqn:finitesamplebound1}, \eqref{eqn:finitesamplebound2}, and \eqref{eqn:finitesamplebound3},
we finally obtain inequality \eqref{eqn:y_n_bound}, with $D_{n_0}$ given in equation \eqref{eqn:toulis:D_n0}.    
\end{proof}

\begin{proof}[Proof of \Cref{lem:l4bound}]
    Recall that $\iter{\theta}{n} - \iter{\theta}{\star} = \iter{\theta}{n-1} - \iter{\theta}{\star} - \gamma_n\nabla\ell(Z_n, \iter{\theta}{n})$. Therefore,
    \begin{align}\label{eqn:quadratic}
        \norm{\iter{\theta}{n} - \iter{\theta}{\star}}^2
        &=
        \norm{\iter{\theta}{n-1} - \iter{\theta}{\star}}^2
        - 2\gamma_n (\iter{\theta}{n-1} - \iter{\theta}{\star})^T\nabla\ell(Z_n, \iter{\theta}{n})
        + \gamma_n^2\norm{\nabla\ell(Z_n, \iter{\theta}{n})}^2
        \nonumber\\
        &\leq
        \norm{\iter{\theta}{n-1} - \iter{\theta}{\star}}^2
        - 2\gamma_n W_n
        + \gamma_n^2 V_n
        ,
    \end{align}
    where $W_n = (\iter{\theta}{n-1} - \iter{\theta}{\star})^T\nabla\ell(Z_n, \iter{\theta}{n})$ and $V_n = \norm{\nabla\ell(Z_n, \iter{\theta}{n-1})}^2$; the last line follows from \Cref{lem:difference}.
    
    We first bound the fourth moment. Squaring both sides of inequality  \eqref{eqn:quadratic} yields
    \begin{align*}
        \norm{\iter{\theta}{n} - \iter{\theta}{\star}}^4
        &\leq
        \norm{\iter{\theta}{n-1} - \iter{\theta}{\star}}^4
        + 4\gamma_n^2 W_n^2 + \gamma_n^4 V_n^2
        \\
        & \quad
        - 4\gamma_n \norm{\iter{\theta}{n-1} - \iter{\theta}{\star}}^2 W_n
        + 2\gamma_n^2 \norm{\iter{\theta}{n-1} - \iter{\theta}{\star}}^2 V_n
        - 4\gamma_n^3 W_n V_n
        .
    \end{align*}
    In order to bound $\E{[\norm{\iter{\theta}{n} - \iter{\theta}{\star}}^4 | \mathcal{F}_{n-1}]}$, first note that
    \begin{align}\label{eqn:Wnbound0}
        \E{[W_n | \mathcal{F}_{n-1}]} 
        &= \E{[(\iter{\theta}{n-1} - \iter{\theta}{\star})^T\nabla\ell(Z_n, \iter{\theta}{n-1}) - \gamma_n^{-1}(\iter{\theta}{n-1} - \iter{\theta}{\star})^T R_n) | \mathcal{F}_{n-1}]}
        \nonumber\\
        &\geq
        (\iter{\theta}{n-1} - \iter{\theta}{\star})^T \E{[\nabla\ell(Z_n, \iter{\theta}{n-1}) | \mathcal{F}_{n-1}]}
        - \gamma_n^{-1}\norm{\iter{\theta}{n-1} - \iter{\theta}{\star}}\E{[\norm{R_n} | \mathcal{F}_{n-1}]}
        \nonumber\\
        &\geq 
        (\iter{\theta}{n-1} - \iter{\theta}{\star})^T \E{[\nabla\ell(Z_n, \iter{\theta}{n-1}) | \mathcal{F}_{n-1}]}
        - \gamma_n\beta_0^2 \norm{\iter{\theta}{n-1} - \iter{\theta}{\star}}^2 - \gamma_n(\beta_0^2 + \sigma^2)\norm{\iter{\theta}{n-1} - \iter{\theta}{\star}}
        \nonumber\\
        &= 
        (\iter{\theta}{n-1} - \iter{\theta}{\star})^T(\nabla L(\iter{\theta}{n-1}) - \nabla L(\iter{\theta}{\star}))
        - \gamma_n\beta_0^2 \norm{\iter{\theta}{n-1} - \iter{\theta}{\star}}^2 - \gamma_n(\beta_0^2 + \sigma^2)\norm{\iter{\theta}{n-1} - \iter{\theta}{\star}}
        \nonumber\\
        &\geq
        \lambda\norm{\iter{\theta}{n-1} - \iter{\theta}{\star}}^2
        - \gamma_n\beta_0^2 \norm{\iter{\theta}{n-1} - \iter{\theta}{\star}}^2 - \gamma_n(\beta_0^2 + \sigma^2)\norm{\iter{\theta}{n-1} - \iter{\theta}{\star}}
        .
    \end{align}
    The first inequality is Cauchy-Schwarz, the second is inequality \eqref{eqn:Rnbound}, and the last inequality is from the strong convexity of $L$.
    In addition,
    \begin{equation}\label{eqn:Vnbound0}
        \E{[V_n | \mathcal{F}_{n-1}]} 
        \leq
        2\beta_0^2\norm{\iter{\theta}{n-1} - \iter{\theta}{\star}}^2 + 2\sigma^2
    \end{equation}
    from inequality \eqref{eqn:gradconditionalbound}, and
    \begin{align*}
        \E{[-W_n V_n | \mathcal{F}_{n-1}]}
        &=
        -\E{[(\iter{\theta}{n-1} - \iter{\theta}{\star})^T\nabla\ell(Z_n, \iter{\theta}{n})\norm{\nabla\ell(Z_n, \iter{\theta}{n-1})}^2 | \mathcal{F}_{n-1}]}
        \\
        &\leq
        \norm{\iter{\theta}{n-1} - \iter{\theta}{\star}}\E{[\norm{\nabla\ell(Z_n, \iter{\theta}{n})}\norm{\nabla\ell(Z_n, \iter{\theta}{n-1})}^2 | \mathcal{F}_{n-1}]}
        \\
        &\leq
        \norm{\iter{\theta}{n-1} - \iter{\theta}{\star}}\E{[\norm{\nabla\ell(Z_n, \iter{\theta}{n-1})}^3 | \mathcal{F}_{n-1}]}
    \end{align*}
    again by \Cref{lem:difference}.
    Since $\norm{\nabla\ell(Z_n, \iter{\theta}{n-1})} \leq \beta_0\norm{\iter{\theta}{n-1} - \iter{\theta}{\star}} + \norm{\nabla\ell(Z_n, \iter{\theta}{\star})}$ by Assumption \ref{enum:Lipschitz2}, combining with Assumption \ref{enum:noiselevelcond2} we have
    \begin{align*}
        \E{[\norm{\nabla\ell(Z_n, \iter{\theta}{n-1})}^3 | \mathcal{F}_{n-1}]}
        &\leq
        4\beta_0^3\norm{\iter{\theta}{n-1} - \iter{\theta}{\star}}^3
        +
        4\E{\norm{\nabla\ell(Z_n, \iter{\theta}{\star})}^3}
        \\
        &\leq
        4\beta_0^3\norm{\iter{\theta}{n-1} - \iter{\theta}{\star}}^3 + 4\sigma^3
        .
    \end{align*}
    by \Cref{prop:normpower}.
    Therefore 
    \begin{equation}\label{eqn:WnVnbound}
        \E{[-W_n V_n | \mathcal{F}_{n-1}]}
        \leq
        4\beta_0^3\norm{\iter{\theta}{n-1} - \iter{\theta}{\star}}^4 + 4\sigma^3\norm{\iter{\theta}{n-1} - \iter{\theta}{\star}}
        .
    \end{equation}
    
    Finally,
    \begin{align}\label{eqn:Wn2bound}
        \E{[W_n^2 | \mathcal{F}_{n-1}]}
        &=
        \E{[((\iter{\theta}{n-1} - \iter{\theta}{\star})^T\nabla\ell(Z_n, \iter{\theta}{n}))^2 | \mathcal{F}_{n-1}]}
        \nonumber\\
        &\leq
        \norm{\iter{\theta}{n-1} - \iter{\theta}{\star}}^2
        \E{[\norm{\ell(Z_n, \iter{\theta}{n})}^2 | \mathcal{F}_{n-1}]}
        \nonumber\\
        &\leq
        \norm{\iter{\theta}{n-1} - \iter{\theta}{\star}}^2
        \E{[\norm{\ell(Z_n, \iter{\theta}{n-1})}^2 | \mathcal{F}_{n-1}]}
        \nonumber\\
        &\leq
        2\beta_0^2\norm{\iter{\theta}{n-1} - \iter{\theta}{\star}}^4 + 2\sigma^2\norm{\iter{\theta}{n-1} - \iter{\theta}{\star}}^2
    \end{align}
    from inequality \eqref{eqn:Vnbound0}, and
    \begin{align}\label{eqn:Vn2bound}
        \E{[V_n^2 | \mathcal{F}_{n-1}]}
        &\leq
        \E{[\norm{\ell(Z_n, \iter{\theta}{n-1})}^4 | \mathcal{F}_{n-1}]}
        \nonumber\\
        &\leq
        8\beta_0^4\norm{\iter{\theta}{n-1} - \iter{\theta}{\star}}^4 + 8\E{[\norm{\ell(Z_n, \iter{\theta}{\star})}^4 | \mathcal{F}_{n-1}]}
        \nonumber\\
        &\leq
        8\beta_0^4\norm{\iter{\theta}{n-1} - \iter{\theta}{\star}}^4 + 8\sigma^4
    \end{align}
    from \Cref{prop:normpower} and Assumption \ref{enum:noiselevelcond2}.
    
    Combining inequalities \eqref{eqn:Wnbound0}, \eqref{eqn:Vnbound0}, \eqref{eqn:WnVnbound}, \eqref{eqn:Wn2bound}, and \eqref{eqn:Vn2bound}, we see
    \begin{equation}\label{eqn:fourthmoment}
    \begin{split}
        \E{[\norm{\iter{\theta}{n} - \iter{\theta}{\star}}^4 | \mathcal{F}_{n-1}]}
        &\leq
        (1 - 4\lambda\gamma_n + 14\beta_0^2\gamma_n^2 + 16\beta_0^3\gamma_n^3 + 8\beta_0^4\gamma_n^4)\norm{\iter{\theta}{n-1} - \iter{\theta}{\star}}^4
        \\
        &\quad
        + 4(\beta_0^2 + \sigma^2)\gamma_n^2 \norm{\iter{\theta}{n-1} - \iter{\theta}{\star}}^3
        + 2\sigma^2(\gamma_n^2 + 4\gamma_n^4)\norm{\iter{\theta}{n-1} - \iter{\theta}{\star}}^2
        \\
        &\quad
        + 16\sigma^3\gamma_n^3\norm{\iter{\theta}{n-1} - \iter{\theta}{\star}}
        + 8\sigma^4\gamma_n^4
        \\
        &\leq
        (1 - 4\lambda\gamma_n + 14\beta_0^2\gamma_n^2 + 16\beta_0^3\gamma_n^3 + 8\beta_0^4\gamma_n^4)\norm{\iter{\theta}{n-1} - \iter{\theta}{\star}}^4
        \\
        &\quad
        + 4(\beta_0^2 + \sigma^2)\gamma_n^2 \norm{\iter{\theta}{n-1} - \iter{\theta}{\star}}^3
        \\
        &\quad
        + 2(5\sigma^2\gamma_n^2 + 4\gamma_n^4)\norm{\iter{\theta}{n-1} - \iter{\theta}{\star}}^2
        + 16\sigma^4\gamma_n^4
        ,
    \end{split}
    \end{equation}
    where the last inequality follows from 
    $16\sigma^3\gamma_n^3\norm{\iter{\theta}{n-1} - \iter{\theta}{\star}} \leq 8\sigma^2\gamma_n^2\norm{\iter{\theta}{n-1} - \iter{\theta}{\star}}^2 + 8\sigma^4\gamma_n^4$.
    
    We now bound the third moment. Multiplying inequality \eqref{eqn:quadratic} with 
    \begin{align*}
        \norm{\iter{\theta}{n} - \iter{\theta}{\star}}
        &\leq
        \norm{\iter{\theta}{n-1} - \iter{\theta}{\star}}
        + \gamma_n\norm{\nabla\ell(Z_n, \iter{\theta}{n})}
        \\
        &\leq
        \norm{\iter{\theta}{n-1} - \iter{\theta}{\star}} + \gamma_n V_n^{1/2}
    \end{align*}
    yields
    \begin{align*}
        \norm{\iter{\theta}{n} - \iter{\theta}{\star}}^3
        &\leq
        \norm{\iter{\theta}{n-1} - \iter{\theta}{\star}}^3
        -2\gamma_n \norm{\iter{\theta}{n} - \iter{\theta}{\star}} W_n
        + \gamma_n^2 \norm{\iter{\theta}{n} - \iter{\theta}{\star}} V_n
        \\
        &\quad
        + \gamma_n \norm{\iter{\theta}{n} - \iter{\theta}{\star}}^2 V_n^{1/2}
        - 2\gamma_n W_n V_n^{1/2}
        + \gamma_n^3 V_n^{3/2}
        .
    \end{align*}
    In addition to inequalities \eqref{eqn:Vnbound0} and \eqref{eqn:Vn2bound}, we have
    \begin{align*}
        \E{[V_n^{3/2} | \mathcal{F}_{n-1}]}
        &= \E{[\norm{\nabla\ell(Z_n, \iter{\theta}{n-1})}^3 | \mathcal{F}_{n-1}]}
        \leq 4\beta_0^3 \norm{\iter{\theta}{n} - \iter{\theta}{\star}}^3 + 4\sigma^3
        \\
        \E{[V_n^{1/2} | \mathcal{F}_{n-1}]}
        &= \E{[\norm{\nabla\ell(Z_n, \iter{\theta}{n-1})} | \mathcal{F}_{n-1}]}
        \leq 
        \beta_0[\norm{\nabla\ell(Z_n, \iter{\theta}{n-1})} 
        + \E{\norm{\nabla\ell(Z_n, \iter{\theta}{\star})}}
        \\
        &\leq
        \beta_0[\norm{\nabla\ell(Z_n, \iter{\theta}{n-1})} 
        + \sigma
        \\
        \E{[-W_nV_n^{1/2} | \mathcal{F}_{n-1}]}
        &=
        -\E{[(\iter{\theta}{n-1} - \iter{\theta}{\star})^T\nabla\ell(Z_n, \iter{\theta}{n})\norm{\nabla\ell(Z_n, \iter{\theta}{n-1})} | \mathcal{F}_{n-1}]}
        \\
        &\leq
        \norm{\iter{\theta}{n-1} - \iter{\theta}{\star}}\E{[\norm{\nabla\ell(Z_n, \iter{\theta}{n})}\norm{\nabla\ell(Z_n, \iter{\theta}{n-1})} | \mathcal{F}_{n-1}]}
        \\
        &\leq
        \norm{\iter{\theta}{n-1} - \iter{\theta}{\star}}\E{[\norm{\nabla\ell(Z_n, \iter{\theta}{n-1})}^2 | \mathcal{F}_{n-1}]}
        \\
        &\stackrel{\eqref{eqn:Vnbound0}}{\leq}
        2\beta_0^2 \norm{\iter{\theta}{n-1} - \iter{\theta}{\star}}^3
        + 2\sigma^2 \norm{\iter{\theta}{n-1} - \iter{\theta}{\star}}
        .
    \end{align*}
    Therefore
    \begin{align}\label{eqn:thirdmoment}
        \E{[\norm{\iter{\theta}{n} - \iter{\theta}{\star}}^3 | \mathcal{F}_{n-1}]}
        &\leq
        [1 + (\beta_0 - 2\lambda)\gamma_n + 8\beta_0^2\gamma_n^2 + 4\beta_0^3\gamma_n^3]\norm{\iter{\theta}{n} - \iter{\theta}{\star}}^3
        \nonumber\\
        &\quad
        +
        [\sigma\gamma_n + 2(\beta_0^2 + \sigma^2)\gamma_n^2]\norm{\iter{\theta}{n} - \iter{\theta}{\star}}^2
        +
        6\sigma^2\gamma_n^2 \norm{\iter{\theta}{n} - \iter{\theta}{\star}}
        +
        4\sigma^3\gamma_n^3
        \nonumber\\
        &\leq
        [1 + (\beta_0 - 2\lambda)\gamma_n + 8\beta_0^2\gamma_n^2 + 4\beta_0^3\gamma_n^3]\norm{\iter{\theta}{n} - \iter{\theta}{\star}}^3
        \nonumber\\
        &\quad
        +
        [4\sigma\gamma_n + 2(\beta_0^2 + \sigma^2)\gamma_n^2]\norm{\iter{\theta}{n} - \iter{\theta}{\star}}^2
        +
        7\sigma^3\gamma_n^3
    \end{align}
    since $6\sigma^2\gamma_n^2\norm{\iter{\theta}{n} - \iter{\theta}{\star}} \leq 3\sigma\gamma_n \norm{\iter{\theta}{n} - \iter{\theta}{\star}}^2 + 3\sigma^3\gamma_n^3$.
    
    Now, let
    \[
        U_n = \norm{\iter{\theta}{n} - \iter{\theta}{\star}}^4 + c\beta_0\gamma_{n+1}\norm{\iter{\theta}{n} - \iter{\theta}{\star}}^3,
        \quad
        c = \frac{8(\beta_0^2 + \sigma^2)}{\beta_0\lambda}
        .
    \]
    Then, from inequalities \eqref{eqn:fourthmoment} and \eqref{eqn:thirdmoment}
    \begin{align*}
        \E{[U_n | \mathcal{F}_{n-1}]}
        &\leq
        \norm{\iter{\theta}{n-1} - \iter{\theta}{\star}}^4
        [1 - 4\lambda\gamma_n + 14\beta_0^2\gamma_n^2 + 16\beta_0^3\gamma_n^3 + 8\beta_0^4\gamma_n^4]
        \\
        &\quad
        +
        \norm{\iter{\theta}{n-1} - \iter{\theta}{\star}}^3
        [4(\beta_0^2 + \sigma^2)\gamma_n^2 + c\beta_0\gamma_{n+1}(1 + (\beta_0 - 2\lambda)\gamma_n + 8\beta_0^2\gamma_n^2 + 4\beta_0^3\gamma_n^3)]
        \\
        &\quad
        +
        \norm{\iter{\theta}{n-1} - \iter{\theta}{\star}}^2
        [10\sigma^2\gamma_n^2 + 8\gamma_n^4 + c\beta_0\gamma_{n+1}(4\sigma\gamma_n + 2(\beta_0^2+\sigma^2)\gamma_n^2)]
        \\
        &\quad
        +
        16\sigma^4\gamma_n^4 + 7c\beta_0\sigma^3\gamma_{n+1}\gamma_n^3
        \\
        &\leq
        \norm{\iter{\theta}{n-1} - \iter{\theta}{\star}}^4
        [\textstyle 1 - \frac{1}{2}\lambda\gamma_n + 14\beta_0^2\gamma_n^2 + 16\beta_0^3\gamma_n^3 + 8\beta_0^4\gamma_n^4]
        \\
        &\quad
        +
        \norm{\iter{\theta}{n-1} - \iter{\theta}{\star}}^3
        [4(\beta_0^2 + \sigma^2)\gamma_n^2 + c\beta_0\gamma_{n}(1 - 2\lambda\gamma_n + 8\beta_0^2\gamma_n^2 + 4\beta_0^3\gamma_n^3)]
        \\
        &\quad
        +
        c\beta_0^2\gamma_n^2\norm{\iter{\theta}{n} - \iter{\theta}{\star}}^3
        \\
        &\quad
        +
        \norm{\iter{\theta}{n-1} - \iter{\theta}{\star}}^2
        [10\sigma^2\gamma_n^2 + 8\gamma_n^4 + c\beta_0\gamma_{n}(4\sigma\gamma_n + 2(\beta_0^2+\sigma^2)\gamma_n^2)]
        \\
        &\quad
        +
        16\sigma^4\gamma_n^4 + 7c\beta_0\sigma^3\gamma_n^4
        \\        
        &\leq
        \norm{\iter{\theta}{n-1} - \iter{\theta}{\star}}^4
        [\textstyle 1 - \frac{1}{2}\lambda\gamma_n + \frac{3c}{4}\beta_0^{5/3}\gamma_n^{5/3} + 14\beta_0^2\gamma_n^2 + 16\beta_0^3\gamma_n^3 + 8\beta_0^4\gamma_n^4]
        \\
        &\quad
        +
        \norm{\iter{\theta}{n-1} - \iter{\theta}{\star}}^3
        [4(\beta_0^2 + \sigma^2)\gamma_n^2 + c\beta_0\gamma_{n}(1 - 2\lambda\gamma_n + 8\beta_0^2\gamma_n^2 + 4\beta_0^3\gamma_n^3)]
        \\
        &\quad
        +
        \norm{\iter{\theta}{n-1} - \iter{\theta}{\star}}^2
        [10\sigma^2\gamma_n^2 + 8\gamma_n^4 + c\beta_0\gamma_{n}(4\sigma\gamma_n + 2(\beta_0^2+\sigma^2)\gamma_n^2)]
        \\
        &\quad
        +
        \frac{c}{4}\beta_0^3\gamma_n^3 + 16\sigma^4\gamma_n^4 + 7c\beta_0\sigma^3\gamma_n^4
        \\
        &\leq
        (\norm{\iter{\theta}{n-1} - \iter{\theta}{\star}}^4 + c\beta_0\gamma_n \norm{\iter{\theta}{n} - \iter{\theta}{\star}}^3)
        \\
        &\quad
        \times
        [\textstyle 1 - \frac{1}{2}\lambda\gamma_n + \frac{3c}{4}\beta_0^{5/3}\gamma_n^{5/3} + 14\beta_0^2\gamma_n^2 + 16\beta_0^3\gamma_n^3 + 8\beta_0^4\gamma_n^4]
        \\
        &\quad
        + 
        \norm{\iter{\theta}{n-1} - \iter{\theta}{\star}}^3
        [4(\beta_0^2 + \sigma^2)\gamma_n^2 + c\beta_0\gamma_{n}(1 - \lambda\gamma_n + 8\beta_0^2\gamma_n^2 + 4\beta_0^3\gamma_n^3)
        \\
        &\quad\quad
        - c\beta_0\gamma_n(\textstyle 1 - \frac{1}{2}\lambda\gamma_n + \frac{3c}{4}\beta_0^{5/3}\gamma_n^{5/3} + 14\beta_0^2\gamma_n^2 + 16\beta_0^3\gamma_n^3 + 8\beta_0^4\gamma_n^4)
        ]
        \\
        &\quad
        +
        \norm{\iter{\theta}{n-1} - \iter{\theta}{\star}}^2
        [10\sigma^2\gamma_n^2 + 8\gamma_n^4 + c\beta_0\gamma_{n}(4\sigma\gamma_n + 2(\beta_0^2+\sigma^2)\gamma_n^2)]
        \\
        &\quad
        +
        \frac{c}{4}\beta_0^3\gamma_n^3 + 16\sigma^4\gamma_n^4 + 7c\beta_0\sigma^3\gamma_n^4
        \\        
        &\leq
        (\norm{\iter{\theta}{n-1} - \iter{\theta}{\star}}^4 + c\beta_0\gamma_n \norm{\iter{\theta}{n} - \iter{\theta}{\star}}^3)
        \\
        &\quad
        \times
        [\textstyle 1 - \frac{1}{2}\lambda\gamma_n + \frac{3c}{4}\beta_0^{5/3}\gamma_n^{5/3} + 14\beta_0^2\gamma_n^2 + 16\beta_0^3\gamma_n^3 + 8\beta_0^4\gamma_n^4]
        \\
        &\quad
        + 
        \norm{\iter{\theta}{n-1} - \iter{\theta}{\star}}^3
        [\textstyle (4(\beta_0^2 + \sigma^2)-\frac{1}{2}c\beta_0\lambda)\gamma_n^2 
                ]
        \\
        &\quad
        +
        \norm{\iter{\theta}{n-1} - \iter{\theta}{\star}}^2
        [10\sigma^2\gamma_n^2 + 8\gamma_n^4 + c\beta_0\gamma_{n}(4\sigma\gamma_n + 2(\beta_0^2+\sigma^2)\gamma_n^2)]
        \\
        &\quad
        +
        \frac{c}{4}\beta_0^3\gamma_n^3 + 16\sigma^4\gamma_n^4 + 7c\beta_0\sigma^3\gamma_n^4    
        \\
        &=
        U_{n-1}
        [\textstyle 1 - \frac{1}{2}\lambda\gamma_n + \frac{6(\beta_0^2 + \sigma^2)}{\lambda}\beta_0^{2/3}\gamma_n^{5/3} + 14\beta_0^2\gamma_n^2 + 16\beta_0^3\gamma_n^3 + 8\beta_0^4\gamma_n^4]
        \\
        &\quad
        +
        \norm{\iter{\theta}{n-1} - \iter{\theta}{\star}}^2
        [\textstyle (10\sigma^2 + \frac{32\sigma(\beta_0^2 + \sigma^2)}{\lambda})\gamma_n^2 + \frac{16(\beta_0^2 + \sigma^2)^2}{\lambda}\gamma_n^3 + 8\gamma_n^4]
        \\
        &\quad
        + \textstyle
        \frac{2(\beta_0^2 + \sigma^2)}{\lambda}\beta_0^2\gamma_n^3 
        + (16\sigma^4 + \frac{56(\beta_0^2 + \sigma^2)}{\lambda}\sigma^3)\gamma_n^4   
        ,
    \end{align*}
    where the third inequality is due to Young's inequality
    \[
        c\beta_0^2\gamma_n^2\norm{\iter{\theta}{n-1} - \iter{\theta}{\star}}^3
        \leq
        \textstyle
        \frac{3c}{4}\beta_0^{5/3}\gamma_n^{5/3}\norm{\iter{\theta}{n-1} - \iter{\theta}{\star}}^4 + \frac{c}{4}\beta_0^3\gamma_n^3
        .
    \]
    Therefore, 
    \begin{align*}
        \E{[U_n]} 
        &\leq
        [\textstyle 1 - \frac{1}{2}\lambda\gamma_n + \frac{6(\beta_0^2 + \sigma^2)}{\lambda}\beta_0^{2/3}\gamma_n^{5/3} + 14\beta_0^2\gamma_n^2 + 16\beta_0^3\gamma_n^3 + 8\beta_0^4\gamma_n^4]\E{[U_{n-1}}]
        \\
        &\quad
        +
        [\textstyle (10\sigma^2 + \frac{32\sigma(\beta_0^2 + \sigma^2)}{\lambda})\gamma_n^2 + \frac{16(\beta_0^2 + \sigma^2)^2}{\lambda}\gamma_n^3 + 8\gamma_n^4]
        \E{\norm{\iter{\theta}{n-1} - \iter{\theta}{\star}}^2}
        \\
        &\quad
        + \textstyle
        \frac{2(\beta_0^2 + \sigma^2)}{\lambda}\beta_0^2\gamma_n^3 
        + (16\sigma^4 + \frac{56(\beta_0^2 + \sigma^2)}{\lambda}\sigma^3)\gamma_n^4   
    \end{align*}
    Since from \Cref{thm:finite_sample} 
    \[
        \E{\norm{\theta_n - \theta_{\star}}^2} \le 
        (K_1 + \norm{\theta_0 - \theta_{\star}}^2 + D_{n_0})n^{-\gamma} 
        = K_2 \gamma_n, 
        \quad K_2 = (K_1 + \norm{\theta_0 - \theta_{\star}}^2 + D_{n_0}) / \gamma_1
    \]
    and $\gamma_{n-1} \leq 2\gamma_n$ (\Cref{prop:stepsize}),
    \begin{align*}
        \E{[U_n]}
        &\leq
        (\textstyle 1 - \frac{1}{2}\lambda\gamma_n + C_0\gamma_n^{5/3})\E{[U_{n-1}]}
        +
        C_1\gamma_n^3
        ,
    \end{align*}
    where
    \begin{align*}
        C_0 &= \textstyle
        \frac{6(\beta_0^2 + \sigma^2)}{\lambda}\beta_0^{2/3} + 14\beta_0^2\gamma_1^{1/3} + 16\beta_0^3\gamma_1^{4/3} + 8\beta_0^4\gamma_1^{7/3}
        \\
        &\quad
        + \textstyle
        2K_2
        [\textstyle (10\sigma^2 + \frac{32\sigma(\beta_0^2 + \sigma^2)}{\lambda})\gamma_1^{1/3} + \frac{16(\beta_0^2 + \sigma^2)^2}{\lambda}\gamma_1^{4/3} + 8\gamma_1^{7/3}]
        \\
        C_1 &= \textstyle
        \frac{2(\beta_0^2 + \sigma^2)}{\lambda}\beta_0^2
        + (16\sigma^4 + \frac{56(\beta_0^2 + \sigma^2)}{\lambda}\sigma^3)\gamma_1
        .
    \end{align*}
    
    It follows from \Cref{cor:toulis} that 
    \begin{align*}
        \E{[U_n]}
        & \leq
        \tilde{K}_1 n^{-2\gamma}
        + \exp\Big(\textstyle
        \nu(1 + \frac{\lambda\gamma_1}{2})\phi_{\frac{5}{3}\gamma}(n) 
        - \frac{1}{2}\log(1 + \frac{\lambda\gamma_1}{2})\phi_{\gamma}(n)\Big)
        (U_0 + \tilde{D}_{\tilde{n}_0})
        ,
    \end{align*}    
    where $\nu = C_0\gamma_1^{5/3}$ and the $\tilde{n}_0$ is as
    given in equation \eqref{eqn:n0tilde}. The other constants are as given in equation \eqref{eqn:tildeconstants}.
                                                                                                                                                                Noting $\E{\norm{\iter{\theta}{n} - \iter{\theta}{\star}}^4} \leq \E{[U_n]}$ completes the proof.
\end{proof}

\end{document}